\documentclass{article}

\PassOptionsToPackage{square, numbers, compress}{natbib}
\usepackage[final]{neurips_2023}

\usepackage[utf8]{inputenc} % allow utf-8 input
\usepackage[T1]{fontenc}    % use 8-bit T1 fonts
\usepackage{hyperref}       % hyperlinks
\usepackage{url}            % simple URL typesetting
\usepackage{booktabs}       % professional-quality tables
\usepackage{amsfonts}       % blackboard math symbols
\usepackage{nicefrac}       % compact symbols for 1/2, etc.
\usepackage{microtype}      % microtypography
\usepackage{xcolor}         % colors

\usepackage{enumitem}
\setlist[itemize]{leftmargin=*}

\title{On Measuring Fairness in Generative Models}

\usepackage{graphicx}
\usepackage{multirow}
\usepackage{arydshln}
\usepackage{colortbl}
\usepackage{placeins}

\usepackage[linesnumbered,lined,ruled,commentsnumbered]{algorithm2e}

\SetKwComment{tcc}{$\triangleright$~}{}
\SetCommentSty{normalfont}
\SetKwInput{KwRequire}{Require}

\usepackage{subcaption}
\usepackage{amsmath,amsfonts,bm}

\def\eqref#1{equation~\ref{#1}}
\def\1{\bm{1}}

\DeclareMathAlphabet{\mathsfit}{\encodingdefault}{\sfdefault}{m}{sl}
\SetMathAlphabet{\mathsfit}{bold}{\encodingdefault}{\sfdefault}{bx}{n}

\DeclareMathOperator*{\argmax}{arg\,max}

\newcommand{\wrt}{\textit{w.r.t.} }
\newcommand{\ie}{\textit{i.e. }}
\newcommand{\eg}{\textit{e.g. }}
\newcommand{\etal}{\textit{et al. }}
\newcommand{\bu}{\mathbf{u}}
\newcommand{\bx}{\mathbf{x}}
\newcommand{\bbE}{\mathbb{E}}

\newcommand{\phat}{\hat{p}}
\newcommand{\bphat}{\bm{\hat{p}}}
\newcommand{\pstar}{p^*}
\newcommand{\bpstar}{\bm{p^*}}

\newcommand{\bpbar}{\bm{\bar{p}}}
\def\z{{\mathbf z}}

\def\balpha{{\boldsymbol \alpha}}

\def\bc{{\mathbf c}}
\def\bM{{\mathbf M}}
\def\bp{{\mathbf p}}

\def\CLEAMP{{\mu_{\texttt{CLEAM}}(\pstar_0)}}

\def\mubase{\mu_{\texttt{Base}}}

\def\mucleam{\mu_{\texttt{CLEAM}}}

\def\rhobase{{\rho_{\texttt{Base}}}}

\def\rhocleam{{\rho_{\texttt{CLEAM}}}}

\def\emubase{e_{\mu_{\texttt{Base}}}}
\def\emudiv{e_{\mu_{\texttt{DiV}}}}
\def\emucleam{e_{\mu_{\texttt{CLEAM}}}}
\def\emubbse{e_{\mu_{\texttt{BBSE}}}}

\newcommand{\st}{\textit{s.t. }}

\author{
   Christopher T. H. Teo \\
   \texttt{christopher\_teo@mymail.sutd.edu.sg}
   \And
    Milad Abdollahzadeh \\
   \texttt{milad\_abdollahzadeh@sutd.sg}%Address \\
   \And
    Ngai-Man Cheung\thanks{Corresponding Author} \\
    \texttt{ngaiman\_cheung@sutd.edu.sg}\and
   \\
   Singapore University of Technology  and Design (SUTD) \\
}

\begin{document}

\maketitle

\begin{abstract}
Recently, there has been increased interest in fair generative models.
In this work, we conduct, for the first time, an in-depth study on \textbf{fairness measurement}, a critical component 
in gauging
progress on fair generative models.
We make 
three  contributions. 
First, we conduct a study that reveals that  
the existing 
fairness measurement
framework has considerable measurement errors, even when  
highly accurate sensitive attribute (SA) classifiers are used.
These findings cast doubts on previously reported fairness improvements.
Second, to address this issue, we propose 
CLassifier Error-Aware Measurement (CLEAM), a new framework which uses a statistical model to account for inaccuracies in SA classifiers.
Our proposed CLEAM 
reduces measurement errors significantly, e.g., 
{\bf 4.98\%$\rightarrow$0.62\%} for StyleGAN2 
\wrt \texttt{Gender}.
Additionally,
CLEAM achieves this with minimal additional overhead.
Third, we utilize CLEAM to measure fairness in important text-to-image generator and 
GANs, revealing considerable biases in these models 
that raise concerns about their applications.
\textbf{Code and more resources:} \url{https://sutd-visual-computing-group.github.io/CLEAM/}.
\end{abstract}

%%%%%%%%%%%%%%%%%%%%%%%%%%%%%%%%%%%%%%%%%%%%%%%%%%%%%%%%%%%%%%%%%%
%             Section: Introduction
%%%%%%%%%%%%%%%%%%%%%%%%%%%%%%%%%%%%%%%%%%%%%%%%%%%%%%%%%%%%%%%%%%

\section{Introduction}

\begin{figure*}[t]
%\vspace{-0.05cm}
\begin{center}
%trim=left bottom right top
\includegraphics[width=\textwidth]{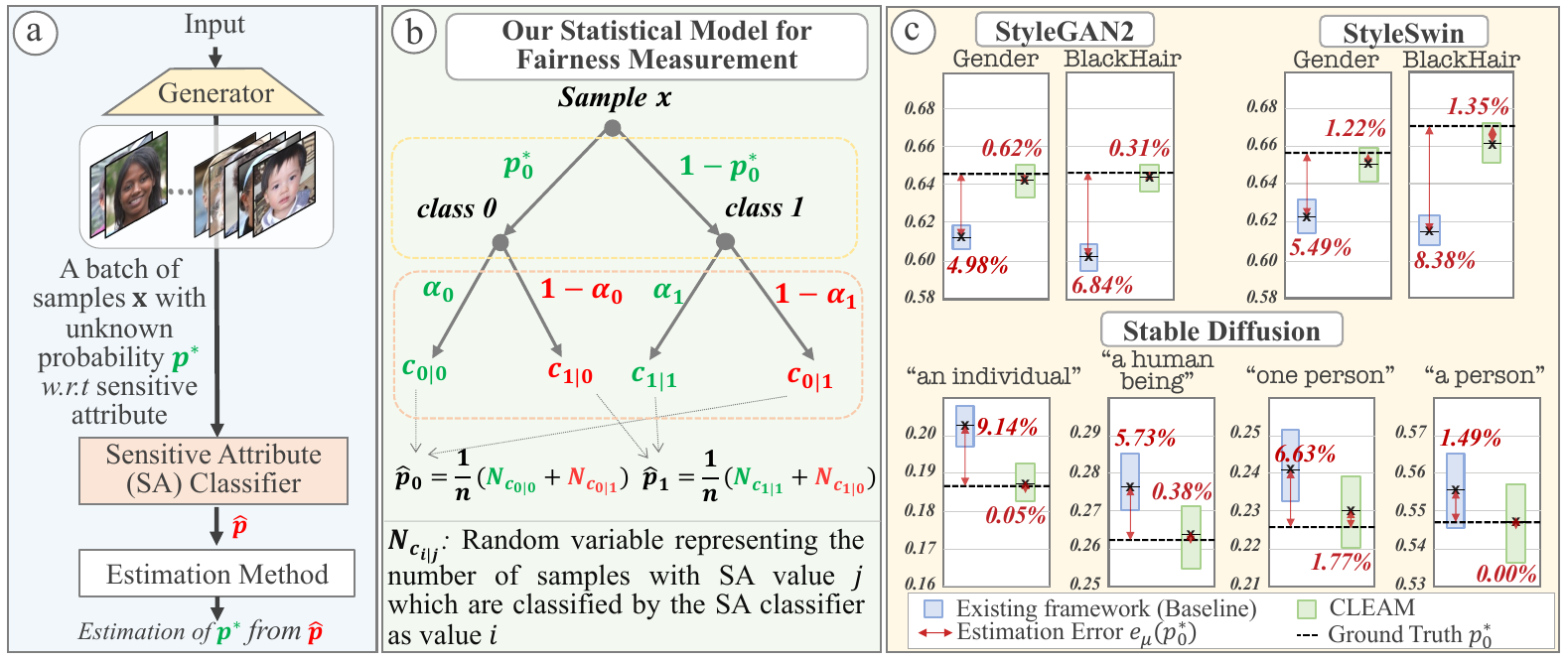}
\caption{
\textcircled{a} 
{\bf General framework for measuring fairness in generative models.}
Generated samples
 with unknown ground-truth (GT) probability $\bpstar$ \wrt sensitive attribute (SA)
are fed into a SA 
classifier to obtain $\bphat$.
Existing framework (Baseline)
uses the classifier output $\bphat$ as  estimation of $\bpstar$. In contrast, our proposed CLEAM includes an improved estimation that accounts for inaccuracies in the SA classifier (see Alg. \ref{alg:cleam}).
\textcircled{\raisebox{-.9pt}b} 
{\bf Our statistical model for fairness measurement.} This model accounts for inaccuracies in the SA classifier and is the base of our proposed CLEAM (see Sec. \ref{sub:Randon_FD_Framework}).
\textcircled{c} 
{\bf Improvements with CLEAM.} 
CLEAM improves upon 
Baseline \cite{choiFairGenerativeModeling2020a,teo2022fair} by reducing the relative error in estimating the GT $\pstar_0$ for SOTA GANs: StyleGAN2 \cite{karrasStyleBasedGeneratorArchitecture2019} and StyleSwin \cite{zhang2021styleswin}, and Stable Diffusion Model \cite{rombach2021highresolution}.
First row displays the Baseline and CLEAM estimates for each GAN, using ResNet-18 as the SA classifier for \texttt{Gender} and \texttt{BlackHair}. The Baseline incurs significant fairness measurement errors (\eg 4.98\%), even when utilizing a highly accurate ResNet-18 ($\approx$97\% accuracy). Meanwhile, CLEAM reduces the error significantly in all setups, \eg in the first panel, the error is reduced: 4.98\% $\rightarrow$ 0.62\%.
Similarly, in the second row, CLEAM reduces measurement error significantly in the Stable Diffusion Model \cite{rombach2021highresolution}, using CLIP \cite{radfordLearningTransferableVisual2021} as the SA classifier for \texttt{Gender}, \eg first panel: 9.14\% $\rightarrow$ 0.05\% 
(Detailed evaluation in Tab. \ref{tab:G_PE} and Tab. \ref{tab:G_PE2}). 
{\bf
Best viewed in color.}
}
\label{fig:overall_framework}
\end{center}
\vspace{-18pt}
\end{figure*}

Fair generative models have been attracting significant attention recently~\cite{choiFairGenerativeModeling2020a,teo2022fair,frankelFairGenerationPrior2020,humayunMaGNETUniformSampling2021a,tanImprovingFairnessDeep2020,yuInclusiveGANImproving2020a,malulekeStudyingBiasGANs2022a,umFairGenerativeModel2021,abdollahzadeh2023survey}.
In generative models~\cite{goodfellow2020generative, karras2020training, zhao2022fewshot,chandrasegaran2022discovering, zhao2023exploring}, 
fairness is commonly defined as 
equal generative quality \cite{malulekeStudyingBiasGANs2022a} 
or
equal representation \cite{choiFairGenerativeModeling2020a,teo2022fair,frankelFairGenerationPrior2020,tanImprovingFairnessDeep2020,umFairGenerativeModel2021,hutchinson50YearsTest2019,teoMeasuringFairnessGenerative2021} \wrt some \textit{Sensitive Attributes} (SA).
In this work, we focus on the more widely utilized definition -- \textit{equal representation.} 
In this definition, as an example,
a generative model 
is regarded as fair 
\wrt \texttt{Gender}, if it generates \texttt{Male} and \texttt{Female} samples with equal probability.
This is an important research topic as
such
biases in
generative models
could impact their application efficacy, e.g., by introducing 
racial bias in  face generation of suspects \cite{jalanSuspectFaceGeneration2020} or reducing accuracy when supplementing data for disease diagnosis~\cite{frid-neurocomputing-2018gan}.

{\bf Fairness measurement for generative models.}
Recognizing the importance of fair generative models, several methods have been proposed
to mitigate biases in generative models
\cite{choiFairGenerativeModeling2020a,teo2022fair,frankelFairGenerationPrior2020,tanImprovingFairnessDeep2020,umFairGenerativeModel2021}.
However, {\em in our work, we focus mainly on the accurate fairness measurement of deep generative models \ie assessing and quantifying the bias of generative models.} This is a critical topic, as accurate measurements are essential to reliably gauge the progress of bias mitigation techniques.
The general fairness measurement framework  
is shown in Fig. \ref{fig:overall_framework} (See 
 Sec. \ref{sec:problemsetup} for details).
This framework
is utilized in existing works
to assess their proposed fair generators.  
Central to the fairness measurement framework is a {\em SA classifier}, which classifies the generated samples \wrt a SA, in order to estimate the bias of the generator.
For example, if eight out of ten generated face images are classified as \texttt{Male}
by the SA classifier, then the generator is deemed biased at $0.8$ towards \texttt{Male} 
(further discussion in Sec. \ref{sec:problemsetup}).
We follow previous works \cite{choiFairGenerativeModeling2020a, teo2022fair, umFairGenerativeModel2021} and focus on binary SA due to dataset limitations.

{\bf Research gap.}
In this paper, we study a critical research gap in fairness measurement.
Existing works assume that
when 
SA classifiers are highly accurate,
measurement errors 
should be
insignificant. 
As a result,
the effect of 
errors in SA classifiers
has not been studied.
However, our study reveals that {\em even with highly accurate SA classifiers, %can result in 
considerable 
fairness
measurement errors could still occur}.
This finding raises concerns about
potential errors in previous works' results, which are
measured using existing  
framework.
Note that the SA classifier is {\em indispensable} in fairness measurement as it enables automated measurement 
of generated samples.

{\bf Our contributions.} We make 
three 
contributions to fairness measurement for generative models.
 {\em As our first contribution}, 
we analyze the 
accuracy of
fairness measurement
on generated samples, which previous works \cite{choiFairGenerativeModeling2020a,teo2022fair,frankelFairGenerationPrior2020,tanImprovingFairnessDeep2020,umFairGenerativeModel2021} have been unable to carry out due to the 
unavailability
of proper
datasets.
We overcome this 
challenge by
proposing new datasets of {\em generated samples} with 
manual labeling \wrt 
various
SAs.
The datasets include 
 generated samples from
Stable Diffusion Model (SDM) \cite{rombach2021highresolution} ---a popular text-to-image generator--- as well as
two State-of-The-Art (SOTA) GANs (StyleGAN2 \cite{karrasStyleBasedGeneratorArchitecture2019} and StyleSwin \cite{zhang2021styleswin}) \wrt different SAs.
Our new datasets are then utilized in our work to evaluate the accuracy of the existing fairness measurement framework.
Our results reveal that the accuracy of the existing fairness measurement framework is not adequate, due to the lack of consideration for the SA classifier inaccuracies.
More importantly, we found that 
{\em even in setups where the accuracy of the SA classifier is high, the error in fairness measurement could still be significant}.
Our finding raises concerns about the accuracy of previous works' results \cite{choiFairGenerativeModeling2020a,teo2022fair,umFairGenerativeModel2021}, especially since some of their reported improvements 
are smaller than the margin of measurement errors
that 
we observe in our study
when evaluated under the same setup;
further discussion in Sec. \ref{subsec:empiricalStudy}.

%%%%%%%%%%%%%%%%%%%%
%Second Contribution
%%%%%%%%%%%%%%%%%%%%
To address this issue, 
{\em as our second (major) contribution}, 
we propose CLassifier Error-Aware Measurement (CLEAM), a new more accurate fairness measurement framework
based on our developed statistical model for SA classification (further details on the statistical model in Sec. \ref{sub:Randon_FD_Framework}).
Specifically, CLEAM utilizes this statistical model to account for the classifier's inaccuracies during SA classification and outputs a more accurate fairness measurement.
We then evaluate the accuracy of CLEAM and validate its improvement over existing fairness measurement framework.
We further 
conduct a series of different ablation studies to validate performance of CLEAM.
We remark that CLEAM is not a new fairness metric,
but
an improved 
fairness measurement framework that could achieve better accuracy in bias estimation 
when used with various fairness metrics for generative models.

%%%%%%%%%%%%%%%%%%%%
%Third Contribution
%%%%%%%%%%%%%%%%%%%%
{\em As our third contribution}, 
we apply CLEAM
as an accurate framework to reliably measure biases in popular generative models. 
Our study reveals that  SOTA GANs have
considerable biases \wrt several SA.
Furthermore, we observe an intriguing property in 
Stable Diffusion Model: 
slight differences in semantically similar prompts could result in markedly different biases for SDM.
These results prompt careful consideration on 
the implication of biases in generative models.
{\bf Our contributions are:
}
\begin{itemize}
    \vspace{-3pt}
    \item We conduct a study to reveal that even highly-accurate SA classifiers 
    could still incur significant 
    fairness measurement 
    errors when using existing framework.
    \item To enable 
    evaluation of fairness measurement frameworks, we propose new datasets based on generated samples from StyleGAN, StyleSwin and SDM, with manual labeling \wrt SA.
    \item We propose a 
    statistically driven
    fairness measurement framework, CLEAM, 
    which accounts for the SA classifier inaccuracies to output 
    a more accurate  
    bias estimate.
    \item Using CLEAM, we reveal considerable biases in several important generative models, prompting careful consideration when applying them to different applications.

\end{itemize}
%}
\vspace{-6pt}

%%%%%%%%%%%%%%%%%%%%%%%%%%%%%%%%%%%%%%%%%%%%%%%%%%%%%%%%%%%%%%%%%%%
%%                          Problem Setup
%%%%%%%%%%%%%%%%%%%%%%%%%%%%%%%%%%%%%%%%%%%%%%%%%%%%%%%%%%%%%%%%%%%
\section{Fairness Measurement Framework}
\label{sec:problemsetup}

Fig.\ref{fig:overall_framework}(a) illustrates the fairness measurement framework for generative models
as in \cite{choiFairGenerativeModeling2020a,teo2022fair,frankelFairGenerationPrior2020,tanImprovingFairnessDeep2020,umFairGenerativeModel2021}. 
Assume that with some input \eg noise vector for a GAN or text prompt for SDM,
a generative model 
synthesizes a sample $\bx$. 
Generally, as the generator does not label synthesized samples, the ground truth (GT) class probability of these samples \wrt a SA (denoted by $\bpstar$) is unknown.  
Thus, an SA classifier $C_\bu$ is utilized to estimate $\bpstar$. 
Specifically,
for each sample 
$\bx \in \{\bx\}$ ,
$C_\bu(\bx)$ is the argmax classification for the respective SA.
In existing works, the expected value of the SA classifier output over a batch of samples,
$\bphat=\bbE_{\bx}[C_\bu(\bx)]$
(or the average of $\bphat$ over multiple batches of samples), is used as an estimation of $\bpstar$.
This estimate may then be used in some fairness metric $f$ to report the fairness value for the generator, \eg fairness discrepancy metric between $\bphat$ and a uniform distribution $\bpbar$ \cite{choiFairGenerativeModeling2020a,teoMeasuringFairnessGenerative2021}
(see Supp A.3 for
details on how to calculate $f$).
Note that {\em the general assumption behind the existing framework is that with a reasonably accurate SA classifier, {$\bphat$} could be an accurate estimation of {{$\bpstar$}}} \cite{choiFairGenerativeModeling2020a,tanImprovingFairnessDeep2020}. 
In the next section, we will present a deeper analysis 
on the effects of an inaccurate SA classifier
on fairness measurement.
Our findings suggest
that there could be a large discrepancy between $\bphat$ and $\bpstar$, even for highly accurate SA classifiers,
indicative of significant fairness measurement errors in the current measurement framework.

One may argue that conditional GANs (cGANs)
\cite{mirza2014conditional,odena2017acgan} may be used to generate samples 
conditioned on the SA, thereby eliminating the need for an SA classifier.
However, cGANs are not considered in previous works 
due to several limitations. These include the limited availability of  
large {\em labeled} training datasets, 
the unreliability of 
sample quality and labels \cite{thekumparampilRobustnessConditionalGANs2018}, and the exponentially increasing conditional terms, per SA. 
Similarly, for SDM, Bianchi \etal \cite{bianchiEasilyAccessibleTexttoImage2022} found that utilizing well-crafted prompts to mitigate biases is ineffective due to the presence of existing biases in its 
training dataset.
Furthermore in Sec. \ref{sec:application}, utilizing CLEAM, we will discuss that even subtle prompt changes (while maintaining the semantics) result in drastically different SA biases.
See Supp G for further comparison between \cite{bianchiEasilyAccessibleTexttoImage2022} and our findings.

%%%%%%%%%%%%%%%%%%%%%%%%%%%%%%%%%%%%%%%%%%%%%
%%  A closer look at fairness Measurement
%%%%%%%%%%%%%%%%%%%%%%%%%%%%%%%%%%%%%%%%%%%%%
\section{A Closer Look at Fairness Measurement}
\label{subsec:empiricalStudy}

In this section, we take a closer look at the existing fairness measurement framework.
In particular, we examine its performance in estimating $\bpstar$ of the samples generated by SOTA GANs
and SDM,
a task previously unstudied due to the lack of 
a labeled generated dataset.
We do so by
designing an experiment to demonstrate these errors while evaluating biases in %\textcolor{blue}{
popular image generators.
Following previous works,
our main focus is on binary SA which takes values in $\{0,1\}$.  
Note that, we assume that the accuracy of the SA classifier $C_\bu$ is known and is characterized by 
$\balpha=\{\alpha_0,\alpha_1\}$, where $\alpha_i$ is the probability of correctly classifying label $i$. For example, for \texttt{Gender} attribute, $\alpha_0$ and $\alpha_1$ are the probability of correctly classifying
\texttt{Female}, and \texttt{Male} classes, respectively.
In practice, 
$C_\bu$ is trained on standard training procedures (more details in the Supp F) and
$\balpha$ can be measured during the validation stage of $C_\bu$ and be considered a constant when the validation dataset is large enough.
Additionally, $\bpstar$ can be  assumed to be a constant vector, given that the samples generated 
can be considered to come from an infinite population, as theoretically there is no limit to the number of samples 
from a generative model like GAN or SDM.

{\bf New dataset by labeling 
generators
output.} 
The
major limitation of evaluating the existing fairness measurement framework is 
the unavailability of $\bpstar$.
{\em To pave the way for an accurate evaluation, we create a new dataset by manually labeling the 
samples generated by GANs and SDM}.
More specifically, we utilize 
the official publicly released pre-trained
StyleGAN2 \cite{karrasStyleBasedGeneratorArchitecture2019} and StyleSwin \cite{zhang2021styleswin} 
on CelebA-HQ \cite{CelebAMask-HQ} for sample generation.
Then, we 
randomly sample from these GANs and
utilize Amazon Mechanical Turks to hand-label the samples \wrt \texttt{Gender} and \texttt{BlackHair}, resulting in $\approx$9K samples for each GAN;
see 
Supp 
H
for more details and examples.
Next, we follow a similar labeling process \wrt \texttt{Gender}, but with a 
SDM \cite{rombach2021highresolution} pre-trained on LAION-5B\cite{schuhmann2022laion}.
Here, we input prompts using best practices \cite{bianchiEasilyAccessibleTexttoImage2022,liu2022design,StableDiffusionPrompt2022,zhouLearningPromptVisionLanguage2022}, %that began
beginning
with a scene description ("A photo with the face of"), followed by
four indefinite (gender-neutral) pronouns or nouns \cite{haspelmathIndefinitePronouns1997,saguy2022little} -- \{"an individual", "a human being", "one person", "a person"\}
to collect $\approx$2k high-quality samples.
We refer to this new dataset as Generated Dataset ({\bf GenData}), which includes generated images from three models with corresponding SA labels: GenData-StyleGAN2, GenData-StyleSwin, GenData-SDM.
We remark that 
these labeled datasets only provide a strong approximation of $\bpstar$ for each generator, however as the datasets are reasonably large, we find this approximation sufficient and simply refer to it as the GT $\bpstar$.
Then utilizing this GT $\bpstar$, we compare it against the estimated baseline ($\bphat$).
One interesting observation revealed by 
GenData
is that all three generators exhibit a considerable amount of bias (see Tab.\ref{tab:G_PE} and \ref{tab:G_PE2}); more detail in Sec. \ref{sec:application}.
Note that for a fair %GAN 
generator
we have $\pstar_0=\pstar_1=0.5$, and measuring the $\pstar_0$ and $\pstar_1$ is a good proxy for measuring fairness.

{\bf Experimental setup.} Here, we follow Choi \etal \cite{choiFairGenerativeModeling2020a} as the {\em Baseline} for measuring fairness.
In particular, to calculate each $\bphat$ value for a generator, a corresponding batch of $n=400$ samples is randomly drawn from 
GenData
and passed into $C_\bu$ for SA classification. 
We repeat this for $s=30$ batches and report the mean results denoted by $\mubase$ %$\mu_{\texttt{Base}}$
and the 95\% confidence interval denoted by $\rhobase$. %$\rho_{\texttt{Base}}$. 
For a comprehensive analysis of the GANs, 
we repeat the experiment
using four different SA classifiers: Resnet-18, ResNet-34 \cite{heDeepResidualLearning2016}, MobileNetv2 \cite{sandlerMobileNetV2InvertedResiduals2018}, and VGG-16 \cite{simonyanVeryDeepConvolutional2014}.
Then, to evaluate the SDM, we utilize CLIP \cite{radfordLearningTransferableVisual2021} to explore the utilization of pre-trained models for zero-shot SA classification; 
more details on the CLIP SA classifier in Supp. E.
As CLIP 
does not have a validation dataset, to measure $\balpha$ for CLIP, we utilize CelebA-HQ, a dataset with a similar domain to our application. We found this to be a very accurate approximation;
see Supp D.7 for validation results.
Note that for SDM, a separate $\bphat$ is measured for each text prompt as SDM's output images are conditioned on the input text prompt.
As seen in Tab. \ref{tab:G_PE} and \ref{tab:G_PE2},  all classifiers demonstrate reasonably high 
average accuracy $\in[84\%,98.7\%]$. %$\in[84\%,96\%]$.
Note that as we focus on binary SA (\eg \texttt{Gender:\{Male, Female\}}), both $\bpstar$ and $\bphat$
have two components
\ie $\bpstar=\{ \pstar_0, \pstar_1\}$, and $\bphat=\{ \phat_0, \phat_1\}$.
After computing the $\mubase$ and $\rhobase$, we calculate {\em normalized $L_1$ point error} $e_\mu$, and {\em interval max error} $e_\rho$ \wrt the $\pstar_0$ (GT) to evaluate the measurement accuracy of the baseline method:
\vspace{-2pt}
\begin{equation}
    \begin{aligned}
       e_{\mu_{\texttt{Base}}}
= \tfrac{1}{\pstar_0} |\pstar_0 - \mu_{\texttt{Base}}| 
\quad; \quad
e_{\rho_{\texttt{Base}}}
        =\tfrac{1}{\pstar_0}\max\{|\min(\rho_{\texttt{Base}})-\pstar_0|,|\max(\rho_{\texttt{Base}})-\pstar_0|\}
    \label{eqn:pointError}
    \end{aligned}
\end{equation}

{\bf Based on our results in Tab. \ref{tab:G_PE}},
for GANs, we observe that despite the use of reasonably accurate SA classifiers, there are significant estimation errors in the existing fairness measurement framework, \ie
$\emubase$$\in [4.98\%,17.13\%]$.
In particular, looking at the SA classifier with the highest average accuracy of $\approx97\%$ (ResNet-18 on \texttt{Gender}), we observe 
significant discrepancies between GT $\pstar_0$ and $\mubase$, with $\emubase=4.98\%$. 
These errors generally worsen as accuracy marginally degrades, \eg MobileNetv2 with accuracy $\approx 96\%$ results in $\emubase=5.45\%$.
These considerably large errors contradict prior assumptions -- that for a reasonably accurate SA classifier, we can assume $\emubase$ to be fairly negligible. 
Similarly, our results in Tab. \ref{tab:G_PE2} for the SDM, show large $\emubase$$\in [1.49\%,9.14\%]$, even though the classifier is very accurate. 
We discuss the reason for this in more detail in Sec. \ref{sec:evalrealG}.

{\em Overall, these results are concerning as they cast doubt on the accuracy of prior reported results.} For example, imp-weighting \cite{choiFairGenerativeModeling2020a} which uses the same ResNet-18 source code as our experiment, reports a 2.35\% relative improvement in fairness against its baseline \wrt \texttt{Gender}, which falls within the range of our experiments smallest relative error, $\emubase$=4.98\%. Similarly, Teo \etal  \cite{teo2022fair} and Um \etal \cite{umFairGenerativeModel2021} report a relative improvement in fairness of 0.32\% and 0.75\%, compared to imp-weighting \cite{choiFairGenerativeModeling2020a}. These findings suggest that some prior results may be affected due to oversight of SA classifier's  
inaccuracies;
see Supp. A.4 for more details on how to calculate these measurements.

{\bf Remark:} In this section, we provide the keystone for the evaluation of measurement accuracy in the current framework by introducing 
a labeled dataset based on generated samples.
These evaluation results raise concerns about the accuracy of existing framework as considerable error rates were observed even when using  accurate SA classifiers, an issue previously seen to be negligible.

%%%%%%%%%%%%%%%%%%%%%%%%%%%%%%%%%%%%%%%%%%%%%%%%%%%%%%%%%%%%%%%%%%%%%%%%%%%%%%%%%%%%%%%%%%
%
%%  Proposed method:  Mitigating Error in Fairness Measurements
%
%%%%%%%%%%%%%%%%%%%%%%%%%%%%%%%%%%%%%%%%%%%%%%%%%%%%%%%%%%%%%%%%%%%%%%%%%%%%%%%%%%%%%%%%%%
\section{Mitigating Error in Fairness Measurements}

The previous section exposes the inaccuracies in the existing fairness measurement framework. Following that, in this section, we 
first
develop a statistical model for the
erroneous output of the
SA classifier, 
$\bphat$, to help draw a more systematic relationship between the inaccuracy of the  SA classifier 
and error in fairness estimation.
Then,
with this statistical model, we propose CLEAM -- a %measurement method 
new measurement framework
that reduces error in 
the measured $\bphat$
by  
accounting
for the SA classifier inaccuracies 
to output a more accurate statistical approximation of $\bpstar$.

\subsection{Proposed Statistical Model for Fairness Measurements}
\label{sub:Randon_FD_Framework}

As shown in Fig.\ref{fig:overall_framework}(a), to measure the fairness of
the generator, %$G_\theta$
we feed $n$ 
generated samples
to the SA classifier $C_\bu$. The output of the SA classifier ($\bphat$) is in fact a random variable that aims to approximate the $\bpstar$.
Here, we propose a statistical model to derive the distribution of $\bphat$.

As Fig.\ref{fig:overall_framework}(b) demonstrates in our running example of a binary SA, each generated sample is from {\em class 0} with probability $\pstar_0$, or from {\em class 1}  with probability $\pstar_1$.
Then, generated sample from {\em class $i$} where $i\in\{0,1\}$,  will be classified correctly with the probability of $\alpha_i$, and wrongly with the probability of $\alpha'_{i}=1-\alpha_i$.  
Thus, for each sample, there are four mutually exclusive possible 
events %outputs 
denoted by $\bc$, with the corresponding probability vector $\bp$:
\begin{equation}
    \begin{aligned}
        \bc^T =
        \begin{bmatrix}
            c_{0|0} \quad
            c_{1|0} \quad
            c_{1|1} \quad
            c_{0|1}
        \end{bmatrix}
        \quad , \quad
            \bp^T=
        \begin{bmatrix}
            \pstar_0\alpha_0  \quad
            \pstar_0\alpha_0' \quad
            \pstar_1\alpha_1  \quad
            \pstar_1\alpha_1'
            %\label{eqn:defining_vector_p}
        \end{bmatrix}
        \label{eqn:defining_vector_cp}
    \end{aligned}
\end{equation}       

%Space saving strategy
\iffalse
\begin{align}
    \bc^T &=
    \begin{bmatrix}
        c_{0|0} \quad
        c_{1|0} \quad
        c_{1|1} \quad
        c_{0|1}
        \label{eqn:defining_vector_c}
    \end{bmatrix} \\
\bp^T&=
    \begin{bmatrix}
        \pstar_0\alpha_0  \quad
        \pstar_0\alpha_0' \quad
        \pstar_1\alpha_1  \quad
        \pstar_1\alpha_1'
        \label{eqn:defining_vector_p}
    \end{bmatrix}
\end{align}
\fi

where $c_{i|j}$ denotes the event of assigning label $i$ to a sample with GT label $j$. 
Given that this process is performed independently for each of the $n$ generated images, the 
probability of the counts 
for each output $\bc^T$ in Eqn. \ref{eqn:defining_vector_cp} (denoted by $\mathbf{N}_\bc$) can be modeled by a multinomial distribution, \ie $\mathbf{N}_\bc \sim Multi(n,\bp)$ \cite{rao1957maximum, kesten1959property,papoulisProbabilityRandomVariables2002}.
Note that $\mathbf{N}_\bc$ models the {\em joint probability distribution} of these outputs, \ie $ \mathbf{N}_\bc \sim \mathbb{P}(N_{c_{0|0}},N_{c_{1|0}},N_{c_{1|1}},N_{c_{0|1}})$
where, $N_{c_{i|j}}$ is the random variable of the count for event $c_{i|j}$ after classifying $n$ generated images.
Since $\bp$ is not near the boundary of the parameter space, 
and as we utilize a large $n$, 
based on the central limit theorem,
$Multi(n,\bp)$ can be approximated by a multivariate Gaussian distribution, 
$\mathbf{N}_\bc \sim \boldsymbol{\mathcal{N}}(\boldsymbol{\mu}, \boldsymbol{\Sigma})$, with $\boldsymbol{\mu}=n\bp$ and $\boldsymbol{\Sigma}= n\bM$ \cite{geyerStat5101Notes2010,papoulisProbabilityRandomVariables2002}, where $\bM$ is defined as:
\begin{equation}
    \begin{aligned}
    \bM=diag(\bp) -\bp\bp^T
    \end{aligned}
\end{equation}
$diag(\bp)$ denotes a square diagonal matrix corresponding to vector $\bp$ (see 
Supp 
A.1
for expanded form).
The {\em marginal distribution} of this multivariate Gaussian distribution gives us a univariate (one-dimensional) Gaussian distribution for the count of each output $\bc^T$ in Eqn. \ref{eqn:defining_vector_cp}. 
For example, the distribution of the count for event $c_{0|0}$, denoted by $N_{c_{0|0}}$, can be modeled as $N_{c_{0|0}} \sim \mathcal{N}(\boldsymbol{\mu}_1, \boldsymbol{\Sigma}_{11})$.

Lastly, we find the total percentage %rate 
of data points labeled as class $i$ when labeling $n$ generated images using the normalized sum of the related random variables, \ie $\hat{p}_i=\frac{1}{n}\sum_j N_{{c}_{i|j}}$.
For our binary example, 
$\hat{p}_i$ can be calculated by summing  random variables with Gaussian distribution, which results in another Gaussian distribution~\cite{goodman1963statistical}, \ie, $\hat{p}_0 \sim \mathcal{N}(\Tilde{\mu}_{\phat_0},\Tilde{\sigma}^2_{\phat_0})$, where:
\begin{align}
    \begin{split}
        \Tilde{\mu}_{\phat_0} = & \tfrac{1}{n} (\boldsymbol{\mu}_1 + \boldsymbol{\mu}_4) = \pstar_0\alpha_0+\pstar_1\alpha_1'
        \label{eqn:meanp0}
    \end{split}\\
    \begin{split}
            \Tilde{\sigma}^2_{\phat_0} 
              = & \tfrac{1}{n^2}(\boldsymbol{\Sigma}_{11} + \boldsymbol{\Sigma}_{44} + 2\boldsymbol{\Sigma}_{14})
               = \tfrac{1}{n}[(\pstar_0\alpha_0-(\pstar_0\alpha_0)^2)+(\pstar_1\alpha_1'-(\pstar_1\alpha_1')^2)]  + \tfrac{2}{n} {\pstar_0\pstar_1\alpha_0\alpha_1'}
              \label{eqn:varp0}
    \end{split}
\end{align}

Similarly $\hat{p}_1 \sim \mathcal{N}(\Tilde{\mu}_{\phat_1},\Tilde{\sigma}^2_{\phat_1})$ with $\Tilde{\mu}_{\phat_1} = (\boldsymbol{\mu}_2 + \boldsymbol{\mu}_3)/n$, and $\Tilde{\sigma}^2_{\phat_1} =(\boldsymbol{\Sigma}_{22} + \boldsymbol{\Sigma}_{33} + 2\boldsymbol{\Sigma}_{23})/n^2$
which is aligned with the fact that $\hat{p}_1 = 1 - \hat{p}_0$.

\vspace{-1mm}

{\bf Remark:} In this section, considering the probability tree diagram in Fig.\ref{fig:overall_framework}(b), we propose a joint distribution for the possible events of classification 
($N_{c_{i|j}}$), 
and use it to compute the marginal distribution of each event, and finally the distribution of the SA classifier outputs ($\phat_0$, and $\phat_1$). 
Note that considering Eqn.~\ref{eqn:meanp0},~\ref{eqn:varp0}, only with a perfect classifier ($\alpha_i=1$,  \ie acc$=100\%$) the $\Tilde{\mu}_{\phat_0}$ converges to 
$\pstar_0$. 
However, training a perfect SA classifier is not practical \eg due to the lack of an appropriate dataset and task hardness~\cite{abdollahzadeh2021revisit, achille2019task2vec}.
As a result, in the following, we will propose CLEAM which instead utilizes this statistical model to mitigate the error of the SA classifier.

\subsection{CLEAM for Accurate Fairness Measurement}
\label{sub:approxMLEpstar}

In this section, we propose a new estimation method in fairness measurement  that considers the inaccuracy of the SA classifier. 
For this, we use the statistical model, introduced in Sec \ref{sub:Randon_FD_Framework}, to compute a more accurate estimation of $\bpstar$. 
Specifically, we first propose a Point Estimate (PE) by approximating the {\it maximum likelihood value} of $\bpstar$. Then, we use the {\it confidence interval} for the 
observed data
($\bphat$)
to propose an Interval Estimate (IE) for $\bpstar$.

\textbf{Point Estimate (PE) for} {$\bpstar$}. Suppose that we have access to $s$ samples of $\bphat$ denoted by $ \{\bphat^1,\dots,\bphat^s\}$, \ie SA classification results on $s$ batches of generated data.
%Recall that each sample $\bphat^i$ contains the prediction \wrt all classes of attribute, \eg  $\bphat^i=\{\phat^i_0, \phat^i_1\}$.
We can then use the proposed statistical model to approximate the $\bpstar$.
In the previous section, we demonstrate that we can model $\phat_{j}^{i}$ using a Gaussian distribution.
Considering this, first, we use the available samples to calculate sample-based statistics including the mean and variance of the $\phat_j$ samples:

% space-saving strategy
\vspace{-0.5cm}
\begin{align}
    \begin{split}
        \Ddot{\mu}_{\phat_j} = &\textstyle\frac{1}{s} \textstyle\sum_{i=1}^s \phat^i_j
        \label{eqn:meanp0samples}
    \end{split}
    \\
    \begin{split}
       \Ddot{\sigma}^2_{\phat_j}=&{\textstyle\frac{1}{s-1} \textstyle\sum^s_{i=1}(\phat^i_j-\Ddot{\mu}_{\phat_j})^2}
        \label{eqn:varp0samples}
    \end{split}
\end{align}

For a Gaussian distribution, the Maximum Likelihood Estimate (MLE) of the population mean is its sample mean $\Ddot{\mu}_{\phat_j}$~\cite{anderson1985maximum}.
Given that $s$ is large enough (\eg $s>30$), we can assume that 
$\Ddot{\mu}_{\phat_j}$ %the MLE 
is a good approximation of the population mean~\cite{krohling2006coevolutionary}, and equate it to the statistical population mean $\Tilde{\mu}_{\phat_j}$ in Eqn. \ref{eqn:meanp0} (see %Supp\ref{apx:proposedmethod}
Supp 
A.2
for derivation).
With that, we get the {\it maximum likelihood approximation of $\bpstar$, which we call the CLEAM's point estimate, $\mucleam$}:
\begin{equation}
    %\CLEAMP =\frac{\Ddot{\mu}_{\hat{p}_0}-\alpha'_{1}}{\alpha_0-\alpha'_{1}}
    \CLEAMP =({\Ddot{\mu}_{\hat{p}_0}-\alpha'_{1}})/({\alpha_0-\alpha'_{1}})
    \quad,\quad \mu_\texttt{CLEAM}(p^*_1)=1-\mu_\texttt{CLEAM}(p^*_0)
    \label{eqn:MLEpstar}
\end{equation}
Notice that $\mucleam$ accounts for the 
inaccuracy of the SA classifier.

%-------ALGORITHM-----------
\begin{figure}[t]
\vspace{-0.1cm}
\begin{algorithm}[H]
%\setstretch{0.3}
\DontPrintSemicolon
\SetAlgoLined
\KwRequire{accuracy of SA classifier, $\balpha$.}
		
Compute SA classifier output $\bphat: \{\bphat^1,\dots,\bphat^s\}$ for $s$ batches of generated data.\\

Compute sample mean $\Ddot{\mu}_{\phat}$ and sample variance $\Ddot{\sigma}^{2}_{\phat}$ using (\ref{eqn:meanp0samples}) and (\ref{eqn:varp0samples}).\\

Use (\ref{eqn:MLEpstar}) to compute point estimate $\mu_\texttt{CLEAM}$.\\

Use (\ref{eqn:CIpstar}) to compute interval estimate $\rho_\texttt{CLEAM}$.\\

\caption{Computing point and interval estimates using CLEAM.}
\label{alg:cleam}
\end{algorithm}
\vspace{-0.3cm}
\end{figure}

\textbf{Interval Estimate (IE) for $\bpstar$.}
In the previous part, we propose a PE for $\bpstar$ using the statistical model, and sample-based mean $\Ddot{\mu}_{\hat{p}_0}$.
However, as we 
use 
only $s$ samples of $\bphat$, $\Ddot{\mu}_{\hat{p}_0}$ may not capture the exact value of the population mean.
This adds some degree of inaccuracy into $\mucleam$. In fact, in our framework, $\Ddot{\mu}_{\hat{p}_0}$ equals $\Tilde{\mu}_{\hat{p}_0}$ when $s \to \infty$. 
However, increasing each unit of $s$ 
significantly
increases the computational complexity, as each $\bphat$ requires $n$ generated samples.
To address this, we recall that $\phat_0$ follows a Gaussian distribution and instead utilize frequentist statistics \cite{goodman1963statistical} to propose a 95\% confidence interval (CI) for  $\bpstar$.
To do this, first  we derive the CI
for $\Tilde{\mu}_{\phat_0}$: 
\begin{equation}
    %\begin{aligned}
    \Ddot{\mu}_{\hat{p}_0} -  1.96 \tfrac{\Ddot{\sigma}_{\phat_0}}{\sqrt{s}}
    \leq 
    \Tilde{\mu}_{\phat_0} 
    \leq 
    \Ddot{\mu}_{\hat{p}_0} +  1.96 \tfrac{\Ddot{\sigma}_{\phat_0}}{\sqrt{s}}
    %\end{aligned}
    \label{eqn:CIbasic}
\end{equation}
Then, applying Eqn.\ref{eqn:meanp0} to Eqn.\ref{eqn:CIbasic} gives the lower and upper bounds of the approximated 95\% CI for $\pstar_0$:
\begin{equation}
    \begin{aligned} 
    \mathcal{L}(\pstar_0),\mathcal{U}(\pstar_0)=
    (\Ddot{\mu}_{\hat{p}_0} \mp 1.96
    (\Ddot{\sigma}_{\hat{p}_0}/\sqrt{s})
    -\alpha_1')/(\alpha_0-\alpha_1')
    \end{aligned}
    \label{eqn:CIpstar}
\end{equation}
This gives us the 
interval estimate of CLEAM,
$\rhocleam=[\mathcal{L}(\pstar_0), \mathcal{U}(\pstar_0)]$, 
a range of values that we can be approximately 95\% confident to contain
$p^*_0$.
The range of possible values for $\pstar_1$ can be simply derived considering $\pstar_1 = 1 - \pstar_0$. The overall procedure of CLEAM is summarized in Alg. \ref{alg:cleam}. 
Now, with the IE, we can provide statistical significance to the reported fairness improvements.

%---------------------------- Waiting for response to push to Supp---------------
\iffalse
\textcolor{cyan}{
{\bf Goodness-of-fit}. Lastly, recognizing CLEAM's dependency on the accurate statistical modeling of $\phat$, we conducted a Kolmogorov–Smirnov (KS) test \cite{prattConceptsNonparametricTheory1981} ($\delta=0.05$ with $D_{crit}=0.24$) to determine the goodness-of-fit between the theoretical distribution, Eqn. \ref{eqn:meanp0} and \ref{eqn:varp0}, and the sample based distribution, Eqn. \ref{eqn:meanp0samples} and \ref{eqn:varp0samples}, on the setup in Sec. \ref{subsec:empiricalStudy} using the GCelebA dataset with ResNet-18. 
Note that we found $\balpha$ evaluated 
during classifier training to be a close approximation of $\balpha_{Gen}$, the accuracy when evaluated on the GCelebA. Hence, {\em we utilize $\balpha$ (which is readily available) for our evaluations and CLEAM}. See Supp\ref{subsec:accOfGenaVsReala} for a detailed comparison.
With this, our results in Tab. \ref{tab:KStestResNet18} shows that the test statistic, $\eta<D_{crit}$, and hence, we are 95\% confident that the two distributions are statistically similar, thereby validating the statistical model and CLEAM. 
}
\fi
%---------------------------- Waiting for response to push to Supp---------------

%%%%%%%%%%%%%%%%%%%%%%%%%%%%%%%%%%%%%%%%%%%%%%%%
%       Table Chunk 1 <start>
%%%%%%%%%%%%%%%%%%%%%%%%%%%%%%%%%%%%%%%%%%%%%%%%
\clearpage
\begin{table*}[!t]
    \centering
    %New overall Caption
    \caption{
    %\textcolor{blue}{
    Comparing the {\em point estimates} and {\em interval estimates} of Baseline \cite{choiFairGenerativeModeling2020a}, Diversity \cite{keswaniAuditingDiversityUsing2021} and our CLEAM in estimating $\bpstar$ of 
    StyleGAN2 \cite{karrasStyleBasedGeneratorArchitecture2019} and StyleSwin \cite{zhang2021styleswin} with the proposed GenData datasets.
    %the GenData-StyleGAN2/StyleSwin datasets. 
    We utilize SA classifiers Resnet-18/34 (R18, R34)\cite{heDeepResidualLearning2016}, MobileNetv2 (MN2)\cite{sandlerMobileNetV2InvertedResiduals2018} and VGG-16 (V16)\cite{simonyanVeryDeepConvolutional2014}, with different accuracies $\balpha$, to classify samples \wrt attributes \texttt{Gender} and \texttt{BlackHair}.
    The $\pstar_0$ value of each GAN \wrt SA is determined by manually hand-labeling the generated data. 
    %(\textcolor{red}{see Supp. XX for more details on the GenData datasets}).
    We repeat this for 5 experimental runs and report the mean error rate, per Eqn. \ref{eqn:pointError}. 
    See Supp D.1 for the standard deviation of PE and IE.
    }
\resizebox{\textwidth}{!}{
    \addtolength{\tabcolsep}{-4pt}
      \begin{tabular}
      {ccc  cc cccc     cc cc cc}
        \toprule
        & & & \multicolumn{6}{c}{\bf Point Estimate} & \multicolumn{6}{c}{\bf Interval Estimate}
        \\
        \cmidrule(lr){4-9}\cmidrule(lr){10-15}
          %\textbf{Classifier} 
          &
          \textbf{  $\balpha=\{\alpha_0,\alpha_1\}$ } &
          Avg. $\balpha$ &
          %\textbf{  $\balpha=\{\alpha_0,\alpha_1\}$ } &
          \multicolumn{2}{c}{\textbf{Baseline}}
          %\cite{choiFairGenerativeModeling2020a}}
          & 
          \multicolumn{2}{c}{\textbf{Diversity}}
          %\cite{keswaniAuditingDiversityUsing2021}} 
          & \multicolumn{2}{c}{\textbf{CLEAM (Ours)}}
          %Interval Estimates
           & \multicolumn{2}{c}{\textbf{Baseline}}
           %\cite{choiFairGenerativeModeling2020a}}
           & 
           \multicolumn{2}{c}{\textbf{Diversity}}
           %\cite{keswaniAuditingDiversityUsing2021}} 
           & \multicolumn{2}{c}{\textbf{CLEAM (Ours)}}
        \\
        %\cmidrule(lr){1-1}
        \cmidrule(lr){2-2}\cmidrule(lr){3-3}\cmidrule(lr){4-5}\cmidrule(lr){6-7} \cmidrule(lr){8-9}
        \cmidrule(lr){10-11}\cmidrule(lr){12-13} \cmidrule(lr){14-15}
         %Baseline
        & & &
        $\mu_{\texttt{Base}}$ & $e_{\mu}(\downarrow)$ 
         %Diversity
         &$\mu_{\texttt{Div}}$ & $e_\mu(\downarrow)$
         %CLEAM
        &$\mu_{\texttt{CLEAM}}$ & $e_\mu(\downarrow)$
        &$\rho_{\texttt{Base}}$ & $e_\rho(\downarrow)$ 
        &$\rho_{\texttt{Div}}$ & $e_\rho(\downarrow)$ 
        &$\rho_{\texttt{CLEAM}}$ & $e_\rho(\downarrow)$\\
        \midrule
        \multicolumn{15}{c}{\cellcolor{lightgray!20} \bf (A) StyleGAN2}\\
        \midrule
        \multicolumn{15}{c}{\texttt{Gender} with GT class probability {\bfseries\boldmath $p^*_0$=0.642} }\\
        \midrule
        R18 & \{0.947, 0.983\} & 0.97 &
        %Point Estimate----------------------------------------------------------
          0.610 &   4.98\% & 
        %Diversity
          --- & 
          --- & 
        %CLEAM
          0.638 &   \textbf{0.62\%} & 
        %Interval Estimate----------------------------------------------------------
        [0.602, 0.618] & 6.23\% & 
        %Diversity
        --- & --- & 
        %CLEAM
       [0.629, 0.646] & \textbf{2.02\%}
        \\  
        %\hdashline
        %Point Estimate----------------------------------------------------------
        R34 & \{0.932, 0.976]\} & 0.95 & 
          0.596 &   7.17\% &
        %Diversity
          --- & 
          --- & 
        %CLEAM
          0.634 & 
          \textbf{1.25\%} &
        %Interval Estimate----------------------------------------------------------
        [0.589, 0.599] & 8.26\% &
        %Diversity
        --- & --- &
        %CLEAM
        [0.628, 0.638] & \textbf{2.18\%} 
        \\ 
        %\hdashline
        %Point Estimate----------------------------------------------------------
        MN2 & \{0.938, 0.975\} & 0.96 & 
          0.607 &   5.45\% & 
        %Diversity
          --- &   --- &  
        %CLEAM
          0.637  &   \textbf{0.78}\% &
        %Interval Estimate----------------------------------------------------------
        [0.602, 0.612] & 6.23\% &
        %Diversity
        --- & --- & 
        %CLEAM
        [0.632, 0.643] & $\textbf{1.56\%}$
        \\ 
        %\hdashline
        %Point Estimate----------------------------------------------------------
        V16 & \{0.801, 0.919\} & 0.86 & 
          0.532 &   17.13\% & 
        %Diversity
          0.550 &   14.30\% & 
        %CLEAM
          0.636  &   \textbf{0.93}\% &
        %Interval Estimate----------------------------------------------------------
        [0.526, 0.538] & 18.06\% &
        %Diversity
        [0.536 , 0.564] & 16.51\% & 
        %CLEAM
        [0.628, 0.644] & \textbf{2.18\%}
        \\
        \midrule
        %Point Estimate----------------------------------------------------------
        \multicolumn{4}{c}{Average Error:}  & 
          8.68\%
        &   & 
          14.30\%&
            &
         \textbf{0.90}\%
        %Interval Estimate----------------------------------------------------------
        &  & 9.70\%
        &  & 16.51\%
        &  & \textbf{1.99}\%
        \\
        \midrule
    %BlackHair
        \multicolumn{15}{c}{\texttt{BlackHair} with GT class probability {\bfseries\boldmath $p^*_0$=0.643}}\\
        \midrule
        %Point Estimate----------------------------------------------------------
        R18 & \{0.869, 0.885\} & 0.88 & 
          0.599 &   6.84\% & 
        %Diversity
          --- &   --- &  
        %CLEAM
          0.641 &   \textbf{0.31\%} &
        %Interval Estimate----------------------------------------------------------
        [0.591, 0.607] & 8.08\% & 
        %Diversity
        --- & --- & 
        %CLEAM
        [0.631, 0.652] & \textbf{1.40\%}
        \\  
        %\hdashline
        %Point Estimate----------------------------------------------------------
        R34 & \{0.834, 0.916\}  & 0.88 & 
          0.566 &   11.98\%&
        %Diversity
          --- &   --- & 
        %CLEAM
          0.644 & 
          \textbf{0.16}\% &
        %Interval Estimate----------------------------------------------------------
        [0.561, 0.572] & 12.75\% &
        %Diversity
        --- & --- &
        %CLEAM
        [0.637, 0.651] & \textbf{1.24\%}
        \\ %\hdashline
        %Point Estimate----------------------------------------------------------
        MN2 & \{0.839, 0.881\} & 0.86 & 
          0.579 &   9.95\% & 
        %Diversity
          --- &   --- &  
        %CLEAM
          0.639  &   \textbf{0.62}\% &
        %Interval Estimate----------------------------------------------------------
        [0.574, 0.584] & 10.73\% &
        %Diversity
        --- & --- & 
        %CLEAM
        [0.632, 0.647] & \textbf{1.71\%}
        \\ %\hdashline
        %Point Estimate----------------------------------------------------------
        V16 & \{0.851, 0.836\} & 0.84 & 
          0.603 &   6.22\% & 
        %Diversity
          0.582 &   9.49\% & 
        %CLEAM
          0.640  &   \textbf{0.47}\%
        %Interval Estimate----------------------------------------------------------
        & 
        [0.597, 0.608] & 7.15\% &
        %Diversity
        [0.568, 0.596] & 11.66\% &
        %CLEAM
        [0.632, 0.648] & \textbf{1.71\%}
        \\
        \midrule
        %Point Estimate----------------------------------------------------------
          \multicolumn{4}{c}{Average Error:} &   8.75\%
        &     &   9.49\% 
        &     &   \textbf{0.39}\%
        %Interval Estimate----------------------------------------------------------
        &  & 9.68\%
        &  & 11.66\%
        &  & \textbf{1.52}\%
        \\
        \midrule
        \multicolumn{15}{c}{\cellcolor{lightgray!20} \bf (B) StyleSwin}\\
        \midrule
        \multicolumn{15}{c}{\texttt{Gender} with GT class probability {\bfseries\boldmath $p^*_0$=0.656}}\\
        \midrule
        R18 & \{0.947, 0.983\} & 0.97 & 
        %Point Estimate----------------------------------------------------------
          0.620 &   5.49\% & 
        %Diversity
          --- &   --- &  
        %CLEAM
          0.648 &   \textbf{1.22\%} 
        %Interval Estimate----------------------------------------------------------
        & [0.612,0.629] & 6.70\% & 
        %Diversity
        --- & --- & 
        %CLEAM
        [0.639,0.658] & \textbf{2.59\%}
        \\  %\hdashline
        R34 & \{0.932, 0.976\}  & 0.95 & 
        %Point Estimate----------------------------------------------------------
          0.610 &   7.01\% &
        %Diversity
          --- &   --- & 
        %CLEAM
          0.649 &   \textbf{1.07}\%  
        %Interval Estimate----------------------------------------------------------
        & [0.605,0.615] & 7.77\%&
        %Diversity
        --- & --- &
        %CLEAM
        [0.643,0.654] & 
        \textbf{1.98}\% 
        \\ %\hdashline
        MN2 & \{0.938, 0.975\} & 0.96 & 
        %Point Estimate----------------------------------------------------------
          0.623 &   5.03\% & 
        %Diversity
          --- &   --- & 
        %CLEAM
          0.655  &   \textbf{0.15}\% 
        %Interval Estimate----------------------------------------------------------
        & [$0.618,0.629$] & $5.79\%$ & 
        %Diversity
        --- & --- & 
        %CLEAM
        [0.649,0.661]  & \textbf{1.07}\%
        \\ %\hdashline
        V16 & \{0.801, 0.919\} & 0.86 & 
        %Point Estimate----------------------------------------------------------
          0.555 &   15.39\% & 
        %Diversity
          0.562 &   14.33\% & 
        %CLEAM
          0.668  &   \textbf{1.83}\% 
         %Interval Estimate----------------------------------------------------------
        &[0.549,0.560] & 16.31\% & 
        %Diversity
        [0.548,0.576] & 16.46\% & 
        %CLEAM
        [0.660,0.675]  & \textbf{2.90}\%
        \\
        \midrule
        %Point Estimate----------------------------------------------------------
          \multicolumn{4}{c}{Average Error:} &   8.23\%
        &     &   14.33\%
        &     &   \textbf{1.07}\%
        %Interval Estimate----------------------------------------------------------
        &   & 9.14\%
        &  & 16.46\% 
        &  & \textbf{2.14}\%
        \\
        \midrule
    %BlackHair
        \multicolumn{15}{c}{\texttt{BlackHair} with GT class probability {\bfseries\boldmath $p^*_0$=0.668}}\\
        \midrule
        R18 & \{0.869, 0.885\} & 0.88 & 
        %Point Estimate----------------------------------------------------------
          0.612 &   8.38\% & 
        %Diversity
          --- &   --- &  
        %CLEAM
          0.659 &   \textbf{1.35\%} 
        %Interval Estimate----------------------------------------------------------
        & [0.605,0.620] & 9.43\% &
        %Diversity
        --- & --- & 
        %CLEAM
        [0.649,0.670] & \textbf{2.84\%} 
        \\  %\hdashline
        R34 & \{0.834, 0.916\} & 0.88 & 
        %Point Estimate----------------------------------------------------------
          0.581 &   13.02\%&
        %Diversity
          --- &   --- & 
        %CLEAM
          0.662 &   \textbf{0.90}\%  
        %Interval Estimate----------------------------------------------------------
        & [0.576,0.586] & 13.77\%&
        %Diversity
        --- & --- &
        %CLEAM
        [0.656,0.669] & 
        \textbf{1.80}\%
        \\ %\hdashline
        MN2 & \{0.839, 0.881\} & 0.86 & 
        %Point Estimate----------------------------------------------------------
          0.596 &   10.78\% & 
        %Diversity
          --- &   --- & 
        %CLEAM
          0.659  &   \textbf{1.35}\% 
        %Interval Estimate----------------------------------------------------------
        & [0.591,0.600] & 11.50\% & 
        %Diversity
        --- & --- & 
        %CLEAM
        [0.652,0.666]  & \textbf{2.40}\%
        \\ %\hdashline
        V16 & \{0.851, 0.836\} & 0.84 & 
        %Point Estimate----------------------------------------------------------
          0.625 &   6.44\% & 
        %Diversity
          0.608 &   8.98\% & 
        %CLEAM
          0.677  &   \textbf{1.35}\% 
        %Interval Estimate----------------------------------------------------------
        & [0.620,0.630] & 7.19\% & 
        %Diversity
        [0.590,0.626] & 11.68\% & 
        %CLEAM
        [0.670,0.684]  & \textbf{2.40}\%
        \\
        \midrule
        %Point Estimate----------------------------------------------------------
          \multicolumn{4}{c}{Average Error:} &   9.66\% 
        &     &   8.98\% 
        &     &   \textbf{1.24}\%
        %Interval Estimate----------------------------------------------------------
        &   & 10.47\% 
        &  & 11.68\%
        &  & \textbf{2.36}\%
        \\
        \midrule
    \end{tabular}
    }
    \label{tab:G_PE}
    \addtolength{\tabcolsep}{4pt}
    %\vspace{-4mm}
    \vspace{-4mm}
\end{table*}
%\vfill
\begin{table}[t]
    \centering
    %New overall Caption
    \caption{Comparing the {\em point estimates} and {\em interval estimates} of Baseline and CLEAM in estimating the $\bpstar$ of the Stable Diffusion Model \cite{rombach2021highresolution} with the GenData-SDM dataset.
   We use prompt input starting with "A photo with the face of" and ending with synonymous (Gender neutral) prompts. We utilized CLIP as the classifier for \texttt{Gender}, to obtain $\bphat$.
    }
\resizebox{\textwidth}{!}{
      \begin{tabular}
      {c c  cc cc cc cc }
        \toprule
        & & \multicolumn{4}{c}{\bf Point Estimate} & \multicolumn{4}{c}{\bf Interval Estimate}\\
        \cmidrule(lr){3-6}\cmidrule(lr){7-10}
          \textbf{Prompt} & \textbf{GT} & \multicolumn{2}{c}{\textbf{Baseline}
          }& \multicolumn{2}{c}{\textbf{CLEAM (Ours)}}
          & \multicolumn{2}{c}{\textbf{Baseline}
          } & \multicolumn{2}{c}{\textbf{CLEAM (Ours)}}
        \\
        \cmidrule(lr){1-1} \cmidrule(lr){2-2} \cmidrule(lr){3-4} \cmidrule(lr){5-6} \cmidrule(lr){7-8} \cmidrule(lr){9-10}
         %Baseline
        & $\pstar_0$ & $\mu_{\texttt{Base}}$ & $e_\mu (\downarrow)$ 
         %CLEAM
        &$\mu_{\texttt{CLEAM}}$ & $e_\mu(\downarrow)$
         %Baseline
        &$\rho_{\texttt{Base}}$ & $e_\rho (\downarrow)$ 
         %CLEAM
        &$\rho_{\texttt{CLEAM}}$ & $e_\rho(\downarrow)$
        \\
        \midrule
        \multicolumn{10}{c}{$\balpha$=[0.998,0.975], Avg. $\alpha$=0.987, CLIP --\texttt{Gender}}\\
        \midrule
        %Point Estimate--------------------------------------------------
        "A photo with the face of \underline{an individual}" & 0.186 & 0.203 & 9.14\% & 0.187 & {\bf 0.05\%} & [ 0.198 , 0.208 ] & 11.83\% & [ 0.182 , 0.192 ] & {\bf 3.23\%} \\
        "A photo with the face of \underline{a human being}"   & 0.262 & 0.277  & 5.73\%  & 0.263  & {\bf  0.38\%} & [ 0.270 , 0.285 ] & 8.78\% & [ 0.255 , 0.271 ] & {\bf 3.44\%}  \\
        "A photo with the face of \underline{one person}"    & 0.226 & 0.241  & 6.63\% & 0.230  & {\bf 1.77\%} & [ 0.232 , 0.251 ] & 11.06\% &  [ 0.220 , 0.239 ] & {\bf 5.75\%} \\
        "A photo with the face of \underline{a person}"      &  0.548 & 0.556 & 1.49\% & 0.548 & {\bf 0.00\%} & [ 0.545 , 0.566 ] & 3.28\% & [ 0.537 , 0.558 ] & {\bf 2.01\%} \\
        \midrule
        \multicolumn{3}{c}{Average Error} & 5.75\% & & {\bf 0.44\%} & & 8.74\% & & {\bf 3.61\%}\\
        %CLEAM
        %Interval Estimate--------------------------------------------------
        %CLEAM
        \bottomrule
    \end{tabular}
    }
    \label{tab:G_PE2}
\end{table}
\begin{figure}[ht!]
        \vspace{-0.2cm}
        \centering
        %\begin{subfigure}[b]{\textwidth}
            \includegraphics[width=\linewidth]{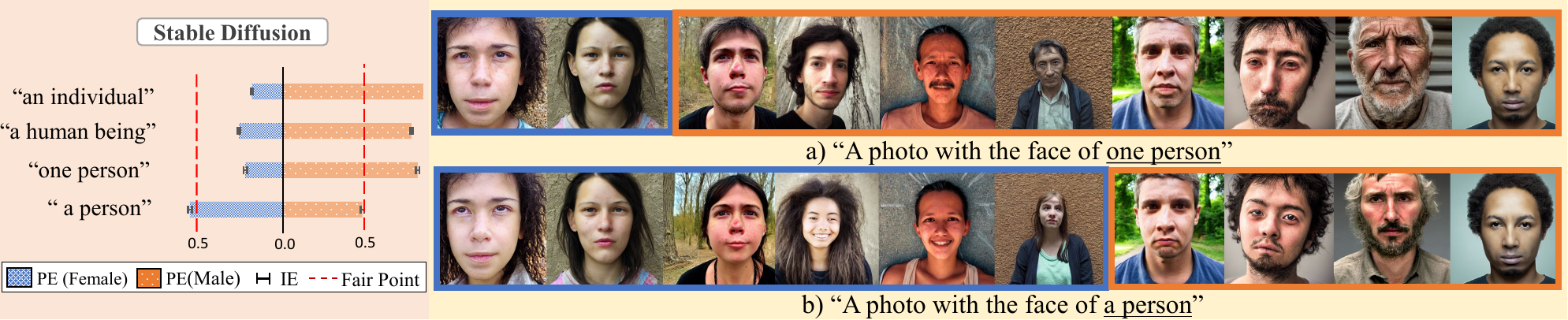}
            \caption{{\bf LHS:} Applying CLEAM to assess \texttt{Gender} bias in SDM\cite{rombach2021highresolution} with CLIP \cite{radfordLearningTransferableVisual2021}. Here, we utilize synonymous neutral prompts, prefixed with "A photo with the face of \_\_" as the input. We found that subtle changes to the prompts resulted in significant changes in the bias. {\bf RHS:} Illustrating the shift in \texttt{Gender} bias. Using the same random seeds (per column), we generate 10 samples from two prompts. Note how in some columns, \texttt{Gender} changes while retaining general features \eg pose. 
            }
             \label{fig:stabledifussionillustration}
       \vspace{-1.0em}
\end{figure}

%%%%%%%%%%%%%%%%%%%%%%%%%%%%%%%%%%%%%%%%%%%%%%%%
%       Table Chunk 1 <end>
%%%%%%%%%%%%%%%%%%%%%%%%%%%%%%%%%%%%%%%%%%%%%%%%

%%%%%%%%%%%%%%%%%%%%%%%%%%%%%%%%%%%%%%%
%        Experiments
%%%%%%%%%%%%%%%%%%%%%%%%%%%%%%%%%%%%%%%
\section{Experiments}
\label{sec:experiments}

In this section, we first
evaluate fairness measurement accuracy of CLEAM on both GANs and SDM (Sec.\ref{sec:evalrealG}) with our proposed GenData dataset.
Then
we evaluate CLEAM's robustness through some ablation studies (Sec. \ref{subsec:ablationstudies}).
To the best of our knowledge, there is no similar 
literature for improving fairness measurements in generative models. Therefore, we compare {\bf CLEAM} with the two most related works: a) the {\bf Baseline} used in previous works \cite{choiFairGenerativeModeling2020a,teo2022fair,frankelFairGenerationPrior2020,tanImprovingFairnessDeep2020,umFairGenerativeModel2021}  
b) {\bf Diversity} \cite{keswaniAuditingDiversityUsing2021} which computes disparity within a dataset
via an intra-dataset pairwise similarity algorithm.
We remark that, as discussed by Keswani \etal \cite{keswaniAuditingDiversityUsing2021} Diversity is model-specific using VGG-16 \cite{simonyanVeryDeepConvolutional2014}; see Supp. D.2 for more details.
Finally, unless specified, we repeat the experiments with $s=30$ batches of 
images from the generators with 
batch size
$n=400$.
For a fair comparison, all three algorithms use the exact same inputs. However, while Baseline and Diversity ignore 
the SA classifier inaccuracies,
CLEAM makes good use of it to rectify the measurement error.
As mentioned in Sec. \ref{sub:approxMLEpstar}, for CLEAM, we utilize $\balpha$ measured on real samples, 
which we found to be a good approximation of the $\balpha$ measured on generated samples (see Supp. D.7 for results).
We repeat each experiment 5 times and report the mean value for each test point for both PE and IE. See Supp 
D.1 
for the standard deviation.

\subsection{Evaluating CLEAM's Performance}
\label{sec:evalrealG}

{\bf CLEAM for fairness measurement of SOTA GANs -- StyleGAN2 and StyleSwin.}
For a fair comparison, we first compute $s$ samples of $\bphat$, one for each batch of $n$ images. For Baseline, we use the mean $\bphat$ value as the PE (denoted by $\mubase$), and the $95\%$ confidence interval as IE ($\rhobase$). 
With the same $s$ samples of $\bphat$, we apply Alg. \ref{alg:cleam} to obtain $\mucleam$ and $\rhocleam$.  
For Diversity, following the original source code \cite{keswaniAuditingDiversityUsing2021}, a controlled dataset 
with 
fair
representation is randomly selected from a held-out dataset of CelebA-HQ \cite{CelebAMask-HQ}. Then, we use a VGG-16 \cite{simonyanVeryDeepConvolutional2014}
feature extractor and compute Diversity, $\delta$.
With $\delta$ we find
$\phat_0=(\delta+1)/2$ and subsequently 
$\mu_\texttt{Div}$ and $\rho_\texttt{Div}$ from the mean and $95\%$ CI (see Supp %\textcolor{blue}{
D.2
for more details on diversity).
We then compute $e_{\mu_\texttt{CLEAM}}$, $e_{\mu_\texttt{Div}}$, $e_{\rho_\texttt{CLEAM}}$ and $e_{\rho_\texttt{Div}}$ with Eqn \ref{eqn:pointError},
by replacing the Baseline estimates with CLEAM and Diversity.

As discussed, our results in Tab.\ref{tab:G_PE} show that the baseline experiences significantly large errors of $4.98\% \leq \emubase \leq 17.13\%$, due to a lack of consideration for the inaccuracies of the SA classifier. We note that this problem is prevalent throughout the different SA classifier architectures, 
even 
with
higher capacity classifiers \eg ResNet-34. 
Diversity, a method similarly unaware of the inaccuracies of the SA classifier, presents a similar issue with $8.98\% \leq \emudiv \leq 14.33\%$ 
In contrast, CLEAM dramatically reduces the error for all classifier architectures. 
Specifically, CLEAM reduces the average point estimate error from $\emubase \geq$  8.23\% to $\emucleam \leq$ 1.24\%,
in both StyleGAN2 and StyleSwin. The IE presents similar results, where in most cases $\rho_\texttt{CLEAM}$
bounds the GT value of
$\bpstar$.

{\bf CLEAM for fairness measurement of SDM.}
We evaluate CLEAM in estimating the bias of the SDM \wrt \texttt{Gender}, based on the synonymous (gender-neutral) prompts
introduced in Sec. \ref{subsec:empiricalStudy}. 
Recall that here we utilize CLIP as the zero-shot SA classifier.
Our results in Tab \ref{tab:G_PE2}, as discussed, show that utilizing the baseline results in considerable error ($1.49\% \leq \emubase \leq 9.14\%$) for all prompts, even though the SA classifier's average accuracy was high, $\approx 98.7\%$ (visual results in Fig.\ref{fig:stabledifussionillustration}). 
A closer look at the theoretical model's Eqn.~\ref{eqn:meanp0} reveals that this is due to the larger inaccuracies observed in the biased class ($\alpha_1'$) coupled with the large bias seen in $\pstar_1$, which results in $\mubase$ deviating from $\pstar_0$.
In contrast, CLEAM accounts for these inaccuracies and significantly  
minimizes the error to $e_{\mu_\texttt{CLEAM}}\leq 1.77\%$. Moreover, CLEAM's IE is able to consistently bound the GT value of $\pstar_0$.

\subsection{Ablation Studies and Analysis}
\label{subsec:ablationstudies}
Here, we perform the ablation studies and compare CLEAM with classifier correction methods. \textbf{\em We remark that detailed results of these experiments are provided in the Supp due to space limitations.}

{\bf CLEAM for measuring varying degrees of bias.}
As we cannot control
the bias in 
trained generative models, to simulate different degrees of bias,
we evaluate CLEAM with a {\em pseudo-generator}.
Our results show that  
CLEAM is effective at different biases ($\pstar_0\in$ [0.5,0.9]) reducing the average error from 2.80\% $\leq \emubase \leq$ 6.93\% to 
$\emucleam \leq$ 0.75\% on CelebA \cite{liuDeepLearningFace2015} \wrt \texttt{\{Gender,BlackHair\}}, and AFHQ \cite{choiStarGANV2Diverse2020}  \wrt \texttt{Cat/Dog}. 
See Supp D.3 and D.4 for full experimental results.

{\bf CLEAM vs Classifier Correction Methods \cite{liptonDetectingCorrectingLabel2018}}.
CLEAM generally accounts for the classifier's inaccuracies, without targeting any particular cause of inaccuracies, for the purpose of 
rectifying the fairness measurements. 
This objective is unlike traditional classifier correction methods as it does not aim to improve the actual classifier's accuracy. 
However, considering that classifier correction methods may improve the fairness measurements by directly rectifying the classifier inaccuracies, we compare its performance against CLEAM. 
As an example, we utilize the Black Box Shift Estimation / Correction (BBSE / BBSC) \cite{liptonDetectingCorrectingLabel2018} which considers the label shift problem and aims to correct the classifier output by detecting the distribution shift.
Our results, based on Sec. \ref{sec:evalrealG} setup, show that while BBSE does improve on the
fairness measurements of the baseline \ie 4.20\% $\leq \emubbse \leq$ 3.38\%, these results are far inferior to CLEAM's results seen in Tab. \ref{tab:G_PE}. In contrast, BBSC demonstrates no improvements in fairness measurements. 
See Supp D.8 for full experimental results.
We postulate that this is likely due to the strong assumption of label shift made by both methods.

{\bf Effect of batch-size.} 
Utilizing experimental setup in Sec. \ref{sec:evalrealG} for
batch size $n \in$[100,600], our results 
in Fig. \ref{fig:ResNet18Permuten}
show that $n$=400 is an ideal batch size, balancing computational cost and measurement accuracy. 
See Supp F for full experimental details and results.

\begin{figure}[h]
    \centering
    \includegraphics[width=0.7\linewidth]{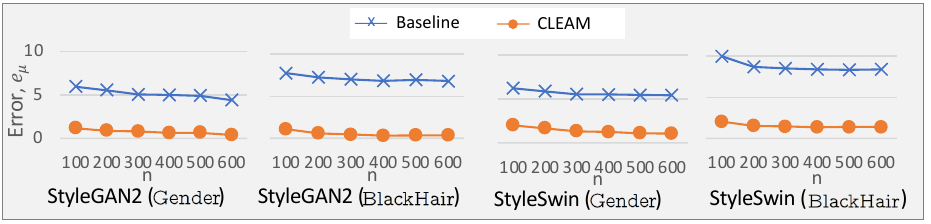}
    \caption{
    Comparing the point error $e_{\mu}$ for Baseline and CLEAM when evaluating the bias of GenData-CelebA
    with ResNet-18, while varying sample size, $n$. 
    }
    \label{fig:ResNet18Permuten}
    \vspace{-0.5cm}
\end{figure}

%%%%%%%%%%%%%%%%%%%%%%%%%%%%%%%%%%%%%%%%%%%%%
%      APPLICATION OF CLEAM
%%%%%%%%%%%%%%%%%%%%%%%%%%%%%%%%%%%%%%%%%%%%%
\section{Applying CLEAM: Bias in Current SOTA Generative Models}
\label{sec:application}

In this section, we leverage 
the improved reliability of CLEAM to 
study
 biases in the popular generative models.
Firstly, with the rise in popularity of text-to-image generators \cite{rombach2021highresolution,rameshZeroShotTexttoImageGeneration2021,galStyleGANNADACLIPGuidedDomain2021,patashnikStyleCLIPTextDrivenManipulation2021}, 
we revisit our results when passing different prompts, with synonymous neutral meanings to an SDM, and take a closer look at how subtle prompt changes can impact bias \wrt \texttt{Gender}. Furthermore, we further investigate if similar results would occur in other SA, \texttt{Smiling}.  
Secondly, with the shift in popularity from convolution to transformer-based architectures \cite{raghuVisionTransformersSee2022,hudsonGenerativeAdversarialTransformers2021,jiaVisualPromptTuning2022}, due to its better sample quality, we determine whether the learned bias would also change.
For this, we compare StylesSwin (transformer) and StyleGAN2 (convolution), which are both 
based on the same architecture backbone.

\begin{figure}[t]
        \centering
        \begin{subfigure}[b]{\textwidth}
             \includegraphics[width=\linewidth]{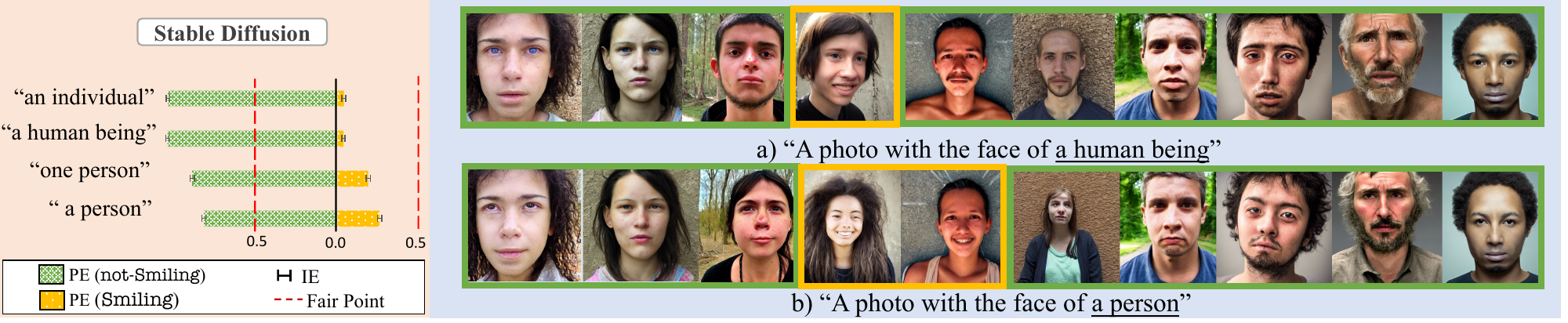}
            \caption{
            Evaluating bias of SDM \cite{rombach2021highresolution} \wrt SA \texttt{Smiling} using CLEAM.
            {\bf LHS:} Bias value with different text prompts as input to SDM. CLEAM identifies that having ``person'' in the prompt results in more \texttt{Smiling} samples generated.
            {\bf RHS:} 
            Visual comparison of the samples generated for two different but semantically similar text prompts. Samples in the same column are generated using the same random seed. Notice how the SA (\texttt{smiling}) of the samples changes with a slight difference in the prompts.
            }
            \label{fig:stabledifussionillustration2}
        \end{subfigure}
        \begin{subfigure}[b]{\textwidth}
            \includegraphics[width=\linewidth]{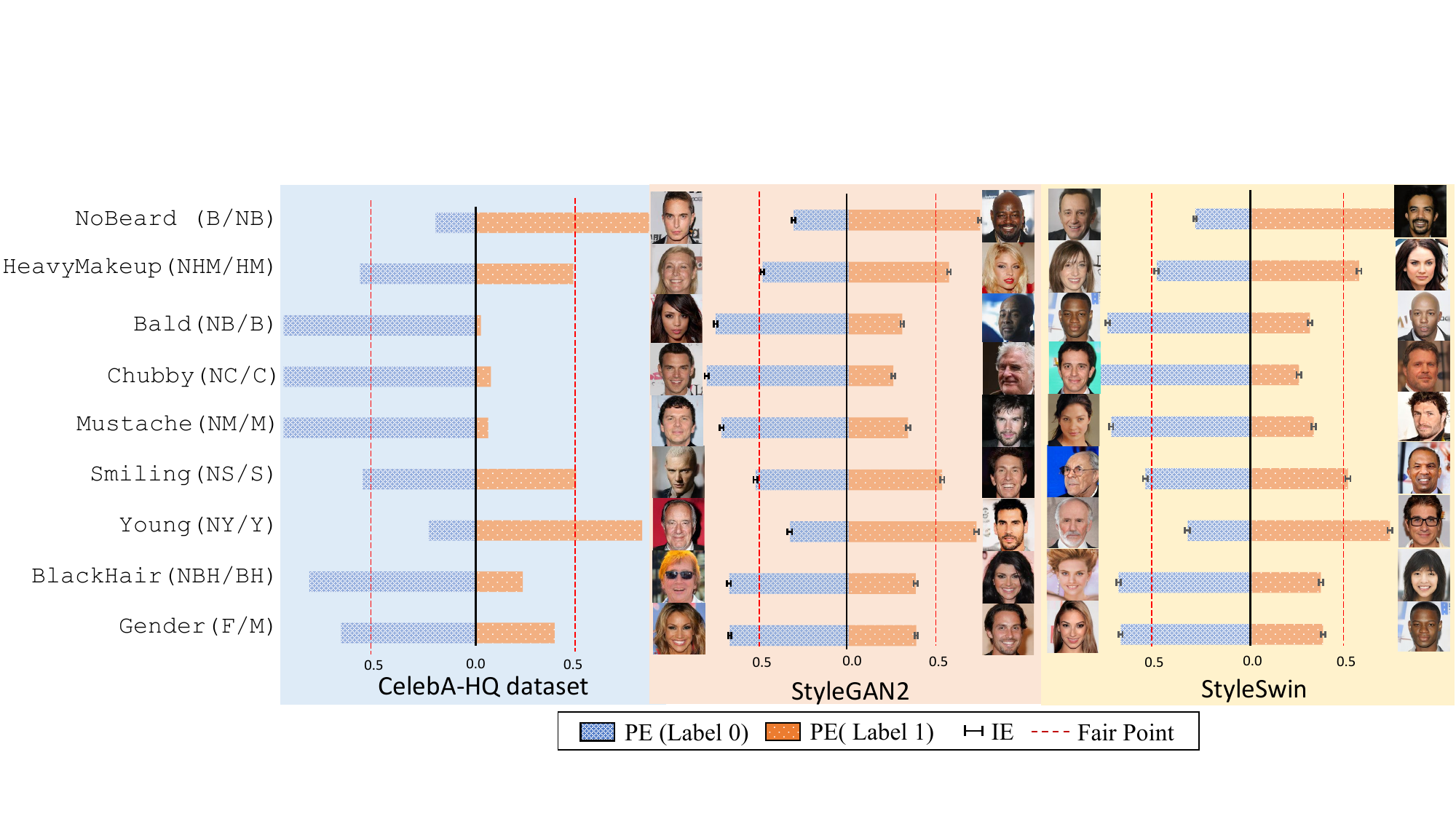}
            \caption{
            CLEAM on StyleGAN2 \cite{karrasStyleBasedGeneratorArchitecture2019} and StyleSwin \cite{zhang2021styleswin}, both pre-trained on CelebA-HQ but are based on different architecture variants.
            We utilize a ResNet-18 and compare the bias \wrt various SAs.
            }
             \label{fig:StyleGANvsStyleSwin}
        \end{subfigure}
        \caption{Applying CLEAM to further assess the bias in popular generative models. %SDM 
        }
        \vspace{-0.3cm}
\end{figure}

Our results, on SDM, demonstrate that the use of different synonymous neutral prompts \cite{haspelmathIndefinitePronouns1997,saguy2022little} results in different degrees of bias \wrt both \texttt{Gender} and \texttt{Smiling} attributes.
For example in Fig. \ref{fig:stabledifussionillustration}, a 
semantically insignificant
prompt change from "\underline{one} person" to "\underline{a} person" results in a significant shift in \texttt{Gender} bias. 
Then, in Fig. \ref{fig:stabledifussionillustration2}, we observe that while the SDM \wrt our prompts appear to be heavily biased to not-\texttt{Smiling}, having "person" in the prompt appears to significantly reduce this bias. 
This suggests that for SDM, even semantically %\textcolor{blue}{
similar
neutral prompts \cite{haspelmathIndefinitePronouns1997,saguy2022little} could result in different degrees of bias,
thereby demonstrating certain instability in SDM.
Next, our results in Fig. \ref{fig:StyleGANvsStyleSwin} compare the bias in StyleGAN2, StylesSwin, and the training CelebA-HQ dataset over an extended number of SAs. 
Overall, we found that while StyleSwin 
produces
better quality samples \cite{zhang2021styleswin}, the same biases still remain statistically unchanged between the two architectures \ie their IE overlap. 
Interestingly, our results also found that both the GANs were less biased than the training dataset itself.

%%%%%%%%%%%%%%%%%%%%%%%%%%%%%%%%%%%%%%%%%%%%%%%%
%       Table Chunk 2 <start>
%%%%%%%%%%%%%%%%%%%%%%%%%%%%%%%%%%%%%%%%%%%%%%%%

%%%%%%%%%%%%%%%%%%%%%%%%%%%%%%%%%%%%%%%%%%%%%%%%
%       Table Chunk 2 <end>
%%%%%%%%%%%%%%%%%%%%%%%%%%%%%%%%%%%%%%%%%%%%%%%%

%%%%%%%%%%%%%%%%%%%%%%%%%%%%%%%%%%%%
%     Conclusion
%%%%%%%%%%%%%%%%%%%%%%%%%%%%%%%%%%%
\section{Discussion}
%\textcolor{blue}{
{\bf Conclusion.}
In this work, we address the limitations of the existing fairness measurement framework.
Since generated samples are typically unlabeled, we first introduce a new labeled dataset based on three state-of-the-art generative models for our studies.
Our findings suggest that the existing framework, which ignores classification inaccuracies, suffers from significant measurement errors, even when the SA classifier is very accurate. %\eg accuracy$>$97\%. 
To rectify this, we propose CLEAM, which considers these inaccuracies in its statistical model and outputs a more accurate fairness measurement.
Overall, CLEAM demonstrates improved accuracy over extensive experimentation, including both real generators and controlled setups.
Moreover, by applying CLEAM to popular generative models, we uncover significant biases that raise efficacy concerns about these models' real-world application.

%%%%%%%%%%%%%%%%%%%%%%%%%%%%%%%%%%%%
%    Broader Impact
%%%%%%%%%%%%%%%%%%%%%%%%%%%%%%%%%%%
{\bf Broader Impact.} Given that generative models are becoming more widely integrated into our everyday society \eg text-to-image generation, it is important that we have reliable means to measure fairness in generative models, thereby allowing us to prevent these biases from proliferating into new technologies. CLEAM provides a step in this direction by allowing for more accurate evaluation.
We remark that our work {\em does not introduce any social harm} but instead improves on the already existing measurement framework.

%%%%%%%%%%%%%%%%%%%%%%%%%%%%%%%%%%%%
%    Limitations
%%%%%%%%%%%%%%%%%%%%%%%%%%%%%%%%%%%

{\bf Limitations.} Despite the effectiveness of the proposed method along various generative models, our work addresses only one facet of the problems in the existing fairness measurement and there is still room for further improvement. For instance, it may be beneficial to consider SA to be non-binary \eg 
when hair color is not necessary fully black (grey).
Additionally, existing datasets used to train classifiers are commonly human-annotated, which may itself contain certain notions of bias. See Supp. I for further discussion.

%%%%%%%%%%%%%%%%%%%%%
% Acknowledgement 
%%%%%%%%%%%%%%%%%%%%%
\section*{Acknowledgements}
This research is supported by the National Research Foundation, Singapore under its AI Singapore Programmes
(AISG Award No.: AISG2-TC-2022-007) and SUTD project PIE-SGP-AI-2018-01. 
This research work is also supported by the Agency for Science, Technology and Research (A*STAR) under its MTC Programmatic Funds (Grant No. M23L7b0021). 
This material is based on the research/work support in part by the Changi General Hospital and Singapore University of Technology and Design, under the HealthTech Innovation Fund (HTIF Award No. CGH-SUTD-2021-004).
We thank anonymous reviewers for their insightful feedback and discussion.

% These research projects are supported by the National Research Foundation, Singapore under its AI Singapore
% Programs (AISG Award No.: AISG2-RP-2021-021; AISG Award No.: AISG2-TC-2022-007).], and SUTD project PIE-SGP-AI-2018-01. We thank anonymous reviewers
% for their insightful comments.

{\small
\bibliographystyle{unsrtnat}
%\bibliography{Bib/eccv,Bib/ICLR23,Bib/ICML23,Bib/NeurIPS23}
\bibliography{Arvix_compiled}
}

\newpage
\appendix
%{\bf \huge Supplementary} 
%\\ \\
%Please find the respective code and Dataset in the following project page \url{https://sutd-visual-computing-group.github.io/CLEAM/}.
%anonymous Google drive links:
%\begin{itemize}
%    \item Code: \url{https://drive.google.com/drive/folders/1g0ChBXQla0wfn5MAGc2EfJq5Ek8gOCqH?usp=sharing}.
%    \item Dataset: \url{https://drive.google.com/drive/folders/1ENslNLyK6EEG2qj5YLZ3Qu3rFijJWEqB?usp=sharing}
%\end{itemize}
\section*{Supplementary Material}

This supplementary provides additional experiments as well as details that are required to reproduce our results. These were not included in the main paper due to space limitations. The supplementary is arranged as follows:

\begin{itemize}
    \item {\bf Section A}: Details on Modelling
    \begin{itemize}
        \item {\bf Section A.1} Details of Theoretical Modelling
        \item {\bf Section A.2} Additional Details on CLEAM Algorithm
        \item {\bf Section A.3} Details on Fairness Metric
        \item {\bf Section A.4} Details of Significance of the Baseline Errors
    \end{itemize}
    \item {\bf Section B}: Deeper Analysis on Error in Fairness Measurement
    \item {\bf Section C}: Validating Statistical Model for Classifier Output
        \begin{itemize}
        \item {\bf Section C.1} Validation of Sample-Based Estimate vs Model-Based Estimate
        \item {\bf Section C.2} Goodness-of-Fit Test: $\bphat$ from the Real GANs with Our Theoretical Model 
    \end{itemize}
    \item {\bf Section D}: Additional Experimental Results
        \begin{itemize}
            \item {\bf Section D.1} {Experimental Results with Standard Deviation}
            %RealGAN and Psuedo-GAN with Standard Deviation}
            %Interval Estimation Results of CLEAM for Bias Mitigation Algorithms 
            \item  {\bf Section D.2} Experimental Setup for Diversity
            \item {\bf Section D.3} Measuring Varying Degrees of Bias ($\texttt{Gender}$ and $\texttt{BlackHair}$)
            \item {\bf Section D.4} Measuring Varying Degrees of Bias with Additional Sensitive Attributes ($\texttt{Young}$ and $\texttt{Attractive}$)
            %\textcolor{red}{[Chris: AFHQ results are in the main paper, need to combine PE and IE table to follow main paper]}
            \item {\bf Section D.5} Measuring Varying Degrees of Bias with Additional Sensitive Attribute Classifiers (MobileNetV2) %\textcolor{green}{[+Experiment: AFHQ on MobileNet2]}
            \item {\bf Section D.6} Measuring SOTA GANs %\textcolor{blue}{
            and Diffusion Models
            %} 
            with Additional Classifier (CLIP) 
            \item {\bf Section D.7} Comparing Classifiers Accuracy on Validation Dataset vs Generated Dataset 
            %\textcolor{green}{[+Experiment: Repeat main papers Tab.2 experiment but with "generated dataset's $\alpha$"]}
            \item {\bf Section D.8} Comparing CLEAM with Classifier Correction Techniques (BBSE/BBSC)
            \item  {\bf Section D.9} Applying CLEAM to Re-evaluate Bias Mitigation Algorithms
        \end{itemize}
    \item {\bf Section E:} {Details on Applying CLIP as a SA Classifier}
    \item {\bf Section F:} Ablation Study: Details of Hyper-Parameter Settings and Selection %\textcolor{green}{[+Experiment: Include experiment with varying $s$] }
    \item {\bf Section G:} Related work 
    \item {\bf Section H:} Details of the GenData: A New Dataset of Labeled Generated Images 
    \item {\bf Section I:} Limitations and Considerations
        
\end{itemize}

\newpage
%\clearpage
\FloatBarrier
\section{Details on Modelling} 
\subsection{Details of Theoretical Modelling }
\label{subsec:expandedFormCLEAM}
In Sec 4.1 of main paper, we have proposed a statistical model for the sensitive attribute classifier output which is then used in CLEAM to rectify current measurement method. In this section, we give more details of this model which is not included in the main paper due to lack of space. 

Recall that in main paper, we mentioned that there are four possible mutually exclusive outputs $\bc$ for each sample with corresponding probability vector $\bp$:

\begin{equation*}
    \bc=
    \begin{bmatrix}
        c_{0|0}\\
        c_{1|0}\\
        c_{1|1}\\
        c_{0|1}\\     
    \end{bmatrix} ; \quad
    \bp=
    \begin{bmatrix}
        \pstar_0\alpha_0\\
        \pstar_0\alpha_0'\\
        \pstar_1\alpha_1\\
        \pstar_1\alpha_1'\\     
    \end{bmatrix}
\end{equation*}
where $c_{i|j}$ denotes the event of assigning label $i$ to a sample with GT label $j$.
Then, we mentioned that the count for each ouptut can be modeled as a multinomial distribution, $\mathbf{N}_{\bc}\sim Multi(n,\bp)$. 
Note that $\mathbf{N}_\bc 
= [N_{c_{0|0}},N_{c_{1|0}},N_{c_{1|1}},N_{c_{0|1}}]^T
$ 
is the random vector of counts for individual outputs of $\bc$.
% models the {\em joint probability distribution} of these outputs, \ie $ \mathbf{N}_\bc= \mathbb{P}(N_{c_{0|0}},N_{c_{1|0}},N_{c_{1|1}},N_{c_{0|1}})$ where, 
$N_{c_{i|j}}$ is the random variable of the count for event $c_{i|j}$ after classifying $n$ generated images.
First, we consider following assumptions:
\begin{enumerate}
    \item \textbf{Classifiers are reasonably accurate.} We state that, given the advancement in classifiers architecture, and the assumption that the sensitive attribute classifier is trained with proper training procedures, it is a reasonable assumption that it achieves reasonable accuracy and hence, $\alpha_0\neq0$ and $\alpha_1\neq0$. Similarly, we assume that it is highly unlikely to have a perfect classifier and as such $\alpha'_0=1-\alpha_0\neq0$ and $\alpha'_1=1-\alpha_1\neq0$.  
    \item \textbf{Generators are not completely biased.} Given that a generator is trained on a reliable dataset with the availability of all classes of a given sensitive attribute, coupled with the advancement in generator's architecture, it is a fair assumption that the generator would learn some representation of each class in the sensitive attribute and not be completely biased, as such $\pstar_0\neq0$ and $\pstar_1\neq0$.
\end{enumerate}

Based on this assumptions, $\bp$ is not near the boundary of the parameter space, and we can conclude that $0<\bp<1$. 
Therefore, we can approximate the multinomial distribution as a Gaussian,
$\mathbf{N}_\bc \sim \boldsymbol{\mathcal{N}}(\boldsymbol{\mu}, \boldsymbol{\Sigma})$, with $\boldsymbol{\mu}=n\bp$ and $\boldsymbol{\Sigma}= n\bM$~\cite{georgii2012stochastics}, where
%$N_{\bC}\sim\mathcal{N}(n\bp,n\bM)$, where
\begin{equation*}
    \boldsymbol{M}=
        \begin{bmatrix}
        \pstar_0\alpha_0 & 0 & 0 & 0\\
        0 & \pstar_0\alpha_0' & 0 & 0\\
        0 & 0 & \pstar_1\alpha_1 & 0\\
        0 & 0 & 0 &  \pstar_1\alpha_1'\\ 
    \end{bmatrix}
    -
        \begin{bmatrix}
        (\pstar_0\alpha_0)^2 & {(\pstar_0)}^2\alpha_0\alpha_0' & \pstar_0\pstar_1\alpha_0\alpha_1 & \pstar_0\pstar_1\alpha_0\alpha_1'
        \\
        {(\pstar_0})^2\alpha_0\alpha_0' & (\pstar_0\alpha_0')^2  & \pstar_0\pstar_1\alpha_0'\alpha_1 & \pstar_0\pstar_1\alpha_0'\alpha_1'
        \\
        \pstar_0\pstar_1\alpha_0\alpha_1 & \pstar_0\pstar_1\alpha_0'\alpha_1 & (\pstar_1\alpha_1)^2 & (\pstar_1)^2\alpha_1\alpha_1'
        \\
        \pstar_0\pstar_1\alpha_0\alpha_1' & \pstar_0\pstar_1\alpha_0'\alpha_1' & (\pstar_1)^2\alpha_1\alpha_1' & (\pstar_1\alpha_1')^2\\ 
    \end{bmatrix}
\end{equation*}
and therefore:
\begin{equation*}
    \boldsymbol{\mu}=
    \begin{bmatrix}
    \boldsymbol{\mu}_1\\
    \boldsymbol{\mu}_2\\
    \boldsymbol{\mu}_3\\
    \boldsymbol{\mu}_4\\     
    \end{bmatrix}
    =
    n\begin{bmatrix}
        \pstar_0\alpha_0\\
        \pstar_0\alpha_0'\\
        \pstar_1\alpha_1\\
        \pstar_1\alpha_1'\\     
    \end{bmatrix}
\end{equation*}
\begin{equation*}
%\begin{split}
\resizebox{\textwidth}{!}{
$
    \boldsymbol{\Sigma}
    =
        \begin{bmatrix}
        \boldsymbol{\Sigma}_{11} & \boldsymbol{\Sigma}_{12} & \boldsymbol{\Sigma}_{13} & \boldsymbol{\Sigma}_{14}\\
        \boldsymbol{\Sigma}_{21} & \boldsymbol{\Sigma}_{22} & \boldsymbol{\Sigma}_{23} & \boldsymbol{\Sigma}_{24}\\
        \boldsymbol{\Sigma}_{31} & \boldsymbol{\Sigma}_{32} & \boldsymbol{\Sigma}_{33} & \boldsymbol{\Sigma}_{34}\\
         \boldsymbol{\Sigma}_{41} & \boldsymbol{\Sigma}_{42} & \boldsymbol{\Sigma}_{43} &  \boldsymbol{\Sigma}_{44}\\ 
    \end{bmatrix}
    \\
    =
        n\begin{bmatrix}
        \pstar_0\alpha_0-(\pstar_0\alpha_0)^2 & {(\pstar_0)}^2\alpha_0\alpha_0' & \pstar_0\pstar_1\alpha_0\alpha_1 & \pstar_0\pstar_1\alpha_0\alpha_1'
        \\
        {(\pstar_0})^2\alpha_0\alpha_0' & \pstar_0\alpha_0'-(\pstar_0\alpha_0')^2  & \pstar_0\pstar_1\alpha_0'\alpha_1 & \pstar_0\pstar_1\alpha_0'\alpha_1'
        \\
        \pstar_0\pstar_1\alpha_0\alpha_1 & \pstar_0\pstar_1\alpha_0'\alpha_1 & \pstar_1\alpha_1-(\pstar_1\alpha_1)^2 & (\pstar_1)^2\alpha_1\alpha_1'
        \\
        \pstar_0\pstar_1\alpha_0\alpha_1' & \pstar_0\pstar_1\alpha_0'\alpha_1' & (\pstar_1)^2\alpha_1\alpha_1' & \pstar_1\alpha_1'-(\pstar_1\alpha_1')^2\\ 
    \end{bmatrix}
$
}  
%\end{split}
\end{equation*}
With this, we note that the {\em marginal distribution} of this multivariate Gaussian distribution gives us a univariate (one-dimensional) Gaussian distribution for the count of each output in $\bc^T$. For example, the distribution of the count for event $c_{0|0}$, denoted by $N_{c_{0|0}}$, can be modeled as $N_{c_{0|0}} \sim \mathcal{N}(\boldsymbol{\mu}_1, \boldsymbol{\Sigma}_{11})$.
Then, we find the total rate of data points labeled as class $i$ when labeling $n$ generated images using the normalized sum of the related random variables \ie $\hat{p}_i=\frac{1}{n}\sum_j N_{{c}_{i|j}}$. More specifically:
%With that, we solve for the total rate of data points labelled as class $i$ when labelling $n$ generated samples using the normalized sum of the related random variable \ie 
%We then normalise $\textbf{N}_{\phat_i}$, to get $\phat_i=\frac{1}{n}\textbf{N}_{\phat_i},\;i\in\{0,1\}$
%via
%the marginal distribution of $\mathbf{N}_{\phat_0}$ and $\mathbf{N}_{\phat_1}$ where, 
\begin{gather*}
%\begin{equation*}
%    \begin{aligned}
    \phat_0=\frac{1}{n}(\mathbf{N}_{\bc_{0|0}}+\mathbf{N}_{\bc_{0|1}})\sim 
    \boldsymbol{\mathcal{N}}(\Tilde{\mu}_{\phat_0},\Tilde{\sigma}^2_{\phat_0})
    \\
    \Tilde{\mu}_{\phat_0}=
    %(\boldsymbol{\mu}_1 + \boldsymbol{\mu}_4)/n=
    (\pstar_0\alpha_0+\pstar_1\alpha_1')
    \\
    \Tilde{\sigma}^2_{\phat_0}=
    %(\boldsymbol{\Sigma_{11}}+\boldsymbol{\Sigma_{44}}+\boldsymbol{\Sigma_{14}})/n^2
    \frac{1}{n}((\pstar_0\alpha_0-(\pstar_0\alpha_0)^2)+(\pstar_1\alpha_1'-(\pstar_1\alpha_1')^2)
    + 2 {\pstar_0\pstar_1\alpha_0\alpha_1'})
    %\sqrt{n(\pstar_0\alpha_0-(\pstar_0\alpha_0)^2)}\sqrt{n(\pstar_1\alpha_1'-(\pstar_1\alpha_1')^2})
%    \end{aligned}
%\end{equation*}
\end{gather*}
\begin{gather*}
%\begin{equation*}
%    \begin{aligned}
    {\phat_1}=\frac{1}{n}(\mathbf{N}_{\bc_{1|0}}+\mathbf{N}_{\bc_{1|1}})\sim \mathcal{N}(\Tilde{\mu}_{\phat_1},\Tilde{\sigma}^2_{\phat_1})
    \\
    \Tilde{\mu}_{\phat_1}=(\pstar_0\alpha_0'+\pstar_1\alpha_1)\\
    \Tilde{\sigma}^2_{\phat_1}=\frac{1}{n}((\pstar_0\alpha_0'-(\pstar_0\alpha_0')^2)+(\pstar_1\alpha_1-(\pstar_1\alpha_1)^2)
    +2{\pstar_0\pstar_1\alpha_0'\alpha_1})
    %\sqrt{n(\pstar_0\alpha_0'-(\pstar_0\alpha_0')^2)}\sqrt{n(\pstar_1\alpha_1+(\pstar_1\alpha_1)^2)}
%    \end{aligned}
%\end{equation*}
\end{gather*}

%\textcolor{red}{To conclude \dots *Main take away*}
{\bf Remark:} In Sec 4.1 of the main paper, considering the probability tree diagram in Fig.1(b) (main paper), we proposed a  distribution for the possible events of classification ($c_{i|j}$), and used it to compute distribution of each event, and finally the distribution of the output of the sensitive attribute classifier ($\phat_0$, and $\phat_1$). 
Here, we provide more information on the necessary assumptions and the expanded forms of the equations.
In the following Sec. \ref{apx:proposedmethod}, we will similarly provide more information on proposed CLEAM, presented in Sec. 4.2 of the main paper, which utilizes this statistical model to mitigate the sensitive attribute classifier's error.

\FloatBarrier
\subsection{Additional Details on CLEAM Algorithm}
\label{apx:proposedmethod}

{\bf MLE value of Population Mean.}
In this section, first, we discuss the  maximum likelihood estimate (MLE) of the population mean for a Gaussian distribution. 
Given a Gaussian distribution with the population mean $\tilde{\mu}_{\phat_0}$ and standard deviation $\tilde{\sigma}_{\phat_0}$, we can first find the joint probability distribution from the product of each probabilistic outcome (we introduce the natural log as a monotonic function, for ease of calculation).
Then, to find the MLE of $\tilde{\mu}_{\phat_0}$, we take the partial derivative of this joint distribution \wrt $\tilde{\mu}_{\phat_0}$, and solve for its maximum value. This maximum value is equal to the sample mean, $\Ddot{\mu}_{\phat_0}$, as detailed below:
\begin{gather*}
%\begin{equation*}
    \label{eqn:MLEAppen}
%    \begin{aligned}
    \frac{\partial}{\partial \Tilde{\mu}_{\phat_0}} \prod_{i=1}^s ln(\frac{1}{\Tilde{\sigma}_{\phat_0} \sqrt{2 \pi}} e^{\frac{-(\hat{p}^i_0 - \Tilde{\mu}_{\phat_0})^2}{2 \Tilde{\sigma}_{\phat_0}^2}})=0\\
    \frac{1}{\Tilde{\sigma}_{\phat_0}^2} \sum_{i=0}^s (\hat{p}^i_0-\Tilde{\mu}_{\phat_0})=0\\ 
    \Tilde{\mu}_{\phat_0}=\frac{1}{s} \sum_i^s \hat{p}^i_0=\Ddot{\mu}_{\phat_0}
%    \end{aligned}
%\end{equation*}
\end{gather*}

{\bf Point Estimate of CLEAM.}
From this, given that $s$ is sufficiently large, we utilize the sample mean as the maximum likelihood approximate of the population mean. As the population mean was modeled in Sec. \ref{subsec:expandedFormCLEAM}, we can equate the sample mean to the expanded theoretical model: 
%From this, given that $s$ is large, we assume that the sample mean is a good approximation for the population mean (modeled in Sec. \ref{subsec:expandedFormCLEAM} ) and equate them 
%\textcolor{red}{\bf we have to provide more information as to how we got to the RHS of the equation}:
\begin{gather*}
\Ddot{\mu}_{\phat_0}=\Tilde{\mu}_{\phat_0}=p^*_0\alpha_{0} + (1-p^*_0)\alpha'_{1}
\end{gather*}
Now given that the classifier's accuracy $\balpha=[\alpha_0,\alpha_1]$ and the sample mean $\Ddot{\mu}_{\phat_0}$ can be measured, we are able to solve for the maximum likelihood point estimate of $\pstar_0$, which we denoted with $\mu_{\texttt{CLEAM}}$ as follows:
\begin{gather*}
%\begin{equation*}
%    \begin{aligned}
    \mu_{\texttt{CLEAM}}=\frac{\Ddot{\mu}_{\phat_0}-\alpha'_{1}}{\alpha_0-\alpha'_{1}}=\frac{\Ddot{\mu}_{\phat_0}-1+\alpha_1}{\alpha_0-1+\alpha_1} 
    %\\
    %\mu_{\texttt{CLEAM}}(p^*_1)=1-\mu_{\texttt{CLEAM}}(p^*_0)
%    \end{aligned}
%\end{equation*}
\end{gather*}
Note that we compute $\mu_\texttt{CLEAM}$ \wrt $\pstar_0$ \ie $\mu_\texttt{CLEAM}(\pstar_0)$  through-out this paper for ease of discussion, however as $\pstar_1=1-\pstar_0$, a similar $\mu_\texttt{CLEAM}$ \wrt $\pstar_1$ \ie $\mu_\texttt{CLEAM}(\pstar_1)$ can be found with $\mu_\texttt{CLEAM}(\pstar_1)=1-\mu_\texttt{CLEAM}(\pstar_0)$.

{\bf Interval Estimate of CLEAM.}
We acknowledge that there exist other statistically probable solutions for $\bpstar$ that could output the $s$ $\bphat$ samples, other than the Maximum likelihood point estimate of $\bpstar$. We thus propose the following approximation for the $95\%$ confidence interval of $\bpstar$.
Recall the notations $\Ddot{\mu}_{\phat_0}$ and $\Ddot{\sigma}_{\phat_0}$ are the sample mean and standard deviation respectively:
\begin{equation*}
    \Ddot{\mu}_{\phat_0}=\frac{1}{s} \sum_i^s \hat{p}^i_0
    \quad ; \quad
    \Ddot{\sigma}_{\phat_0}=\sqrt{\frac{\sum^s_{i=1}(\phat^i_0-\Ddot{\mu}_{\phat_0})^2}{s-1}}
\end{equation*}

Since $\bphat$ follows a Gaussian distribution, we can propose the following equation:
\begin{equation*}
    \begin{aligned}
    Pr(-z_{\frac{\delta}{2}}\leq \frac{\Ddot{\mu}_{\phat_0}-\Tilde{\mu}_{\phat_0}}{\frac{\Ddot{{\sigma}}_{\phat_0}}{\sqrt{s}}}\leq \z_{\frac{\delta}{2}})=1-\delta
    \quad 
    \end{aligned}
\end{equation*}
Solving for $\Tilde{\mu}_{\phat}$, we get:
%\begin{equation*}
%    \begin{aligned}
%    Pr(-z_{\frac{\delta}{2}}\leq \frac{\Ddot{\mu}_{\phat_0}-\Tilde{\mu}_{\phat_0}}{\frac{\Ddot{\sigma}_{\phat_0}}{\sqrt{s}}}\leq \z_{\frac{\delta}{2}})=1-\delta
%    \end{aligned}
%\end{equation*}
%\textcolor{red}{\bf [Milad: The upper equation is repetitive?][Yes, noted thanks]}
\begin{equation*}
    \begin{aligned}
    Pr(\Ddot{\mu}_{\phat_0}+z_\frac{\delta}{2}(\frac{\Ddot{\sigma}_{\phat_0}}{\sqrt{s}})\geq \Tilde{\mu}_{\phat_0} \geq \Ddot{\mu}_{\phat_0}- z_{\frac{\alpha}{2}}\frac{\Ddot{\sigma}_{\phat_0}}{\sqrt{s}})=1-\delta
    \end{aligned}
\end{equation*}
Then, given that $\Tilde{\mu}_{\phat}=\pstar_0\alpha_0+\pstar_1\alpha_1'=\pstar_0(\alpha_0-\alpha_1')+\alpha_1'$ we formulate the following:
\begin{equation}
    \label{eqn:CIwithsamplestd}
    \begin{aligned}
    Pr(\frac{\Ddot{\mu}_{\phat_0}+z_\frac{\delta}{2}(\frac{\Ddot{\sigma}_{\phat_0}}{\sqrt{s}})-\alpha_1'}{\alpha_0-\alpha_1'}
    \geq 
    \pstar_0
    \geq 
    \frac{\Ddot{\mu}_{\phat_0}- z_{\frac{\alpha}{2}}\frac{\Ddot{\sigma}_{\phat_0}}{\sqrt{s}}-\alpha_1'}{\alpha_0-\alpha_1'})=1-\delta
    \end{aligned}
\end{equation}
As such when $\delta=0.05$, we can determine that the $95\%$ approximated confidence interval of $\pstar_0$ is :
\begin{equation*}
    \begin{aligned}
    \rho_{\texttt{CLEAM}}(\pstar_0)=[\mathcal{L}(\pstar_0),\mathcal{U}(\pstar_0)]=
    [\frac{\Ddot{\mu}_{\phat_0}-1.96(\frac{\Ddot{\sigma}_{\phat_0}}{\sqrt{s}})-\alpha_1'}{\alpha_0-\alpha_1'}
    \quad,\quad
    \frac{\Ddot{\mu}_{\phat_0}+ 1.96\frac{\Ddot{\sigma}_{\phat_0}}{\sqrt{s}}-\alpha_1'}{\alpha_0-\alpha_1'}]
    \end{aligned}
\end{equation*}

\iffalse
\textbf{Alternative solution:} 
As an alternative solution to Eqn. \ref{eqn:CIwithsamplestd}, 
the theoretical standard deviation i.e., $\Tilde{\sigma}_{\phat_0}$, 
could be used, as shown in Eqn. \ref{eqn:CIwithpopstd}. 
But as we will discuss in Sec. \ref{subsec:sampleBasedEstimate}, the sample-based estimates and the model-based estimates are very close, and the sample-based estimates can be easily calculated. Therefore,  
we apply the sample-based standard deviation as discussed above.

\begin{equation}
    \label{eqn:CIwithpopstd}
    \begin{aligned}
    Pr(-z_{\frac{\delta}{2}}\leq \frac{\Ddot{\mu}_{\phat_0}-(\pstar_0(\alpha_0-\alpha_1')+\alpha_1')}{\sqrt{\frac{1}{n}(\pstar_0\alpha_0-(\pstar_0\alpha_0)^2)+\frac{1}{n}(\pstar_1\alpha_1'-(\pstar_1\alpha_1')^2)
    + \frac{2}{n} {\pstar_0\pstar_1\alpha_0\alpha_1'}}}\leq z_{\frac{\delta}{2}})=1-\delta
    \end{aligned}
\end{equation}
\fi

%\textcolor{blue}{
{\bf Extending the point estimate to a multiple label setup.}
We remark that in current literature, fairness of generative models has been studied for binary sensitive attributes mainly due to the lack of availability of a large labeled dataset needed for systematic experimentation. As a result, CLEAM similarly focuses on binary SA to address a common flaw in the evaluation process of the many proposed State-of-the-Art methods.
%}

\begin{figure*}[h!]
    \centering
    \includegraphics[width=\textwidth]{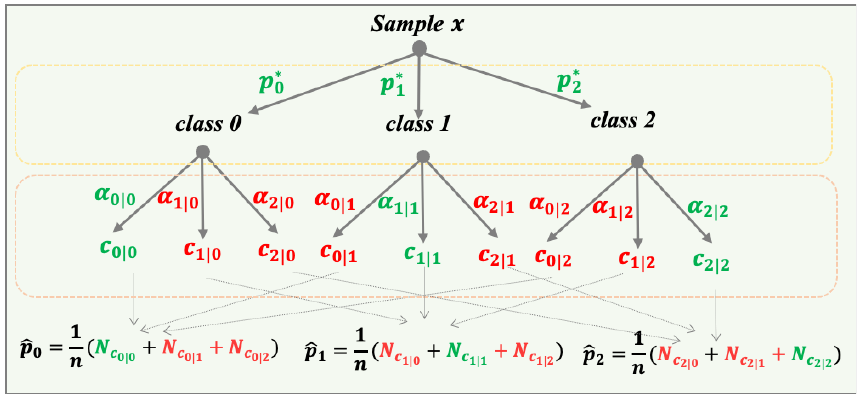}
    \caption{
  %  \textcolor{blue}{
    Our statistical model for fairness measurement when considering a multi-label SA. For illustration purposes, we utilize 3 labels for a given SA. Note that, {\bf our same approach can be applied to other multi-label settings.} This statistical model accounts for inaccuracies in the SA classifier and is the base of our proposed CLEAM (see Sec. 4.1). Here, $p^*_j$ is the ground truth probability of a generator outputting a sample with label $j$ and ${\alpha}_{i|j}$ denotes the probability (“accuracy”) of the SA classifier classifying a sample with GT label $j$ as $i$ for $i,j \in \{0,1,2\}$.
    %\textcolor{red}{[Milad: This figure seems a bit blurred.]}
    %Chris: I believe this is the overleaf rendering (seems fine when downloaded), I've replaced it anyways to make sure.
    } 
  %  }
    \label{fig:multilabel}
\end{figure*}

%\textcolor{blue}{
Assuming that the constraint of the dataset is addressed, our same CLEAM approach can be easily extended to a multi-label setup. For example, given a 3 label sensitive attribute where $\pstar_j$ is the probability of generating a sample with label $j$ and $\alpha_{i|j}$ denotes the probability (“accuracy”) of the SA classifier in classifying a sample with GT label $j$ as $i$ for $i$,$j\in\{0,1,2\}$. Fig. \ref{fig:multilabel} shows our statistical model for this setting. We can then similarly solve for the point estimate by solving the matrix:
\begin{equation}
    \begin{bmatrix}
        \alpha_{0|0} & \alpha_{0|1} & \alpha_{0|2}\\
        \alpha_{1|0} & \alpha_{1|1} & \alpha_{1|2}\\
        \alpha_{2|0} & \alpha_{2|1} & \alpha_{2|2}\\
        \end{bmatrix}
        \begin{bmatrix}
            p^*_0 \\
            p^*_1 \\
            p^*_2
        \end{bmatrix}
        = 
        \begin{bmatrix}
            \ddot{\mu}_{\hat{p}_0} \\
            \ddot{\mu}_{\hat{p}_1} \\
            \ddot{\mu}_{\hat{p}_2}
        \end{bmatrix}
\end{equation}
%}

\subsection{Details on Fairness Metric}
\label{subsec:MoreOnFD}
Fairness in generative models is defined as {\em Equal Representation} meaning that the generator is supposed to generate an equal number of samples for each element of an attribute, e.g., an equal number of generated \texttt{Male} and \texttt{Female} samples when the sensitive attribute is \texttt{Gender}. 
Therefore, the expected distribution for a fair generator is usually a uniform distribution denoted by $\bpbar$. Considering this, the fairness discrepancy (FD) metric \cite{choiFairGenerativeModeling2020a} measures the L2 norm between $\bpbar$ and the estimated 
%SA distribution
class probability of the generator 
by %the sensitive attribute classifier $C_\bu$
each measurement method, \ie $\boldsymbol{\mu}_{\dagger}$, where $\dagger\in$\{\texttt{Base}, \texttt{CLEAM}, \texttt{Div}\}, as follows:
\begin{equation}
f  
= 
%|\bar{p} - \mathbb{E}_{z\sim p_{z}(z)} [C_{\bu}(\mathbf{G(z)})] |_2 = 
|\bpbar - \boldsymbol{\mu}_{\dagger}|_2
    \label{eqn:FD}
\end{equation}
%where $C_\bu(G(z))$ is the one-hot vector for the classified label of the generated sample, $G(z)$. $z$ is sampled from a Gaussian noise distribution $p_z(z)$. 
Note that for a fair generator the fairness discrepancy $f$ would be zero, which also indicates zero bias.

%\textcolor{red}{\bf [Milad: $\boldsymbol{\mu}_{\dagger}$ should be bold as we use a vector?] Yes I was just correcting that haha}

\subsection{Details of Significance of the Baseline Errors}
%\textcolor{red}{[Chris: Discuss relative improvement vs relative error, provide an example] [{\bf Addressed}]}

%\textcolor{blue}{

In the main manuscript
(Sec. 3 of the main paper), we discussed that the relative improvement of 
the previous fair generative models could be small, e.g. 
Teo \etal \cite{teo2022fair} and  Um \etal \cite{umFairGenerativeModel2021} 
report a relative improvement in the fairness of
0.32\% and 0.75\%, compared to imp-weighting \cite{choiFairGenerativeModeling2020a}, and they  
fall within the range of our experiment's smallest relative error, $\emubase$=4.98\%.
Here, we provide more detail on how we calculate the relative improvements in the main manuscript. 
Specifically, we calculate the relative change of the proposed work against the previous work with the following:
%}

\begin{align}
    Relaitve\;Improvement=\frac{|(\phat_0\;of\;previous\;work)-(\phat_0 \; of \; proposed \; work)|}{(\phat_0\;of\;previous\;work)}
    \label{eqn:relativeimprovement}
\end{align}

%\textcolor{blue}{
Notice that this is similar to $\emubase$ of Eqn. 1 in the main paper. For example, Teo \etal (Tab. 1 90\_10 and perc=0.1 settings) \cite{teo2022fair}
reports that fairTL measures a $f=0.105$ which is compared against the previous work's (Choi \etal \cite{choiFairGenerativeModeling2020a}) $f=0.107$. 
Utilizing Eqn. \ref{eqn:FD}, we find that this is equivalent to $\pstar_0=$ 0.4257 or 0.5743, and 0.4243 or 0.5757, respectively. We remark that here we report two values per $f$, as the FD metric is a symmetric metric. Then applying Eqn. \ref{eqn:relativeimprovement}, and taking the maximum of the values, we find the relative improvement to be $0.32\%$, at best. Note that as we mentioned in the main paper for this setup the baseline measurement framework results in $4.98\%$ error rate (with the best performing sensitive attribute classifier), meaning that it may not be reliable for gauging the improvement.
%}

%\textcolor{red}{\bf [Milad: Hi Chris. I added a sentence above. Please fill in the XX value.]}
\newpage
\FloatBarrier
\section{Deeper Analysis on Error in Fairness Measurement}
\label{sec:DeeperAnlaysis}
%\textcolor{red}{
%[Chris: To include:]
%\begin{enumerate}
%    \item Why we didn't include diversity in the main paper [{\bf Addressed}]
%    \item Large error despite large error in R34 [{\bf Addressed}]
%\end{enumerate}
%}

In the main paper Sec.3, we discussed that there could be considerable error in the fairness measurement, $\bphat$, even though the sensitive attribute classifier's accuracy is considerably high. In addition, we further develop on this and discuss two additional factors that could result in a variation of errors.
%\textcolor{blue}{
{\em We remark that in the main manuscript, we report diversity only using VGG-16, as specified by Keswani \etal \cite{keswaniAuditingDiversityUsing2021}. 
%This is because Diversity specifies the use of a pre-trained VGG-16 on ImageNet as a good feature extractor. 
Further discussion in Sec. \ref{sec:DiversitySetup} %}
}

{\bfseries\boldmath Accuracy of Individual Classes} ($\balpha=\{\alpha_0, \alpha_1\}$) {\bfseries\boldmath Impacts the Degree of Error}.
Notice that in some cases even though the sensitive attribute classifier may have a very similar average accuracy, different degrees of errors could exist for the two different classifiers  \eg R18 and R34 in Tab. \ref{tab:G_PE_Extract}. This is because {\em the fairness measurement error is not only dependent on the average accuracy but on the individual class accuracy} \ie $\alpha_0$ and $\alpha_1$. More specifically, given that there is a 
%lower
%\textcolor{blue}{
larger
%}
error in $\alpha_0$ for R34 and the bias exists in $\pstar_0=0.643$, this results in a compounded effect and hence a larger error of $\emubase$=11.98\% is observed as compared to R18 $\emubase$=6.84\%. 

{\bfseries\boldmath Uniform Inaccuracies at Unbiased Test-Point} ($\pstar_0=\pstar_1=$0.5).
%Notice that as discussed in Sec. 5.3 of the main paper,
%\textcolor{blue}{
In our extended experiments in Sec. \ref{sec:psuedoG} for a Pseudo-generator, we discuss that
%}
%in Tab. \ref{tab:fakeG_PE_Exrtacted} for a Pseudo-generator setup, 
for some sensitive attribute classifiers \eg ResNet-18 for \texttt{Gender} and \texttt{BlackHair}, the Baseline performs better than CLEAM at the unbiased test-point \ie $\pstar_0=0.5$. 
This is just due to the \texttt{Gender} and \texttt{Blackhair} setups having a specific combination of (i) the Pseudo-Generator producing almost perfectly unbiased data with $\bpstar=[0.5,0.5]$, (ii) sensitive attribute classifier with almost perfectly uniform inaccuracies $\alpha_0' \approx \alpha_1'$, thereby leading to {\em uniform misclassification} and 
%thereby giving 
hence
the false impression of better accuracy by the baseline method, at $\bpstar=[0.5,0.5]$ (See Tab. \ref{tab:fakeG_PE_Exrtacted} for extracted table) . To further illustrate this, notice how the ResNet-18 trained on \texttt{Cat/Dog} did not demonstrate this better performance in the Baseline due to its non-uniform $\balpha$. Nevertheless, we note this situation whereby the Baseline outperform CLEAM is specific to the test-point $\pstar_0=0.5$ and does not impact the overall effectiveness of CLEAM. Furthermore, CLEAM still demonstrates outstanding results with low error for both the PE and IE at $\pstar_0=0.5$.

To further demonstrate these effects, we repeat this same experiment,
%as Sec. 5.3 in the main paper
but with sensitive attributes \texttt{Young} and \texttt{Attractive} from the CelebA dataset. As seen in Tab. \ref{tab:skewAblation}, both $\texttt{Young}$ or $\texttt{Attractive}$ have similar average accuracy, $\alpha_{Avg}=\frac{\alpha_0+\alpha_1}{2}$ of $0.801$ and $0.794$ but a different $skew_\balpha=|\alpha_0-\alpha_1|$ of 0.103 and 0.027. As such, we are able to investigate the effects that $skew_\balpha$ has on both CLEAM and Baseline. We did not include Diversity in this study, due to its poor performance on harder sensitive attribute, as discussed in Sec. \ref{sec:DiversitySetup}.
\textbf{\em From our results in Tab. \ref{tab:skewAblation}}, we observe that as the $skew_\balpha$ increases from sensitive attribute $\texttt{Attractive}$ to $\texttt{Young}$, the error becomes much more significant in the baseline method. The average $\emubase$ increases from $12.69\%$ to $17.63\%$. 
Furthermore, 
%we observe that $skew_\balpha$ has influence on the errors observed in the lower biases ($\pstar=[0.5,0.5]$). 
unlike $\texttt{Gender}$ and $\texttt{Blackhair}$, who have relatively negligible skew, $\texttt{Young}$ and $\texttt{Attractive}$ observes a significantly larger error at $\bpstar=[0.5,0.5]$.

\begin{table*}[!h]
    \centering
    %New overall Caption
    \caption{
    {\bf Extracted from Tab. \ref{tab:fakeG_PE} for ease of viewing.}
    %Tab. 2  of main paper.}
    Comparing the {\em point estimates} and {\em interval estimates} of Baseline \cite{choiFairGenerativeModeling2020a} and our proposed CLEAM measurement framework in estimating $\bpstar$ of the GenData datasets sampled from (A) StyleGAN2 \cite{karrasStyleBasedGeneratorArchitecture2019}. 
    The $\pstar_0$ value for each GAN with a certain attribute is determined by manually hand-labeling the generated data.
    We utilize two different classifiers Resnet-18/34 (R18, R34)\cite{heDeepResidualLearning2016} with different accuracy $\balpha$ to obtain $\bphat$ by classifying samples \wrt \texttt{BlackHair}.
    For calculating each $\bphat$, we utilize $n=400$ samples and evaluate for a batch size of $s=30$.
    We repeat this for 5 experimental runs and report the mean error rate, per Eqn. 1 of the main paper. 
   }
\resizebox{\textwidth}{!}{
    \addtolength{\tabcolsep}{-4pt}
      \begin{tabular}
      {c@{\hspace{-0.01cm}}c c@{\hspace{-0.01cm}} cc cc cc @{\hspace{0.7cm}} cc cc cc}
        \toprule
        & & & \multicolumn{6}{c}{\bf Point Estimate} & \multicolumn{6}{c}{\bf Interval Estimate}
        \\
        \cmidrule(lr){4-9}\cmidrule(lr){10-15}
          \textbf{Classifier} &
          \textbf{  $\balpha=\{\alpha_0,\alpha_1\}$ } &
          Avg. $\balpha$ &
          %\textbf{  $\balpha=\{\alpha_0,\alpha_1\}$ } &
          \multicolumn{2}{c}{\textbf{Baseline}
          %\cite{choiFairGenerativeModeling2020a}
          }& \multicolumn{2}{c}{\textbf{Diversity}
          %\cite{keswaniAuditingDiversityUsing2021}
          } & \multicolumn{2}{c}{\textbf{CLEAM (Ours)}}
          %Interval Estimates
            & \multicolumn{2}{c}{\textbf{Baseline}
            %\cite{choiFairGenerativeModeling2020a}
            }& \multicolumn{2}{c}{\textbf{Diversity}
            %\cite{keswaniAuditingDiversityUsing2021}
            } & \multicolumn{2}{c}{\textbf{CLEAM (Ours)}}
        \\
        \cmidrule(lr){1-1}\cmidrule(lr){2-2}\cmidrule(lr){3-3}\cmidrule(lr){4-5}\cmidrule(lr){6-7} \cmidrule(lr){8-9}
        \cmidrule(lr){10-11}\cmidrule(lr){12-13} \cmidrule(lr){14-15}
         %Baseline
        & & &
        $\mu_{\texttt{Base}}$ & $e_{\mu}(\downarrow)$ 
         %Diversity
         &$\mu_{\texttt{Div}}$ & $e_\mu(\downarrow)$
         %CLEAM
        &$\mu_{\texttt{CLEAM}}$ & $e_\mu(\downarrow)$
        &$\rho_{\texttt{Base}}$ & $e_\rho(\downarrow)$ 
        &$\rho_{\texttt{Div}}$ & $e_\rho(\downarrow)$ 
        &$\rho_{\texttt{CLEAM}}$ & $e_\rho(\downarrow)$\\
        \midrule
        \multicolumn{15}{c}{\bf (A) StyleGAN2}\\
        \midrule
    %BlackHair
        \multicolumn{14}{c}{\texttt{BlackHair} with GT class probability {\bfseries\boldmath $p^*_0$=0.643}}\\
        \midrule
        %Point Estimate----------------------------------------------------------
        R18 & \{0.869, 0.885\} & 0.88 & 
        0.599 & 6.84\% & 
        %Diversity
        --- & --- &  
        %CLEAM
        0.641 & \textbf{0.31\%} &
        %Interval Estimate----------------------------------------------------------
        [0.591, 0.607] & 8.08\% & 
        %Diversity
        --- & --- & 
        %CLEAM
        [0.631, 0.652] & \textbf{1.40\%}
        \\  
        %\hdashline
        %Point Estimate----------------------------------------------------------
        R34 & \{0.834, 0.916\}  & 0.88 & 
        0.566 & 11.98\%&
        %Diversity
        --- & --- & 
        %CLEAM
        0.644 & 
        \textbf{0.16}\% &
        %Interval Estimate----------------------------------------------------------
        [0.561, 0.572] & 12.75\% &
        %Diversity
        --- & --- &
        %CLEAM
        [0.637, 0.651] & \textbf{1.24\%} \\
        \bottomrule
    \end{tabular}
    }
    \label{tab:G_PE_Extract}
    \addtolength{\tabcolsep}{4pt}
\end{table*}

\begin{table*}[h]
    \centering
    %New overall Caption
    \caption{ {\bf Extracted from Tab. \ref{tab:fakeG_PE} for ease of viewing.}
    Comparing the {\em point estimates} and {\em interval estimate}  of Baseline \cite{choiFairGenerativeModeling2020a}, Diversity \cite{keswaniAuditingDiversityUsing2021} and proposed CLEAM measurement frameworks in estimating different $\bpstar$ of a {\em pseudo-generator}, based on the CelebA \cite{liuDeepLearningFace2015} and AFHQ \cite{choiStarGANV2Diverse2020} dataset. The $\bphat$ is computed with a ResNet-18 sensitive attribute classifier and
    the error rate is reported using Eqn. 1 of the main paper.
    We repeat this for \texttt{Gender}, \texttt{BlackHair} and \texttt{Cat/Dog} attributes.
    }
\resizebox{0.98\linewidth}{!}{
    \addtolength{\tabcolsep}{-2pt}
      \begin{tabular}
      %{c  c @{\hspace{0.5em}} c @{\hspace{1em}} c @{\hspace{0.5em}} c @{\hspace{1em}} c @{\hspace{0.5em}} c}
      {c  cc cc cc @{\hspace{1cm}} cc cc cc}
        \toprule
        %\multicolumn{7}{c}{\bf (A) Pseudo-Generator}\\
        %\midrule
        & \multicolumn{6}{c}{\bf Point Estimate} & \multicolumn{6}{c}{\bf Interval Estimate}\\
        \cmidrule(lr){2-7}\cmidrule(lr){8-13}
          \textbf{GT} & \multicolumn{2}{c}{\textbf{Baseline}
          %\cite{choiFairGenerativeModeling2020a}
          }& \multicolumn{2}{c}{\textbf{Diversity}
          %\cite{keswaniAuditingDiversityUsing2021}
          } & \multicolumn{2}{c}{\textbf{CLEAM (Ours)}}
          & \multicolumn{2}{c}{\textbf{Baseline}
          %\cite{choiFairGenerativeModeling2020a}
          }& \multicolumn{2}{c}{\textbf{Diversity}
          %\cite{keswaniAuditingDiversityUsing2021}
          } & \multicolumn{2}{c}{\textbf{CLEAM (Ours)}}
        \\
         \cmidrule(lr){1-1}\cmidrule(lr){2-3}\cmidrule(lr){4-5} \cmidrule(lr){6-7} 
         \cmidrule(lr){1-1}\cmidrule(lr){8-9}\cmidrule(lr){10-11} \cmidrule(lr){12-13}
         %Baseline
        &$\mu_{\texttt{Base}}$ & $e_\mu (\downarrow)$ 
         %Diversity
         &$\mu_{\texttt{Div}}$ & $e_\mu(\downarrow)$ 
         %CLEAM
        &$\mu_{\texttt{CLEAM}}$ & $e_\mu(\downarrow)$
         %Baseline
        &$\rho_{\texttt{Base}}$ & $e_\rho (\downarrow)$ 
         %Diversity
         &$\rho_{\texttt{Div}}$ & $e_\rho(\downarrow)$ 
         %CLEAM
        &$\rho_{\texttt{CLEAM}}$ & $e_\rho(\downarrow)$
        \\
        \midrule
        \multicolumn{13}{c}{$\balpha$=[0.976,0.979], \texttt{Gender} (CelebA)}\\
        \midrule
        $p^*_0$=0.5 & 
        %Point Estimate--------------------------------------------------
        0.501 & \textbf{0.20}\% &
        %Diversity
        0.481 & 3.80\% & 
        %CLEAM
        0.502  & 0.40\%  &
        %Interval Estimate--------------------------------------------------
        [0.495  , 0.507 ] & \textbf{1.40}\% &
        %Diversity
        [0.473 , 0.490 ] & 5.40\% &
        %CLEAM
        [0.497, 0.508]  & 1.60\%
        \\
        \midrule
        \multicolumn{13}{c}{$\balpha$=[0.881,0.887], \texttt{BlackHair} (CelebA)}\\
        \midrule
        %Point Estimate--------------------------------------------------
        $p^*_0$=0.5 & 
        0.500 & \textbf{0.00}\% & 
        %Diversity
        0.521 & 4.20\% & 
        %CLEAM
        0.504  & 0.8\%  &
        %Interval Estimate--------------------------------------------------
        [ 0.495  , 0.505  ] & \textbf{1.00}\% &
        %Diversity
        [0.506 , 0.536 ] & 7.20\% &
        %CLEAM
        [0.497, 0.511]  & 2.20\% \\
        \midrule
        \multicolumn{13}{c}{$\balpha$=[0.953,0.0.990], \texttt{Cat/Dog} (AFHQ)}\\
        \midrule
        %Point Estimate--------------------------------------------------
        $p^*_0$=0.5 & 
        0.486  & 2.80\% & 
        %Diversity
        0.469 & 6.20\% & 
        %CLEAM
        0.505  & \textbf{1.00}\%  &
        %Interval Estimate--------------------------------------------------
        [ 0.480 , 0.493 ] & 4.00\% &
        %Diversity
        [ 0.458, 0.480  ]  & 8.40\% &
        %CLEAM
        [ 0.498 , 0.511 ]  & \textbf{2.20}\% \\
        \bottomrule
    \end{tabular}
    \addtolength{\tabcolsep}{2pt}
    }
    \label{tab:fakeG_PE_Exrtacted}
\end{table*}

\begin{table*}[h]
    \centering
    %New overall Caption
    \caption{{\bf Duplicate of Tab. \ref{Tab:YoungAndAttractivePEIE} for ease of viewing.} Comparing \textbf{point estimate} and \textbf{interval estimate} of 
    Baseline \cite{choiFairGenerativeModeling2020a},
    and proposed CLEAM measurement framework on a pseudo-generator with sensitive attribute $\{\texttt{Young},\texttt{Attractive}\}$
    }
\resizebox{0.98\linewidth}{!}{
    \addtolength{\tabcolsep}{-2pt}
      \begin{tabular}
      %{c  c @{\hspace{0.5em}} c @{\hspace{1em}} c @{\hspace{0.5em}} c @{\hspace{1em}} c @{\hspace{0.5em}} c}
      {c  cc cc cc @{\hspace{1cm}} cc cc cc}
        \toprule
        %\multicolumn{7}{c}{\bf (A) Pseudo-Generator}\\
        %\midrule
        & \multicolumn{6}{c}{\bf Point Estimate} & \multicolumn{6}{c}{\bf Interval Estimate}\\
        \cmidrule(lr){2-7}\cmidrule(lr){8-13}
          \textbf{GT} & \multicolumn{2}{c}{\textbf{Baseline}
          %\cite{choiFairGenerativeModeling2020a}
          }& \multicolumn{2}{c}{\textbf{Diversity}
          %\cite{keswaniAuditingDiversityUsing2021}
          } & \multicolumn{2}{c}{\textbf{CLEAM (Ours)}}
          & \multicolumn{2}{c}{\textbf{Baseline}
          %\cite{choiFairGenerativeModeling2020a}
          }& \multicolumn{2}{c}{\textbf{Diversity}
          %\cite{keswaniAuditingDiversityUsing2021}
          } & \multicolumn{2}{c}{\textbf{CLEAM (Ours)}}
        \\
         \cmidrule(lr){1-1}\cmidrule(lr){2-3}\cmidrule(lr){4-5} \cmidrule(lr){6-7} 
         \cmidrule(lr){1-1}\cmidrule(lr){8-9}\cmidrule(lr){10-11} \cmidrule(lr){12-13}
         %Baseline
        &$\mu_{\texttt{Base}}$ & $e_\mu (\downarrow)$ 
         %Diversity
         &$\mu_{\texttt{Div}}$ & $e_\mu(\downarrow)$ 
         %CLEAM
        &$\mu_{\texttt{CLEAM}}$ & $e_\mu(\downarrow)$
         %Baseline
        &$\rho_{\texttt{Base}}$ & $e_\rho (\downarrow)$ 
         %Diversity
         &$\rho_{\texttt{Div}}$ & $e_\rho(\downarrow)$ 
         %CLEAM
        &$\rho_{\texttt{CLEAM}}$ & $e_\rho(\downarrow)$
        \\
                \midrule
        \multicolumn{13}{c}{$\balpha$=[0.749,0.852], \texttt{Young}}\\
        \midrule
         $p^*_0=0.9$ & 
         %Point estimate---------------------------------------------
        %Baseline
        0.690 & 23.33\%  & 
        %Diversity
        --- & --- & 
        %CLEAM
        0.905& \textbf{0.56}\%  &
        %Interval estimate---------------------------------------------
        %Baseline
        [0.684,0.695] & 24.00\% & 
        %Diversity
         --- & ---  & 
        %CLEAM
        [0.890,0.920] & \textbf{2.22}\%  
        \\  %\hdashline
        $p^*_0=0.8$ & 
        %Point estimate---------------------------------------------
        %Baseline
        0.630 & 21.25\%  &  
        %Diversity
        --- & --- & 
        %CLEAM
        0.804 & \textbf{0.50}\%  &
        %Interval estimate---------------------------------------------
        %Baseline
        [0.625,0.635] & 21.88\%   & 
        %Diversity
          --- &  --- & 
        %CLEAM
        [0.795,0.813] & \textbf{1.63}\%  
        \\ %\hdashline
        $p^*_0=0.7$ & 
        %Point estimate---------------------------------------------
        %Baseline
        0.570 & 18.57\%  & 
        %Diversity
        --- & ---  & 
        %CLEAM
        0.698 & \textbf{0.29}\% &
        %Interval estimate---------------------------------------------
        %Baseline
        [0.565,0.575] & 19.29\% & 
        %Diversity
        --- & ---  & 
        %CLEAM
        [0.690,0.706] & \textbf{1.43}\% 
        \\ %\hdashline
        $p^*_0=0.6$ & 
        %Point estimate---------------------------------------------
        %Baseline
        0.510 & 15.00\%  & 
        %Diversity
        --- & ---  & 
        %CLEAM
        0.595 & \textbf{0.83}\% &
        %Interval estimate---------------------------------------------
        %Baseline
        [0.505,0.515] &  15.83\% & 
        %Diversity
        --- & ---  & 
        %CLEAM
        [0.590,0.600]  & \textbf{1.67}\%
        \\ %\hdashline
        $p^*_0=0.5$ & 
        %Point estimate---------------------------------------------
        %Baseline
        0.450 & 10.0\%  & 
        %Diversity
        --- & --- &
        %CLEAM
        0.506 & \textbf{1.20}\% &
        %Interval estimate---------------------------------------------
        %Baseline
        [0.445,0.455] & 11.00\%  & 
        %Diversity
        --- & ---  & 
        %CLEAM
        [0.502,0.510] & \textbf{2.00}\%
        \\ \midrule
        %Point estimate---------------------------------------------
        \multicolumn{2}{c}{Avg Error}& 17.63\%
        && ---\% 
        && \textbf{0.68}\% 
        %Interval estimate---------------------------------------------
        & & 18.40\% 
        & & ---\% 
        & & \textbf{1.79}\% \\
        \midrule
        \multicolumn{13}{c}{$\balpha$=[0.780,0.807], \texttt{Attractive}}\\
        \midrule
        %Point estimate---------------------------------------------
        $p^*_0=0.9$ & 
        %Baseline
        0.730 & 18.89\%  & 
        %Diversity
        --- & ---  & 
        %CLEAM
        0.908 & \textbf{0.89}\% &
        %Interval estimate---------------------------------------------
        %Baseline
        [0.724,0.736] & 19.56\%  & 
        %Diversity
        --- & ---  & 
        %CLEAM
        [0.900,0.916] &  \textbf{1.78}\%
        \\ %\hdashline
        $p^*_0=0.8$ & 
        %Point estimate---------------------------------------------
        %Baseline
        0.670 & 16.25\%  & 
        %Diversity
        --- & ---  & 
        %CLEAM
        0.804 & \textbf{0.50}\% &
        %Interval estimate------------------------------------------
        %Baseline
        [0.665,0.675] & 16.88\% & 
        %Diversity
        --- & ---  &  
        %CLEAM
        [0.795,0.813] & \textbf{1.63}\%
        \\ %\hdashline
        $p^*_0=0.7$ & 
        %Point estimate---------------------------------------------
        %Baseline
        0.600 & 14.29\%  & 
        %Diversity
        --- &  --- & 
        %CLEAM
        0.696 & \textbf{0.57}\% &
        %Interval estimate---------------------------------------------
        %Baseline
        [0.594,0.606] & 15.14\% & 
        %Diversity
        --- & ---  & 
        %CLEAM
        [0.690,0.712] & \textbf{1.71}\%
        \\ %\hdashline
        $p^*_0=0.6$ & 
        %Point estimate---------------------------------------------
        %Baseline
        0.540 & 10.00\% & 
        %Diversity
        --- & ---  & 
        %CLEAM
        0.592  & \textbf{1.33}\% &
        %Interval estimate---------------------------------------------
        %Baseline
        [0.534,0.546] & 11.00\% & 
        %Diversity
        --- & ---  & 
        %CLEAM
        [0.580,0.604] & \textbf{3.33}\%
        \\ %\hdashline
        $p^*_0=0.5$ & 
        %Point estimate---------------------------------------------
        0.480 & 4.00\%  & 
        %Diversity
        --- & ---  & 
        %CLEAM
        0.493 & \textbf{1.40}\%  &
        %Interval estimate---------------------------------------------
        [0.475,0.485] & 5.00\% & 
        %Diversity
        --- & ---  & 
        %CLEAM
        [0.487,0.499] & \textbf{2.60}\% 
        \\
        \hline 
        %Point estimate---------------------------------------------
        \multicolumn{2}{c}{Avg Error}& 12.69\% 
        && ---\% 
        && \textbf{0.94}\%
        %Interval estimate---------------------------------------------
        && 13.52\% &
         & ---\% &
         & \textbf{2.22}\%
        \\
        \bottomrule
    \end{tabular}
    \addtolength{\tabcolsep}{2pt}
    }
    \label{tab:skewAblation}
\end{table*}
\FloatBarrier
\clearpage
\section{Validating Statistical Model for Classifier Output}
\subsection{ Validation of Sample-Based Estimate vs Model-Based Estimate}
\label{subsec:sampleBasedEstimate}
As described in the main paper, we utilize the sample-based estimate, $\Ddot{\mu}_{\phat_0}$,  ${\Ddot{\sigma}^2_{\phat_0}}$ as an approximate for the model-based estimate $\Tilde{\mu}_{\phat_0}$,  ${\Tilde{\sigma}^2_{\phat_0}}$. 
As discussed in Sec. \ref{apx:proposedmethod}, $\Ddot{\mu}_{\phat_0}$ allows us to find the maximum likelihood approximate of $\bpstar$.
%and ${\Ddot{\sigma}^2_{\phat_0}}$ allows us to ease computation.

\textit{\bf To validate this approximation, } we utilize a ResNet-18 trained on \texttt{Gender} and \texttt{BlackHair} to compute $\bphat$. Then with the samples from the pseudo-generators with different $\bpstar$ (following Sec. \ref{sec:psuedoG}) we computed $\bphat$ with a batch-size of $s=30$ and sample size $n=400$.
%s different $\phat$ values from the pseudo-generators (Sec. 5.3 of the main paper) , with different GT $\pstar$. 
Finally, we calculate the sample-based estimates as given in 
Eqn. 6, 7 of the main paper. As the GT $\bpstar$ and classifier's accuracy $\balpha$ is known, we also calculate the model-based estimates as given in Eqn. 4, 5 of the main manuscript and compare it against the sample-based estimates.

\textit{\bf Our results} in Tab. \ref{tab:theoreticaleval} shows that both the sample and theoretical means and standard deviations are \hypertarget{sampleStatsGoodApporx}{close approximate} to one another. Thus, we can utilise the sample statistics as a close approximation in our proposed method, CLEAM.
Additional results for different values of batch-sizes ($s$) and sample-sizes ($n$) are tabulated in Tab. \ref{tab:theoreticalevals20n400}, \ref{tab:theoreticalevals30n200} and \ref{tab:theoreticalevals200n400}.
Notice that a reduction in $s$ and $n$ values contributed to increased errors between the sample-based and model-based estimates. While making $s$ very large ($s=200$), results in the sample based estimate almost a perfectly approximating the model based estimates.
 
\begin{table}[!h]
    \centering
    \caption{
        \textbf{Comparing sample-based estimates ($\Ddot{\boldsymbol\mu}_{\boldsymbol\phat_0}$, $\Ddot{\boldsymbol\sigma}_{\boldsymbol\phat_0}$) against model-based estimates ($\tilde{\boldsymbol\mu}_{\boldsymbol\phat_0}$, $\tilde{\boldsymbol\sigma}_{\boldsymbol\phat_0}$).}
        %Validation of using sample-based estimates as given in Eqn. (5) and (6)  in place of model-based estimates as given in Eqn. (3) and (4) of the main paper, using samples from Pseudo Generators with different GT $\pstar$. Number of $\phat $ samples $s=30$ and number of images per batch $n=1k$. The GT $\pstar$ are used to compute the model-based estimates following Eqn. (3) and (4). As $\pstar$ is unknown in practice, we instead use sample-based estimates in our method. 
        The results show that sample-based estimates are close to 
        model-based estimates. 
        Furthermore, note the discrepancy between $p^*_0$ and $\Ddot{\mu}_{\phat_0}$,
        and that between $p^*_0$ and
        $\Tilde{\mu}_{\phat_0}$, highlighting the issue of using $\phat_0$ directly
        to estimate $p^*_0$
        and the
        need to compensate for the sensitive attribute classifier error as we discussed.
        We utilize a $s=30$ and $n=400$.
        }
    \resizebox{8.0cm}{!}{
    \begin{tabular}{c @{\hspace{2em}} c @{\hspace{0.1 em}} c @{\hspace{2em}} c @{\hspace{0.1em}} c}
        
        %GT & Mean $\hat{p}_0$ & Std $\hat{p}_0$ & Mean $\hat{p}_0$ &  Std $\hat{p}_0$ \\
      
        \toprule
        GT & \multicolumn{2}{c}{Sampled-based estimates} & \multicolumn{2}{c}{Model-based estimates}
        \\\cmidrule(lr){1-1} \cmidrule(lr){2-3} \cmidrule(lr){4-5} 
        &$\Ddot{\mu}_{\phat_0}$ & $\sqrt{\Ddot{\sigma}^2_{\phat_0}}$ & $\Tilde{\mu}_{\phat_0}$ &  $\sqrt{\Tilde{\sigma}^2_{\phat_0}}$ \\
        
        \midrule
        \multicolumn{5}{c}{Gender, $\balpha$=[0.976,0.979]}\\
        \midrule
        $p^*_0=0.9$ &0.881 & 0.0101 & 0.881  & 0.0106\\
        $p^*_0=0.8$ &0.781 & 0.0133 & 0.785   & 0.0135\\
        $p^*_0=0.7$ &0.692 & 0.0149 & 0.690 & 0.0152\\
        $p^*_0=0.6$ &0.590 & 0.0165 & 0.594   & 0.0162\\
        $p^*_0=0.5$ &0.503 & 0.0164 & 0.499   & 0.0164\\
        \midrule
        \multicolumn{5}{c}{$\balpha$=[0.881,0.887], Black-Hair}\\
        \midrule
        $p^*_0=0.9$ &0.802 & 0.0130 & 0.804  &	0.0139\\
        $p^*_0=0.8$ &0.723 & 0.0151 & 0.727  &	0.0162\\
        $p^*_0=0.7$ &0.653 & 0.0169 & 0.650 &	0.0177\\
        $p^*_0=0.6$ &0.580 & 0.0180 & 0.574   & 0.0186\\
        $p^*_0=0.5$ &0.502 & 0.0180 & 0.497  &	0.0189\\
        \bottomrule

    \end{tabular}}
    \label{tab:theoreticaleval}
\end{table}

\begin{table}[!h]
    \centering
    \caption{We repeat the same experiment as Tab.\ref{tab:theoreticaleval} with $s=20$ and $n=400$ samples.
        }
    \resizebox{8.0cm}{!}{
    \begin{tabular}{c @{\hspace{2em}} c @{\hspace{0.1 em}} c @{\hspace{2em}} c @{\hspace{0.1em}} c}
        
        %GT & Mean $\hat{p}_0$ & Std $\hat{p}_0$ & Mean $\hat{p}_0$ &  Std $\hat{p}_0$ \\
      
        \toprule
        GT & \multicolumn{2}{c}{Sampled-based estimates} & \multicolumn{2}{c}{Model-based estimates}
        \\\cmidrule(lr){1-1} \cmidrule(lr){2-3} \cmidrule(lr){4-5} 
        &$\Ddot{\mu}_{\phat_0}$ & $\sqrt{\Ddot{\sigma}^2_{\phat_0}}$ & $\Tilde{\mu}_{\phat_0}$ &  $\sqrt{\Tilde{\sigma}^2_{\phat_0}}$ \\
        
        \midrule
        \multicolumn{5}{c}{Gender, $\balpha$=[0.976,0.979]}\\
        \midrule
        $p^*_0=0.9$ &0.855 & 0.0201 & 0.881  & 0.0106\\
        $p^*_0=0.8$ &0.774 & 0.0211 & 0.785  & 0.0135\\
        $p^*_0=0.7$ &0.672 & 0.0219 & 0.690  & 0.0152\\
        $p^*_0=0.6$ &0.580 & 0.0181 & 0.594  & 0.0162\\
        $p^*_0=0.5$ &0.510 & 0.0230 & 0.499  & 0.0164\\
        \midrule
        \multicolumn{5}{c}{$\balpha$=[0.881,0.887], Black-Hair}\\
        \midrule
        $p^*_0=0.9$ &0.768 & 0.180 & 0.804  &	0.0139\\
        $p^*_0=0.8$ &0.712 & 0.210 & 0.727  &	0.0162\\
        $p^*_0=0.7$ &0.658 & 0.190 & 0.650 &	0.0177\\
        $p^*_0=0.6$ &0.554 & 0.230 & 0.574   & 0.0186\\
        $p^*_0=0.5$ &0.508 & 0.242 & 0.497  &	0.0189\\
        \bottomrule
    \end{tabular}}
    \label{tab:theoreticalevals20n400}
\end{table}
\begin{table}[!t]
    \centering
    \caption{We repeat the same experiment as per Tab.\ref{tab:theoreticaleval} with $s=30$ and $n=200$ samples.
        }
    \resizebox{8.0cm}{!}{
    \begin{tabular}{c @{\hspace{2em}} c @{\hspace{0.1 em}} c @{\hspace{2em}} c @{\hspace{0.1em}} c}
        
        %GT & Mean $\hat{p}_0$ & Std $\hat{p}_0$ & Mean $\hat{p}_0$ &  Std $\hat{p}_0$ \\
      
        \toprule
        GT & \multicolumn{2}{c}{Sampled-based estimates} & \multicolumn{2}{c}{Model-based estimates}
        \\\cmidrule(lr){1-1} \cmidrule(lr){2-3} \cmidrule(lr){4-5} 
        &$\Ddot{\mu}_{\phat_0}$ & $\sqrt{\Ddot{\sigma}^2_{\phat_0}}$ & $\Tilde{\mu}_{\phat_0}$ &  $\sqrt{\Tilde{\sigma}^2_{\phat_0}}$ \\
        
        \midrule
        \multicolumn{5}{c}{Gender, $\balpha$=[0.976,0.979]}\\
        \midrule
        $p^*_0=0.9$ &0.860 & 0.0232 & 0.881  & 0.0149\\
        $p^*_0=0.8$ &0.780 & 0.0286 & 0.785  & 0.0191\\
        $p^*_0=0.7$ &0.710 & 0.0294 & 0.690  & 0.0215\\
        $p^*_0=0.6$ &0.578 & 0.0380 & 0.594  & 0.0228\\
        $p^*_0=0.5$ &0.520 & 0.0321 & 0.499  & 0.0233\\
        \midrule
        \multicolumn{5}{c}{$\balpha$=[0.881,0.887], Black-Hair}\\
        \midrule
        $p^*_0=0.9$ &0.742 & 0.0312 & 0.804  &	0.0197\\
        $p^*_0=0.8$ &0.740 & 0.0332 & 0.727  &	0.0229\\
        $p^*_0=0.7$ &0.610 & 0.0291 & 0.650  &	0.0250\\
        $p^*_0=0.6$ &0.582 & 0.350 & 0.574  & 0.0262\\
        $p^*_0=0.5$ &0.542 & 0.388 & 0.497  &	0.0267\\
        \bottomrule
        
    \end{tabular}}
    \label{tab:theoreticalevals30n200}
\end{table}

\begin{table}[!t]
    \centering
    \caption{We repeat the same experiment as per Tab.\ref{tab:theoreticaleval} with $s=200$ and $n=400$ samples.
        }
    \resizebox{8.0cm}{!}{
    \begin{tabular}{c @{\hspace{2em}} c @{\hspace{0.1 em}} c @{\hspace{2em}} c @{\hspace{0.1em}} c}
        
        %GT & Mean $\hat{p}_0$ & Std $\hat{p}_0$ & Mean $\hat{p}_0$ &  Std $\hat{p}_0$ \\
      
        \toprule
        GT & \multicolumn{2}{c}{Sampled-based estimates} & \multicolumn{2}{c}{Model-based estimates}
        \\\cmidrule(lr){1-1} \cmidrule(lr){2-3} \cmidrule(lr){4-5} 
        &$\Ddot{\mu}_{\phat_0}$ & $\sqrt{\Ddot{\sigma}^2_{\phat_0}}$ & $\Tilde{\mu}_{\phat_0}$ &  $\sqrt{\Tilde{\sigma}^2_{\phat_0}}$ \\
        
        \midrule
        \multicolumn{5}{c}{Gender, $\balpha$=[0.976,0.979]}\\
        \midrule
        $p^*_0=0.9$ &0.881 & 0.0104 & 0.881  & 0.0106\\
        $p^*_0=0.8$ &0.784 & 0.0133 & 0.785   & 0.0135\\
        $p^*_0=0.7$ &0.690 & 0.0153 & 0.690 & 0.0152\\
        $p^*_0=0.6$ &0.594 & 0.0160 & 0.594   & 0.0162\\
        $p^*_0=0.5$ &0.500 & 0.0164 & 0.499   & 0.0164\\
        \midrule
        \multicolumn{5}{c}{$\balpha$=[0.881,0.887], Black-Hair}\\
        \midrule
        $p^*_0=0.9$ &0.804 & 0.0137 & 0.804  &	0.0139\\
        $p^*_0=0.8$ &0.726 & 0.0160 & 0.727  &	0.0162\\
        $p^*_0=0.7$ &0.650 & 0.0179 & 0.650 &	0.0177\\
        $p^*_0=0.6$ &0.573 & 0.0185 & 0.574   & 0.0186\\
        $p^*_0=0.5$ &0.498 & 0.0191 & 0.497  &	0.0189\\
        \bottomrule
        
    \end{tabular}}
    \label{tab:theoreticalevals200n400}
\end{table}

\FloatBarrier
\clearpage
\subsection{Goodness-of-Fit Test: \texorpdfstring{$\bphat$}{} from the Real GANs with Our Theoretical Model}
\label{subsec:goodnessOfFit}

%As discussed in the main manuscript (Sec. 5.3), in 
In order to make sure that our proposed theoretical model in Eqn. 4 and Eqn. 5 of the main paper, is also a good representation of the $\bphat$ distribution when using a %real GAN as a 
generator, we perform a
\hypertarget{goodnessoffit}{goodness of fit} test between the proposed model for the distribution of $\bphat$ and sample data generated by a GAN. 
%Here, we provide more details, in addition to those discussed in the main paper.

\iffalse
\begin{table}[!h]
    \centering
   \caption{\textbf{Validating theoretical model on GAN:} KS-test on $s=30$ and $\delta=0.05$ with $D_{crit}=0.24$. As seen from the table, since $\eta<D_{crit}$, all of the generated samples by GANs are statistically similar to the respective Gaussian at a 95\% confidence of the K-S test. }
   \resizebox{6cm}{!}{
    \begin{tabular}{c @{\hspace{1em}} c @{\hspace{1em}} c}
    \toprule
        Model Type & Sensitive Attribute & $\eta$  \\
    \midrule
        StyleGAN2 &  Gender & 0.1048 \\
        StyleSwin &  Gender & 0.1509 \\
        StyleGAN2 &  Blackhair & 0.1065\\
        StyleSwin &  Blackhair & 0.1079\\
        \bottomrule
        
    \end{tabular}
    }
    \label{tab:KStest}
\end{table}
\fi

\begin{table}[h]
    \centering
   \caption{
   %\textcolor{blue}{
   \textit{Validating goodness-of-fit of the proposed theoretical model against generated samples.} A KS-test \cite{prattConceptsNonparametricTheory1981} is conducted between the samples distribution of $\bphat$ - measured from GenData with a ResNet-18, and the theoretical distribution of $\bphat$. We utilize $s$=30, $n$=400 with $D_{crit}$=0.24. %As seen from the table, 
   Since $\eta<D_{crit}$, all of the $\bphat$ are statistically similar to the theoretical Gaussian at 95\% confidence.
   This is further observed by the sample-based mean ($\Ddot{\mu}$) $\approx$ model-based mean ($\tilde{\mu}$).}
   %}
   \resizebox{0.5\linewidth}{!}{
    \begin{tabular}{c  c  c @{\hspace{2em}} c c}
    \toprule
        Model Type & Sensitive Attribute & $\eta$ &   $\tilde{\mu}$ & $\Ddot{\mu}$  \\
    \midrule 
        \multirow{2}{*}{StyleGAN2} & \texttt{Gender}    & 0.1048 & 0.610 & 0.609\\
                                   & \texttt{Blackhair} & 0.1065 & 0.601 & 0.601\\
    \midrule
        \multirow{2}{*}{StyleSwin} & \texttt{Gender}    & 0.1509 & 0.628 & 0.629\\
                                   & \texttt{Blackhair} & 0.1079 & 0.619 & 0.614\\    
        \iffalse
        StyleGAN2 &  Gender & 0.1048 \\ 
        StyleSwin &  Gender & 0.1509 \\ 
        StyleGAN2 &  Blackhair & 0.1065\\ 
        StyleSwin &  Blackhair & 0.1079 \\
        \fi
    \bottomrule
    \end{tabular}
    }
    \label{tab:KStest}
    \vspace{-0.35cm}
\end{table}
%Similar to \cref{sub:validatetheoreticalfakeG}, we would like to first evaluate if our Gaussian model is also a good representation for the $\phat$ samples from our GAN. 
%-------------Setup/objective that we agreed upon--------------------------

%the empirical model estimated from the GAN samples.
To do this, we first obtain $s=30$ values of $\bphat$ from framework shown in Fig. 1 of the main paper, and use StyleGAN2 \cite{karrasStyleBasedGeneratorArchitecture2019} and StyleSwin \cite{zhang2021styleswin} as the generative model.
Then using ResNet-18 with known $\balpha$ and GAN's GT $\bpstar$, as discussed in Sec. 4.1 of the main paper, we form the theoretical model's Gaussian distribution, $\mathcal{N}(\tilde{\mu}_{\phat_0},\tilde{\sigma}^2_{\phat_0})$.
%-------------Setup/objective that we agreed upon--------------------------
%Recall that we showed in Sec. \ref{sec:sampleBasedEstimate} that the sample-based estimate is a good approximation of the theoretical model. Hence, we use the sample-based estimates in Eqn. 5 and Eqn. 6 of the main paper to approximated mean and variances of the theoretical model's Gaussian distribution, $\mathcal{N}(\Ddot{\mu}_{\phat_0},\Ddot{\sigma}^2_{\phat_0})$.

Now with both our model distribution and the GAN samples,
%sample-based empirical distribution, 
we utilise the Kolmogorov-Smirnov goodness of fit test (K-S test) to determine if the samples distribution is statistically similar to the proposed Gaussian model.
We thus propose the following hypothesis test for the samples $\phat^i_j, i \in \{1, \cdots, s\}$: 

\begin{flalign*}
&\mathbf{H_0}: \textit{the samples $\phat^i_j$
 belong to the modelled distribution.}&&\\
&\mathbf{H_1}: \textit{at least one of the samples $\phat^i_j$ does not match the modelled distribution.}&&
\end{flalign*}

The K-S test then measures a D-statistic ($\eta$) and compares it against a $D_{crit}$ for a given $s$. As we use $s=30$, and a significance level $\delta=0.05$ in our setup, we have $D_{crit}=0.24$. 
As seen from Tab. \ref{tab:KStest}, all of the measured $\eta$ values are below $D_{crit}$, 
thus we cannot reject the null hypothesis at a $95\%$ confidence with the K-S test. 
Therefore, we conclude that the distribution of the obtained samples from the framework (by GANs as generator) are statistically similar to the proposed Gaussian distribution. As a result, we can utilise CLEAM to approximate the $\bpstar$ range in the presence of a real GAN as the generator.

\begin{figure*}[!h]
    \centering
    \begin{subfigure}[b]{0.5\textwidth}
        \centering
        \includegraphics[height=1.6in]{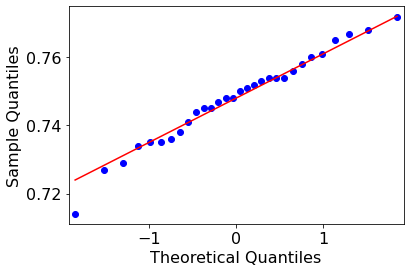}
        \caption{StyleGAN2, $\texttt{Gender}$.}
    \end{subfigure}%
    \begin{subfigure}[b]{0.5\textwidth}
        \centering
        \includegraphics[height=1.6in]{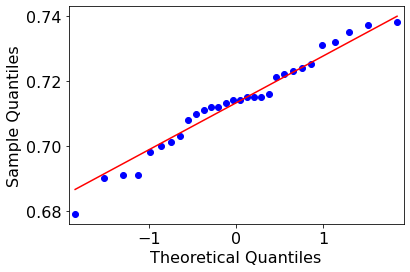}
        \caption{StyleSwin2, $\texttt{Gender}$}
    \end{subfigure}%
    \newline
        \begin{subfigure}[b]{0.5\textwidth}
        \centering
        \includegraphics[height=1.6in]{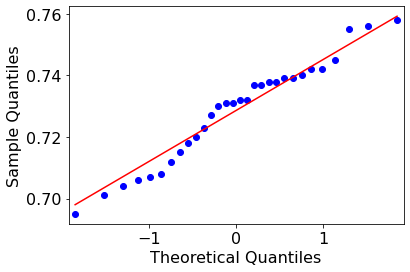}
        \caption{StyleGAN2, $\texttt{Blackhair}$}
    \end{subfigure}%
    \begin{subfigure}[b]{0.5\textwidth}
        \centering
        \includegraphics[height=1.6in]{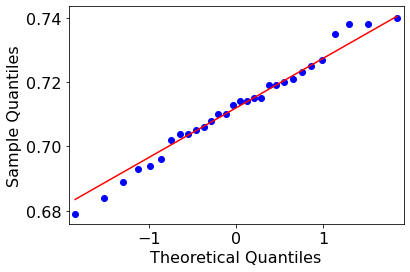}
        \caption{StyleSwin, $\texttt{BlackHair}$}
    \end{subfigure}%
     \caption{Quartile-Quartile(QQ) plot between $s=30$ $\bphat$ samples calculated for StyleGAN2 \cite{karrasStyleBasedGeneratorArchitecture2019} and StyleSwin \cite{zhang2021styleswin} generators and proposed theoretical model for $\bphat$}
     %(a Gaussian distribution in Eqn. 3, 4 of the main paper)}
    \label{fig:QQplot}
\end{figure*}

%In Section 6.1 of the main paper, we discuss that samples of $\phat$ obtained from real GAN generators in \cite{choiFairGenerativeModeling2020a} follow a Gaussian distribution.
%Here, we 
We further perform a Quantile-Quantile(QQ) analysis to provide a more visual representation. In particular, we plot the Quantile-Quantile(QQ) plot between the $\bphat$ samples (produced for the data generated by the GAN) and proposed model.
%man_24Nov: what are the parameters of this Gaussian model? how did you calculate?
As seen in Fig. \ref{fig:QQplot}, the  $\bphat$ samples from GAN correlate tightly with the standardised line (in red), a line indicating a perfect correlation between theoretical and sample quantiles. This analysis supports our claim that the $\bphat$ samples from a real generator (GAN) follow the distribution estimated by the proposed model.

%the  $\phat$ samples from the 
%real GAN generators in \cite{choiFairGenerativeModeling2020a} \textcolor{red}{(Can we say this a bit general? e.g.: 'the $\phat$ samples from a real generator (GAN) follow the distribution estimated by the proposed model.')} follow a Gaussian distribution.
%\newpage
\clearpage
\FloatBarrier
\section{Additional Experiments}
\label{sec:additionalExp}
\subsection{Experimental Results with Standard Deviation}
%RealGAN and Psuedo-GAN with standard deviation}
\label{subsec:resultsWithError}
\FloatBarrier
%\textcolor{red}{TO INCLUDE ERROR BARS**********************************************}

%\textcolor{blue}{
In the main manuscript, we did not include the error bars of our experiments due to space constraints. Hence,
in this section, we provide the full tables for Tab.
%\ref{tab:G_PE} 
1 and 2 of the main manuscript
%, and additionally Tab. \ref{tab:fakeG_PE}, 
with the standard deviation over 5 runs.
%}
%We were unable to include this in the main manuscript, due to space constraints. 
Note that generally, the standard deviation at each test point is relatively small and hence can be considered as negligible. This is likely due to the large $s$ and $n$ utilized. As a result, we can utilize the mean results (as seen in the main manuscript) to compare CLEAM against Diversity and the Baseline. 

\begin{table}[!h]
    \centering
    %New overall Caption
    \caption{
    %{\color{blue} Man:   need to push these tables to later section in experiment even this may be referred earlier; }
    Comparing the {\em point estimates} and {\em interval estimates} of Baseline \cite{choiFairGenerativeModeling2020a}, Diversity \cite{keswaniAuditingDiversityUsing2021} and our proposed CLEAM measurement framework in estimating $\bpstar$ of the GenData datasets sampled from (A) StyleGAN2 \cite{karrasStyleBasedGeneratorArchitecture2019} and (B) StyleSwin \cite{zhang2021styleswin}. 
    The $\pstar_0$ value for each GAN with a certain sensitive attribute is determined by manually hand-labeling the generated data.
    We utilize four different sensitive attribute classifier Resnet-18/34 (R18, R34)\cite{heDeepResidualLearning2016}, MobileNetv2 (MN2)\cite{sandlerMobileNetV2InvertedResiduals2018} and VGG-16 (V16)\cite{simonyanVeryDeepConvolutional2014}, with different accuracy $\balpha$, to classify attributes \texttt{Gender} and \texttt{BlackHair}, to obtain $\bphat$.
    Each $\bphat$ utilizes $n=400$ samples and is evaluated for a batch size of $s=30$.
    We repeat this for 5 experimental runs and report the mean error rate, per Eqn. 1 of the main manuscript.
    %\textcolor{red}{\bf [Milad: How do we measure the $e_{\mu}$ considering the standard deviation value?]}
    %\ref{eqn:pointError} and \ref{eqn:intervalError}. 
    %\textcolor{red}{More analysis with different $n$ and $s$ are included in Supp \ref{subsec:trainingParam}.}
   }
    % We then compared the point estimate against the GT $p^*$ of the datasets and reported the mean error rate, Eqn. \ref{eqn:pointError}, for 5 experimental runs.}
\resizebox{\textwidth}{!}{
    \addtolength{\tabcolsep}{-4pt}
      \begin{tabular}
      {ccc cc cc @{\hspace{0.3cm}} cc cc cc}
        \toprule
        %& & 
        & \multicolumn{6}{c}{\bf Point Estimate} & \multicolumn{6}{c}{\bf Interval Estimate}
        \\
        %\cmidrule(lr){4-9}\cmidrule(lr){10-15}
        \cmidrule(lr){2-7}\cmidrule(lr){8-13}
          %\textbf{Classifier} 
          &
          %\textbf{  $\balpha=\{\alpha_0,\alpha_1\}$ } &
          %Avg. $\balpha$ &
          %\textbf{  $\balpha=\{\alpha_0,\alpha_1\}$ } &
          \multicolumn{2}{c}{\textbf{Baseline}}
          %\cite{choiFairGenerativeModeling2020a}}
          & 
          \multicolumn{2}{c}{\textbf{Diversity}}
          %\cite{keswaniAuditingDiversityUsing2021}} 
          & \multicolumn{2}{c}{\textbf{CLEAM (Ours)}}
          %Interval Estimates
           & \multicolumn{2}{c}{\textbf{Baseline}}
           %\cite{choiFairGenerativeModeling2020a}}
           & 
           \multicolumn{2}{c}{\textbf{Diversity}}
           %\cite{keswaniAuditingDiversityUsing2021}} 
           & \multicolumn{2}{c}{\textbf{CLEAM (Ours)}}
        \\
        %\cmidrule(lr){1-1}\cmidrule(lr){2-2}\cmidrule(lr){3-3}\cmidrule(lr){4-5}\cmidrule(lr){6-7} \cmidrule(lr){8-9} \cmidrule(lr){10-11}\cmidrule(lr){12-13} \cmidrule(lr){14-15}
        \cmidrule(lr){1-1}\cmidrule(lr){2-3}\cmidrule(lr){4-5} \cmidrule(lr){6-7} \cmidrule(lr){8-9}\cmidrule(lr){10-11} \cmidrule(lr){12-13}
        %& 
        %& 
        &
        $\mu_{\texttt{Base}}$ & $e_{\mu}(\downarrow)$ 
         %Diversity
         &$\mu_{\texttt{Div}}$ & $e_\mu(\downarrow)$
         %CLEAM
        &$\mu_{\texttt{CLEAM}}$ & $e_\mu(\downarrow)$
        &$\rho_{\texttt{Base}}$ & $e_\rho(\downarrow)$ 
        &$\rho_{\texttt{Div}}$ & $e_\rho(\downarrow)$ 
        &$\rho_{\texttt{CLEAM}}$ & $e_\rho(\downarrow)$\\
        \midrule
        \multicolumn{13}{c}{\bf (A) StyleGAN2}\\
        \midrule
        \multicolumn{13}{c}{\texttt{Gender} with GT class probability {\bfseries\boldmath $p^*_0$=0.642} }\\
        \midrule
        R18 
        %& \{0.947, 0.983\} & 0.97 
        &
        %Point Estimate----------------------------------------------------------
        0.610 $\pm$ 0.004 & 4.98\% & 
        %Diversity
        --- & --- & 
        %CLEAM
        0.638 $\pm$ 0.006 & \textbf{0.62\%} & 
        %Interval Estimate----------------------------------------------------------
        [0.602$\pm$ 0.004, 0.618$\pm$ 0.004] & 6.23\% & 
        %Diversity
        --- & --- & 
        %CLEAM
       [0.629 $\pm$ 0.006, 0.646$\pm$ 0.006] & \textbf{2.02\%}
        \\  
        %\hdashline
        %Point Estimate----------------------------------------------------------
        R34 & 
        %\{0.932, 0.976]\} & 0.95 & 
        0.596$\pm$ 0.003 & 7.17\% &
        %Diversity
        --- & --- & 
        %CLEAM
        0.634$\pm$ 0.002 & 
        \textbf{1.25\%} &
        %Interval Estimate----------------------------------------------------------
        [0.589$\pm$ 0.003, 0.599$\pm$ 0.003] & 8.26\% &
        %Diversity
        --- & --- &
        %CLEAM
        [0.628$\pm$ 0.002, 0.638$\pm$ 0.002] & \textbf{2.18\%} 
        \\ 
        %\hdashline
        %Point Estimate----------------------------------------------------------
        MN2 & 
        %\{0.938, 0.975\} & 0.96 & 
        0.607 $\pm$ 0.003 & 5.45\% & 
        %Diversity
        --- & --- &  
        %CLEAM
        0.637 $\pm$ 0.002  & \textbf{0.78}\% &
        %Interval Estimate----------------------------------------------------------
        [0.602 $\pm$ 0.003, 0.612 $\pm$ 0.003] & 6.23\% &
        %Diversity
        --- & --- & 
        %CLEAM
        [0.632 $\pm$ 0.002, 0.643 $\pm$ 0.002] & $\textbf{1.56\%}$
        \\ 
        %\hdashline
        %Point Estimate----------------------------------------------------------
        V16 & 
        %\{0.801, 0.919\} & 0.86 & 
        0.532 $\pm$ 0.007 & 17.13\% & 
        %Diversity
        0.550 $\pm$ 0.011 & 14.3\% & 
        %CLEAM
        0.636 $\pm$ 0.007  & \textbf{0.93}\% &
        %Interval Estimate----------------------------------------------------------
        [0.526 $\pm$ 0.007, 0.538 $\pm$ 0.007] & 18.06\% &
        %Diversity
        [0.536 $\pm$ 0.011 , 0.564 $\pm$ 0.011] & 16.51\% & 
        %CLEAM
        [0.628 $\pm$ 0.007, 0.644 $\pm$ 0.007] & \textbf{2.18\%}
        \\
        \midrule
        %Point Estimate----------------------------------------------------------
        %& & 
        \multicolumn{2}{c}{Avg Error}& 8.68\% 
        & & 14.30\%
        & &\textbf{0.90}\%
        %Interval Estimate----------------------------------------------------------
        & & 9.70\%
        & & 16.51\%
        & & \textbf{1.99}\%
        \\
        \midrule
    %BlackHair
        \multicolumn{13}{c}{\texttt{BlackHair} with GT class probability {\bfseries\boldmath $p^*_0$=0.643}}\\
        \midrule
        %Point Estimate----------------------------------------------------------
        R18 & 
        %\{0.869, 0.885\} & 0.88 & 
        0.599 $\pm$ 0.006 & 6.84\% & 
        %Diversity
        --- & --- &  
        %CLEAM
        0.641 $\pm$ 0.004 & \textbf{0.31\%} &
        %Interval Estimate----------------------------------------------------------
        [0.591 $\pm$ 0.006, 0.607 $\pm$ 0.005] & 8.08\% & 
        %Diversity
        --- & --- & 
        %CLEAM
        [0.631 $\pm$ 0.004, 0.652 $\pm$ 0.003] & \textbf{1.40\%}
        \\  
        %\hdashline
        %Point Estimate----------------------------------------------------------
        R34 & 
        %\{0.834, 0.916\}  & 0.88 & 
        0.566 $\pm$ 0.007 & 11.98\%&
        %Diversity
        --- & --- & 
        %CLEAM
        0.644 $\pm$ 0.008 & 
        \textbf{0.16}\% &
        %Interval Estimate----------------------------------------------------------
        [0.561 $\pm$ 0.007, 0.572 $\pm$ 0.006] & 12.75\% &
        %Diversity
        --- & --- &
        %CLEAM
        [0.637 $\pm$ 0.009, 0.651 $\pm$ 0.008] & \textbf{1.24\%}
        \\ %\hdashline
        %Point Estimate----------------------------------------------------------
        MN2 & 
        %\{0.839, 0.881\} & 0.86 & 
        0.579 $\pm$ 0.007 & 9.95\% & 
        %Diversity
        --- & --- &  
        %CLEAM
        0.639 $\pm$ 0.007  & \textbf{0.62}\% &
        %Interval Estimate----------------------------------------------------------
        [0.574 $\pm$ 0.008, 0.584 $\pm$ 0.008] & 10.73\% &
        %Diversity
        --- & --- & 
        %CLEAM
        [0.632 $\pm$ 0.007, 0.647 $\pm$ 0.007] & \textbf{1.71\%}
        \\ %\hdashline
        %Point Estimate----------------------------------------------------------
        V16 & 
        %\{0.851, 0.836\} & 0.84 & 
        0.603 $\pm$ 0.004 & 6.22\% & 
        %Diversity
        0.582 $\pm$ 0.011 & 9.49\% & 
        %CLEAM
        0.640 $\pm$ 0.005 & \textbf{0.47}\%
        %Interval Estimate----------------------------------------------------------
        & 
        [0.597 $\pm$ 0.004, 0.608 $\pm$ 0.003] & 7.15\% &
        %Diversity
        [0.568 $\pm$ 0.010, 0.596 $\pm$ 0.011] & 11.66\% &
        %CLEAM
        [0.632 $\pm$ 0.004, 0.648 $\pm$ 0.005] & \textbf{1.71\%}
        \\
        \midrule
        %Point Estimate----------------------------------------------------------
        %& & 
        \multicolumn{2}{c}{Avg Error}& 8.75\%
        && 9.49\% 
        && \textbf{0.39}\%
        %Interval Estimate----------------------------------------------------------
        && 9.68\%
        && 11.66\%
        && \textbf{1.52}\%
        \\
        \midrule
        \multicolumn{13}{c}{\bf (B) StyleSwin}\\
        \midrule
        \multicolumn{13}{c}{\texttt{Gender} with GT class probability {\bfseries\boldmath $p^*_0$=0.656}}\\
        \midrule
        R18 & 
        %\{0.947, 0.983\} & 0.97 & 
        %Point Estimate----------------------------------------------------------
        0.620 $\pm$ 0.005 & 5.49\% & 
        %Diversity
        --- & --- &  
        %CLEAM
        0.648 $\pm$ 0.004 & \textbf{1.22\%} 
        %Interval Estimate----------------------------------------------------------
        & [0.612 $\pm$ 0.004,0.629 $\pm$ 0.005] & 6.70\% & 
        %Diversity
        --- & --- & 
        %CLEAM
        [0.639 $\pm$ 0.005,0.658 $\pm$ 0.005] & \textbf{2.59\%}
        \\  %\hdashline
        R34 & 
        %\{0.932, 0.976\}  & 0.95 & 
        %Point Estimate----------------------------------------------------------
        0.610 $\pm$ 0.002 & 7.01\% &
        %Diversity
        --- & --- & 
        %CLEAM
        0.649 $\pm$ 0.005 & \textbf{1.07}\%  
        %Interval Estimate----------------------------------------------------------
        & [0.605 $\pm$ 0.003,0.615 $\pm$ 0.003] & 7.77\%&
        %Diversity
        --- & --- &
        %CLEAM
        [0.643 $\pm$ 0.006,0.654 $\pm$ 0.006] & 
        \textbf{1.98}\% 
        \\ %\hdashline
        MN2 & 
        %\{0.938, 0.975\} & 0.96 & 
        %Point Estimate----------------------------------------------------------
        0.623 $\pm$ 0.008 & 5.03\% & 
        %Diversity
        --- & --- & 
        %CLEAM
        0.655 $\pm$ 0.005  & \textbf{0.15}\% 
        %Interval Estimate----------------------------------------------------------
        & [0.618 $\pm$ 0.007,0.629$\pm$ 0.007] & $5.79\%$ & 
        %Diversity
        --- & --- & 
        %CLEAM
        [0.649 $\pm$ 0.006,0.661 $\pm$ 0.006]  & \textbf{1.07}\%
        \\ %\hdashline
        V16 & 
        %\{0.801, 0.919\} & 0.86 & 
        %Point Estimate----------------------------------------------------------
        0.555 $\pm$ 0.004 & 15.39\% & 
        %Diversity
        0.562 $\pm$ 0.015 & 14.33\% & 
        %CLEAM
        0.668 $\pm$ 0.006  & \textbf{1.83}\% 
         %Interval Estimate----------------------------------------------------------
        &[0.549 $\pm$ 0.004,0.560 $\pm$ 0.004] & 16.31\% & 
        %Diversity
        [0.548 $\pm$ 0.014,0.576 $\pm$ 0.014] & 16.46\% & 
        %CLEAM
        [0.660 $\pm$ 0.007,0.675 $\pm$ 0.007]  & \textbf{2.90}\%
        \\
        \midrule
        %Point Estimate----------------------------------------------------------
        %& & 
        \multicolumn{2}{c}{Avg Error}& 8.23\%
        && 14.33\%
        && \textbf{1.07}\%
        %Interval Estimate----------------------------------------------------------
        & & 9.14\%
        && 16.46\% 
        && \textbf{2.14}\%
        \\
        \midrule
    %BlackHair
        \multicolumn{13}{c}{\texttt{BlackHair} with GT class probability {\bfseries\boldmath $p^*_0$=0.668}}\\
        \midrule
        R18 & 
        %\{0.869, 0.885\} & 0.88 & 
        %Point Estimate----------------------------------------------------------
        0.612 $\pm$ 0.005 & 8.38\% & 
        %Diversity
        --- & --- &  
        %CLEAM
        0.659 $\pm$ 0.006 & \textbf{1.35\%} 
        %Interval Estimate----------------------------------------------------------
        & [0.605 $\pm$ 0.005,0.620 $\pm$ 0.006] & 9.43\% &
        %Diversity
        --- & --- & 
        %CLEAM
        [0.649 $\pm$ 0.004,0.670 $\pm$ 0.004] & \textbf{2.84\%} 
        \\  %\hdashline
        R34 & 
        %\{0.834, 0.916\} & 0.88 & 
        %Point Estimate----------------------------------------------------------
        0.581 $\pm$ 0.006 & 13.02\%&
        %Diversity
        --- & --- & 
        %CLEAM
        0.662 $\pm$ 0.006 & \textbf{0.90}\%  
        %Interval Estimate----------------------------------------------------------
        & [0.576 $\pm$ 0.005,0.586 $\pm$ 0.006] & 13.77\%&
        %Diversity
        --- & --- &
        %CLEAM
        [0.656 $\pm$ 0.005,0.669 $\pm$ 0.005] & 
        \textbf{1.80}\%
        \\ %\hdashline
        MN2 & 
        %\{0.839, 0.881\} & 0.86 & 
        %Point Estimate----------------------------------------------------------
        0.596 $\pm$ 0.006 & 10.78\% & 
        %Diversity
        --- & --- & 
        %CLEAM
        0.659 $\pm$ 0.005  & \textbf{1.35}\% 
        %Interval Estimate----------------------------------------------------------
        & [0.591 $\pm$ 0.006,0.600 $\pm$ 0.007] & 11.50\% & 
        %Diversity
        --- & --- & 
        %CLEAM
        [0.652 $\pm$ 0.005,0.666$\pm$ 0.005]  & \textbf{2.40}\%
        \\ %\hdashline
        V16 & 
        %\{0.851, 0.836\} & 0.84 & 
        %Point Estimate----------------------------------------------------------
        0.625  $\pm$ 0.006 & 6.44\% & 
        %Diversity
        0.608 $\pm$ 0.014 & 8.98\% & 
        %CLEAM
        0.677 $\pm$ 0.005  & \textbf{1.35}\% 
        %Interval Estimate----------------------------------------------------------
        & [0.620 $\pm$ 0.005,0.630 $\pm$ 0.006] & 7.19\% & 
        %Diversity
        [0.590 $\pm$ 0.012,0.626 $\pm$ 0.013] & 11.68\% & 
        %CLEAM
        [0.670 $\pm$ 0.005,0.684 $\pm$ 0.006]  & \textbf{2.40}\%
        \\
        \midrule
        %Point Estimate----------------------------------------------------------
        %& & 
        \multicolumn{2}{c}{Avg Error}& 9.66\% 
        &&8.98\% 
        &&\textbf{1.24}\%
        %Interval Estimate----------------------------------------------------------
        && 10.47\% 
        && 11.68\%
        && \textbf{2.36}\%
        \\
        \midrule
    \end{tabular}
    }
    \addtolength{\tabcolsep}{4pt}
\end{table}

\begin{table}[h]
    \centering
    %New overall Caption
    \caption{ 
    %\textcolor{blue}{
    Comparing the {\em point estimates} and {\em interval estimates} of Baseline and CLEAM in estimating the $\bpstar$ of the Stable Diffusion Model \cite{rombach2021highresolution} with the GenData-SDM dataset.
   We use prompt input starting with "A photo of with the face of" and ending with synonymous (Gender neutral) prompts. We utilized CLIP as the sensitive attribute classifier for \texttt{Gender}, to obtain $\bphat$. 
    %\textcolor{red}{Chris: this table has been flipped back to F/M}
   % }
    }
\resizebox{\textwidth}{!}{
    %\addtolength{\tabcolsep}{-5pt}
      \begin{tabular}
      {c c  cc cc cc cc }
        \toprule
        & & \multicolumn{4}{c}{\bf Point Estimate} & \multicolumn{4}{c}{\bf Interval Estimate}\\
        \cmidrule(lr){3-6}\cmidrule(lr){7-10}
          \textbf{Prompt} & \textbf{GT} & \multicolumn{2}{c}{\textbf{Baseline}
          }& \multicolumn{2}{c}{\textbf{CLEAM (Ours)}}
          & \multicolumn{2}{c}{\textbf{Baseline}
          } & \multicolumn{2}{c}{\textbf{CLEAM (Ours)}}
        \\
        \cmidrule(lr){1-1} \cmidrule(lr){2-2} \cmidrule(lr){3-4} \cmidrule(lr){5-6} \cmidrule(lr){7-8} \cmidrule(lr){9-10}
         %\cmidrule(lr){1-1}\cmidrule(lr){2-2\cmidrule(lr){3-4}\cmidrule(lr){5-6} \cmidrule(lr){7-8} \\
         %Baseline
        & & $\mu_{\texttt{Base}}$ & $e_\mu (\downarrow)$ 
         %CLEAM
        &$\mu_{\texttt{CLEAM}}$ & $e_\mu(\downarrow)$
         %Baseline
        &$\rho_{\texttt{Base}}$ & $e_\rho (\downarrow)$ 
         %CLEAM
        &$\rho_{\texttt{CLEAM}}$ & $e_\rho(\downarrow)$
        \\
        %\midrule
        %\multicolumn{10}{c}{\cellcolor{lightgray!20} \bf (A) Stable Diffusion}\\
        \midrule
        \multicolumn{10}{c}{$\balpha$=[0.998,0.975], Avg. $\alpha$=0.987, CLIP --\texttt{Gender}}\\
        \midrule
        %GT $\pstar_0$ & --- & 0.112 & 0.814 & 0.262 & 0.548 & 0.226 \\
        %Point Estimate--------------------------------------------------
        %"A photo with the face of \underline{Somebody}"      & 0.112 & 0.132 & 17.86\% & 0.114 & 1.79\%  & [ 0.128 , 0.136 ] & 21.43\%  & [ 0.110 , 0.118 ] & 5.36\% \\
        "A photo with the face of \underline{an individual}" & 0.186 & 0.203 $\pm$ 0.011 & 9.14\% & 0.187 $\pm$ 0.11 & {\bf 0.05\%} 
        & [ 0.198 $\pm$ 0.10 , 0.208 $\pm$ 0.10 ] & 11.83\% & [ 0.182 $\pm$ 0.10  , 0.192 $\pm$ 0.10 ] & {\bf 3.23\%} \\
        "A photo with the face of \underline{a human being}"   & 0.262 & 0.277 $\pm$ 0.10  & 5.73\%  & 0.263 $\pm$ 0.10  & {\bf  0.38\%} 
        & [ 0.270 $\pm$ 0.10 , 0.285 $\pm$ 0.10 ] & 8.78\% & [ 0.255 $\pm$ 0.10 , 0.271 $\pm$ 0.10 ] & {\bf 3.44\%}  \\
        "A photo with the face of \underline{one person}"    & 0.226 & 0.241 $\pm$ 0.009  & 6.63\% & 0.230 $\pm$ 0.08  & {\bf 1.77\%} 
        & [ 0.232 $\pm$ 0.10 , 0.251 $\pm$ 0.10 ] & 11.06\% &  [ 0.220 $\pm$ 0.09 , 0.239 $\pm$ 0.09 ] & {\bf 5.75\%} \\
        "A photo with the face of \underline{a person}"      &  0.548 & 0.556 $\pm$ 0.12 & 1.49\% & 0.548 $\pm$ 0.11 & {\bf 0.00\%} 
        & [ 0.545 $\pm$ 0.11 , 0.566 $\pm$ 0.11 ] & 3.28\% & [ 0.537 $\pm$ 0.11 , 0.558 $\pm$ 0.11 ] & {\bf 2.01\%} \\
        \midrule
        \multicolumn{3}{c}{Average Error} & 5.75\% & & {\bf 0.44\%} & & 8.74\% & & {\bf 3.61\%}\\
        %CLEAM
        %Interval Estimate--------------------------------------------------
        %CLEAM
        \bottomrule
    \end{tabular}
    }
    \label{tab:G_PE2_SD}
\end{table}

\clearpage
\subsection{Experimental Setup for  Diversity\cite{keswaniAuditingDiversityUsing2021}}
\label{sec:DiversitySetup}
In this section, we describe our \hypertarget{diversitySetup}{experimental} setup for Diversity \cite{keswaniAuditingDiversityUsing2021}, as utilized in 
%Sec. 5 and 6 of 
the main paper.
Recall that as discussed by Kewsani \etal \cite{keswaniAuditingDiversityUsing2021} a VGG-16 \cite{simonyanVeryDeepConvolutional2014} model pre-trained on ImageNet \cite{dengImageNetLargescaleHierarchical2009} is utilized as a feature extractor. 
Then, this feature extractor is applied to both the unknown (generator's data) and the controlled dataset.
Finally, the unknown sample's features are compared against the controlled one's via a similarity algorithm to compute diversity, $\delta$. 
%Then these features are fed into a similarity algorithm to measure Diversity.

\begin{figure}[h!]
\begin{subfigure}[b]{\textwidth}
       \centering
    \includegraphics[width=\textwidth]{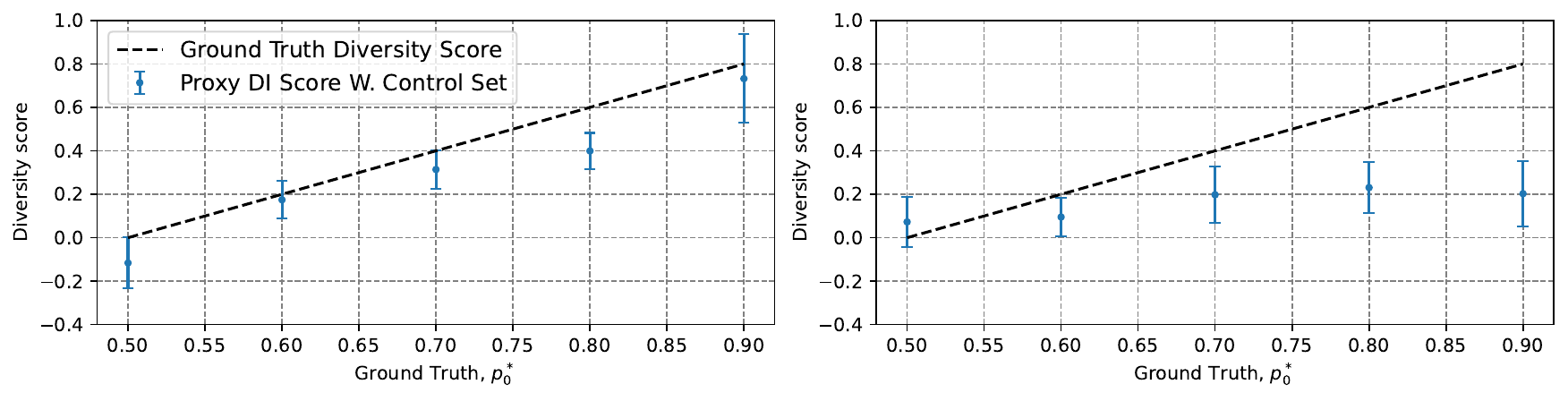}
    \caption{VGG-16 pre-trained on ImageNet}
    \label{fig:VGG16imageNet} 
\end{subfigure}
\begin{subfigure}[b]{\textwidth}
       \centering
    \includegraphics[width=\textwidth]{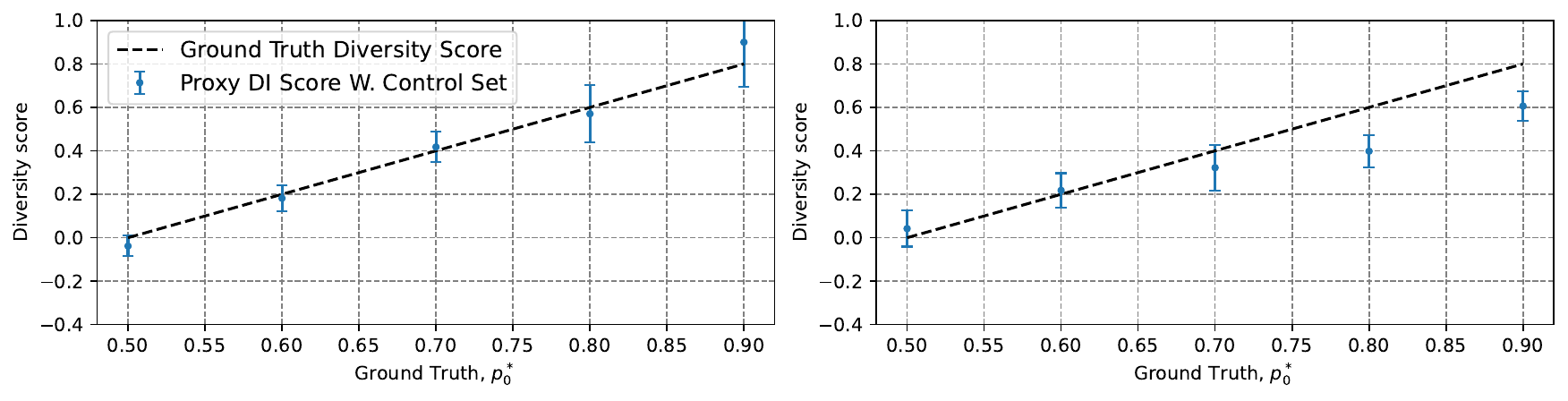}
    \caption{VGG-16 pre-trained on ImageNet then fine-tuned on CelebA}
    \label{fig:VGG16CelebA}  
\end{subfigure}
    \caption{{\bf Improvement in Diversity by fine-tuning the VGG-16, as a feature extractor}:
    %\textcolor{red}{(as feature extractor? if yes, please also include a brief description in paragraph 2 of D2)}:} 
    \textbf{(a)} Diversity implementation by \cite{keswaniAuditingDiversityUsing2021} with VGG-16 pre-trained on ImageNet as the feature extractor testing on the pseudo-generator's with $\pstar_0=\{0.9,0.8,0.7,0.6,0.5\}$ for sensitive attribute $\texttt{Gender}$(Left) and $\texttt{BlackHair}$(Right).
    \textbf{(b)} We re-implemented VGG-16 and furter fine-tune it with CelebA as the feature extractor. We observed  improvement in predicting the GT $\bpstar$}
\end{figure}

From our results in Fig. \ref{fig:VGG16imageNet} (LHS) based on the pseudo-generator's setup (discussed in more details in Sec. \ref{sec:psuedoG}), we recognize that the original implementation with VGG-16 trained on ImageNet works well on the $\texttt{Gender}$ sensitive attribute.
This is seen by the close approximation made by the proxy diversity score when compared against the GT diversity score evaluated with Eqn. \ref{eqn:GTDiversity}, as per \cite{keswaniAuditingDiversityUsing2021}.
\begin{equation}
    GT \, Diversity =\pstar_0 - \pstar_1
    \label{eqn:GTDiversity}
\end{equation}
However, when evaluated on the harder $\texttt{BlackHair}$  sensitive attribute, our results in Fig. \ref{fig:VGG16imageNet} (RHS) observed significant error between the GT Diversity scores and the proxy Diversity scores. 
This error was especially prevalent in the larger biases \eg $\pstar_0=0.9$. We theorized that this was due to the differences between the domains of the feature extractor and the generated/controlled images \ie ImageNet versus CelebA/CelebA-HQ.

To verify this, we 
%retrained
fine-tune
the VGG-16 model on the CelebA dataset with the respective sensitive attribute. 
Then we removed the last fully connected layer of the classifier model, and utilise the 4096 feature vector for the diversity measurement, as per \cite{keswaniAuditingDiversityUsing2021}. Our results in Fig. \ref{fig:VGG16CelebA} demonstrate significant improvement on both $\texttt{Gender}$ and $\texttt{BlackHair}$, based on the new improved VGG-16 model implementation. This thereby verifies our intuition that there exists a mismatch of domains in the VGG-16 pretrained on ImageNet when utilized with CelebA samples.  

\begin{figure}[h!]
\begin{subfigure}[b]{0.5\textwidth}
       \centering
    \includegraphics[width=\textwidth]{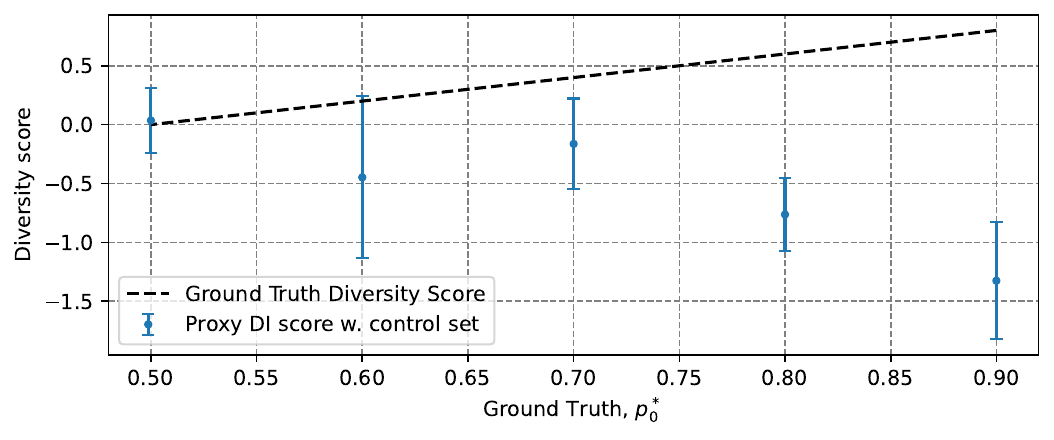}
    \caption{VGG-16(CelebA) on $\texttt{Attractive}$}
\end{subfigure}
\begin{subfigure}[b]{0.5\textwidth}
       \centering
    \includegraphics[width=\textwidth]{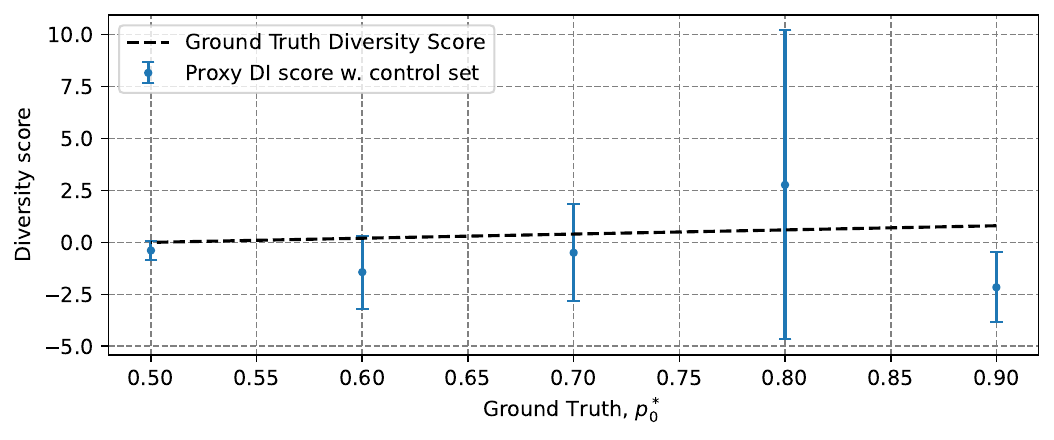}
    \caption{VGG-16(CelebA) on $\texttt{Young}$}
\end{subfigure}
    \caption{\textbf{Limitations Of Diversity
    algorithm.} %Implementation 
    %(\textcolor{red}{implementation or algorithm?})}.
    Our implementation of VGG-16 fine-tuned on CelebA \wrt sensitive attribute $\texttt{Attractive}$ and $\texttt{Young}$. VGG-16 Classifier achieved an accuracy of $78.44\%$ and $84.1\%$ for sensitive attribute $\texttt{Attractive}$ and $\texttt{Young}$. However, the same VGG-16 performs poorly on the diversity metric, demonstrating the limitations of the diversity framework. }
    \label{fig:VGG16AttractiveYoung}
\end{figure}

However, upon further experimentation, we recognize certain limitations still exist in the Diversity measure when used on more ambiguous and harder sensitive attribute \eg $\texttt{Young}$ and $\texttt{Attractive}$. Similar to before, we fine-tuned the sensitive attribute classifier (feature extractor) which achieved accuracies of $78.44\%$ and $84.41\%$ for $\texttt{Young}$ and $\texttt{Attractive}$, respectively. However even with this re-implementation, the diversity persistent to perform poorly, as seen in Fig. \ref{fig:VGG16AttractiveYoung}.   

%As per the senstive attribute \texttt{BlackHair}, we first re-trained the VGG-16 model on the respective SA and measured their accuracies on a validation dataset. 
%We noted an average accuracy of $78.44\%$ and $84.41\%$ for $\texttt{Young}$ and $\texttt{Attractive}$. Next, we removed the last layer of the classifier and utilised the 4096 feature vector for the implementation of diversity, as per \cite{keswaniAuditingDiversityUsing2021}. Unlike the original classifier (with the last fully connected layer) the same models were unable to accurately measure the respective $\pstar_0$ (biases) with the diversity algorithm, as seen in Fig. \ref{fig:VGG16AttractiveYoung}. From this, we observe that even our fined tuned VGG-16 feature extractor did not improve the Diversity framework on these harder SA.
Regardless, given the improvement seen on the \texttt{BlackHair} sensitive attribute, we utilized our improved VGG-16 feature extractor in the main paper, in place of the pre-trained VGG-16 (ImageNet).

%We conclude that the final fully connected layer is vital for the recognition of these harder LA. \textcolor{red}{(We don't need to include this as we don't have enough analysis. We may simply say that 'even fine-tuned version of Diversity is not working well for harder LAs'.)}


\clearpage
\FloatBarrier
\subsection{
%\textcolor{blue}{
Measuring Varying Degrees of Bias}
%}
\label{sec:psuedoG}

%\textcolor{red}{Chris: Reviewing \dots}

{\bf CLEAM for Measuring Varying Degrees of Bias.}
In previous experiments, we show the performance of different methods in measuring the fairness of generators and evaluating bias mitigation techniques. Another interesting analysis would be to see how these methods fare with 
different 
%degrees of 
bias, \ie different $\bpstar$ values. 
A challenge of this analysis is that we cannot control the training dynamics of 
%the GANs
%\textcolor{blue}{
either the GANs nor the Stable Diffusion Model
%}
to obtain an exact value of $\bpstar$.
Thus, we introduce a new setup and use a {\em pseudo-generator} instead of real GANs.
%instead of the real one in this experiment.

In this setup, we utilize the CelebA \cite{liuDeepLearningFace2015} and the AFHQ \cite{choiStarGANV2Diverse2020} dataset to construct different modified datasets that follow different values of $\bpstar$ \wrt the sensitive attribute \eg
%For example, for 
\texttt{BlackHair} attribute, when $\bpstar=\{0.9,0.1\}$, the modified dataset contains 4880 \texttt{BlackHair} and 542 \texttt{Non-BlackHair} samples.
Then, a pseudo-generator with bias $\bpstar$ works by random sampling from the corresponding datasets.
Note that the samples in the modified dataset are unseen to the sensitive attribute classifier.
For our experiment, we use different GT values, $\bpstar=\{p_0^*,p_1^*\}$, where $p_0^* \in \{0.9,0.8,0.7,0.6,0.5\}$, and $p_1^*=1-p_0^*$. 
For a pseudo-generator, to calculate each value of $\bphat$, a batch of $n$ samples is randomly drawn from the corresponding dataset and fed into the $C_\bu$ for classification. 
We utilize a ResNet-18 to evaluate our pseudo-generator.
%the same ResNet-18 as per Sec. \ref{subsec:biasmitigation}, to evaluate our pseudo-generator. 
%See Supp. \ref{subsec:pseudoGMoreLA} and \ref{subsec:pseudoGMoreClassifiers} for more analysis. 
{\bf The results in Tab. \ref{tab:fakeG_PE}} for $p^*_0$ demonstrate that CLEAM is effective for different degrees of bias, reducing the average error ($e_{\mu}$) of the Baseline from 1.43\%$\rightarrow$0.27\% and 6.23\%$\rightarrow$0.49\% for \texttt{Gender} and \texttt{BlackHair} on celebA respectively, and 3.52\%$\rightarrow$0.75\% for \texttt{Cat/Dog} on AFHQ. Additionally, note how measurement error in Baseline and Diversity increases by increasing 
%the bias in the data
the data bias, while CLEAM remains consistently low.
See Sec.
\ref{subsec:pseudoGMoreLA} and \ref{subsec:pseudoGMoreClassifiers} 
for analysis with more attributes and classifiers.

%\subsection{Evaluating CLEAM with different $n$} 
%\textcolor{blue}{\bf Chris: Should we move this before Sec 5.2, so that we dont have to keep jumping between setups?}
%\label{subsec:CLEAMwithDiffn}

%Chris: Merged above to one row
\iffalse
\begin{figure}[t!]
    \centering
    \includegraphics[width=\linewidth]{figures/VaryingN_ResNet18_v2.pdf}
    \caption{Comparing the point error $e_{\mu}$ for Baseline and CLEAM when evaluating the bias of GenData
    %the generated data 
    with ResNet-18, with varying sample size, $n$. }
    \label{fig:ResNet18Permuten}
    \vspace{-0.5cm}
\end{figure}
\fi

%%%%%%%%%%%%%%%%%%%%%%%%%%%%%%%%%%%%
%       Experiment Pseudo G
%%%%%%%%%%%%%%%%%%%%%%%%%%%%%%%%%%%%
\begin{table}[h]
    \centering
    %New overall Caption
    \caption{
    Comparing the {\em point estimates} and {\em interval estimate}  of Baseline \cite{choiFairGenerativeModeling2020a}, Diversity \cite{keswaniAuditingDiversityUsing2021} and CLEAM 
    %measurement frameworks 
    in estimating different $\bpstar$ of a {\em pseudo-generator}, based on CelebA \cite{liuDeepLearningFace2015} and AFHQ \cite{choiStarGANV2Diverse2020}, for sensitive attribute \texttt{Gender}, \texttt{BlackHair} and \texttt{Cat/Dog}. The $\bphat$ is computed with a ResNet-18 and
    the error rate is reported per
    Eqn.1 of the main manuscript
    %Eqn.\ref{eqn:pointError}. % and \ref{eqn:intervalError}.
    }
\resizebox{\textwidth}{!}{
    \addtolength{\tabcolsep}{-5pt}
      \begin{tabular}
      %{c  c @{\hspace{0.5em}} c @{\hspace{1em}} c @{\hspace{0.5em}} c @{\hspace{1em}} c @{\hspace{0.5em}} c}
      {c  cc cc cc @{\hspace{5pt}} cc cc cc}
        \toprule
        %\multicolumn{7}{c}{\bf (A) Pseudo-Generator}\\
        %\midrule
        & \multicolumn{6}{c}{\bf Point Estimate} & \multicolumn{6}{c}{\bf Interval Estimate}\\
        \cmidrule(lr){2-7}\cmidrule(lr){8-13}
          \textbf{GT} & \multicolumn{2}{c}{\textbf{Baseline}
          %\cite{choiFairGenerativeModeling2020a}
          }& \multicolumn{2}{c}{\textbf{Diversity}
          %\cite{keswaniAuditingDiversityUsing2021}
          } & \multicolumn{2}{c}{\textbf{CLEAM (Ours)}}
          & \multicolumn{2}{c}{\textbf{Baseline}
          %\cite{choiFairGenerativeModeling2020a}
          }& \multicolumn{2}{c}{\textbf{Diversity}
          %\cite{keswaniAuditingDiversityUsing2021}
          } & \multicolumn{2}{c}{\textbf{CLEAM (Ours)}}
        \\
         \cmidrule(lr){1-1}\cmidrule(lr){2-3}\cmidrule(lr){4-5} \cmidrule(lr){6-7} 
         \cmidrule(lr){1-1}\cmidrule(lr){8-9}\cmidrule(lr){10-11} \cmidrule(lr){12-13}
         %Baseline
        &$\mu_{\texttt{Base}}$ & $e_\mu (\downarrow)$ 
         %Diversity
         &$\mu_{\texttt{Div}}$ & $e_\mu(\downarrow)$ 
         %CLEAM
        &$\mu_{\texttt{CLEAM}}$ & $e_\mu(\downarrow)$
         %Baseline
        &$\rho_{\texttt{Base}}$ & $e_\rho (\downarrow)$ 
         %Diversity
         &$\rho_{\texttt{Div}}$ & $e_\rho(\downarrow)$ 
         %CLEAM
        &$\rho_{\texttt{CLEAM}}$ & $e_\rho(\downarrow)$
        \\
        \midrule
        \multicolumn{13}{c}{$\balpha$=[0.976,0.979], \texttt{Gender} (CelebA)}\\
        \midrule
        %Point Estimate--------------------------------------------------
        $p^*_0$=0.9 & 
          0.880 &   2.22\% & 
        %Diversity
          0.950 &   5.55\% & 
        %CLEAM
          0.899 &   \textbf{0.11\%} &
        %Interval Estimate--------------------------------------------------
        [0.876 , 0.884 ] & 2.67\% &
        %Diversity
        [0.913 , 0.986 ] & 9.56\% &
        %CLEAM
        [0.895, 0.904] & \textbf{0.56}\% 
        \\  %\hdashline
        %Point Estimate--------------------------------------------------
        $p^*_0$=0.8 & 
          0.783 &   2.10\%&
        %Diversity
          0.785 &   1.88\% &
        %CLEAM
          0.798 & 
          \textbf{0.25}\%  &
        %Interval Estimate--------------------------------------------------
        [ 0.778   , 0.788  ] & 2.75\%
        %Diversity
        & [0.762 , 0.809 ] & 4.75\%
        %CLEAM
        & [0.794$, $0.803] & \textbf{0.75}\%
        \\ %\hdashline
        %Point Estimate--------------------------------------------------
        $p^*_0$=0.7 & 
          0.691 &   1.29\% & 
        %Diversity
          0.709 &   1.29\% & 
        %CLEAM
          0.701  &   \textbf{0.14}\% &
        %Interval Estimate--------------------------------------------------
        [ 0.687  , 0.695  ] & 1.86\% &
        %Diversity
        [0.696 , 0.722 ] & 3.14\% &
        %CLEAM
        [0.697, 0.707]  & \textbf{0.10}\%
        \\ %\hdashline
         %Point Estimate--------------------------------------------------
        $p^*_0$=0.6 & 
          0.592 &   1.33\% & 
        %Diversity
          0.591 &   1.50\% &
        %CLEAM
          0.597 &   \textbf{0.50}\% &
        %Interval Estimate--------------------------------------------------
        [0.586  , 0.598 ] & 2.33\%&
        %Diversity
        [0.581 , 0.612 ] & 3.17\% &
        %CLEAM
        [0.591,0.603]   & \textbf{1.50}\% 
        \\ %\hdashline
        $p^*_0$=0.5 & 
        %Point Estimate--------------------------------------------------
          0.501 &   \textbf{0.20}\% &
        %Diversity
          0.481 &   3.80\% & 
        %CLEAM
          0.502  &   0.40\%  &
        %Interval Estimate--------------------------------------------------
        [0.495  , 0.507 ] & \textbf{1.40}\% &
        %Diversity
        [0.473 , 0.490 ] & 5.40\% &
        %CLEAM
        [0.497, 0.508]  & 1.60\%
        \\
        \midrule
        %Point Estimate--------------------------------------------------
          \multicolumn{2}{c}{Average Error:} &   1.43\% 
        &     &   2.80\% 
        &     &   \textbf{0.27}\%
        %Interval Estimate--------------------------------------------------
        &    & $2.20\%$ 
        &  & $5.20\%$
        &  & $\mathbf{0.90}\%$ 
        \\
        \midrule
        \multicolumn{13}{c}{$\balpha$=[0.881,0.887], \texttt{BlackHair} (CelebA)}\\
        \midrule
        %Point Estimate--------------------------------------------------
        $p^*_0$=0.9 & 
          0.803 &   10.77\%& 
        %Diversity
          0.803 &   10.77\% & 
        %CLEAM
          0.899 &   \textbf{0.11}\% &
        %Interval Estimate--------------------------------------------------
        [ 0.800  , 0.806 ] & 11.11\% &
        %Diversity
        $[0.791 , 0.815 ]$ & 12.11\% &
        %CLEAM
       [0.893, 0.905] & \textbf{0.78}\% 
        \\ %\hdashline
        %Point Estimate--------------------------------------------------
        $p^*_0$=0.8 & 
          0.723 &   9.63\% & 
        %Diversity
          0.699 &   12.63\% &
        %CLEAM
          0.796 &   \textbf{0.50}\% &
        %Interval Estimate--------------------------------------------------
        [0.719  , 0.727 ] & 10.13\% &
        %Diversity
        [0.686 , 0.713 ] & 14.25\% &
        %CLEAM
        [0.790, 0.803] & \textbf{1.25}\%
        \\ %\hdashline
        %Point Estimate--------------------------------------------------
        $p^*_0$=0.7 & 
          0.654 &   6.57\% & 
        %Diversity
          0.661 &   5.57\% &
        %CLEAM
          0.705  &   \textbf{0.71}\% &
        %Interval Estimate--------------------------------------------------
        [ 0.648  , 0.660  ] & 7.43\%  &
        %Diversity
        [ 0.643 , 0.68 ] & 8.14\% &
        %CLEAM
        [0.698, 0.712]  & \textbf{1.71}\%
        \\ %\hdashline
        %Point Estimate--------------------------------------------------
        $p^*_0$=0.6 & 
          0.575 &   4.17\% & 
        %Diversity
          0.609 &   1.50\% & 
        %CLEAM
          0.602 &   \textbf{0.33}\%  &
        %Interval Estimate--------------------------------------------------
        [ 0.564  , 0.586  ] & 6.00\% &
        %Diversity
        [0.604 , 0.614 ] & 2.30\% &
        %CLEAM
        [0.599, 0.606]   & \textbf{1.00}\% 
        \\ %\hdashline
        %Point Estimate--------------------------------------------------
        $p^*_0$=0.5 & 
          0.500 &   \textbf{0.00}\% & 
        %Diversity
          0.521 &   4.20\% & 
        %CLEAM
          0.504  &   0.8\%  &
        %Interval Estimate--------------------------------------------------
        [ 0.495  , 0.505  ] & \textbf{1.00}\% &
        %Diversity
        [0.506 , 0.536 ] & 7.20\% &
        %CLEAM
        [0.497, 0.511]  & 2.20\% \\
        \hline 
         %Point Estimate--------------------------------------------------
          \multicolumn{2}{c}{Average Error:} &   6.23\% &     &   6.93\% &     &   \textbf{0.49}\%
        %Interval Estimate--------------------------------------------------
        &   &7.13\%
        &  & 8.80\%
        &  & \textbf{1.39}\% \\
        \midrule
        \multicolumn{13}{c}{$\balpha$=[0.953,0.0.990], \texttt{Cat/Dog} (AFHQ)}\\
        \midrule
        %Point Estimate--------------------------------------------------
        $p^*_0$=0.9 & 
          0.862  &   4.44\%& 
        %Diversity
          0.855 &   5.00\% & 
        %CLEAM
          0.903 &   \textbf{0.33}\%  &
        %Interval Estimate--------------------------------------------------
         [ 0.859 , 0.865 ] & 4.56\% &
        %Diversity
        [ 0.844 , 0.866  ]  & 6.22\% &
        %CLEAM
       [ 0.900 , 0.907 ] & \textbf{0.78}\% 
        \\ %\hdashline
        %Point Estimate--------------------------------------------------
        $p^*_0$=0.8 & 
           0.766 &   4.25\% & 
        %Diversity
          0.774 &   3.25\% &
        %CLEAM
          0.802 &   \textbf{0.25}\% &
        %Interval Estimate--------------------------------------------------
         [ 0.762 , 0.771 ] & 4.75\% &
        %Diversity
         [ 0.765 , 0.784 ] & 4.38\% &
        %CLEAM
        [ 0.797 , 0.807 ] & \textbf{0.88}\%
        \\ %\hdashline
        %Point Estimate--------------------------------------------------
        $p^*_0$=0.7 & 
          0.677 &   3.29\% & 
        %Diversity
          0.670 &   4.29\% &
        %CLEAM
          0.707  &   \textbf{1.00}\% &
        %Interval Estimate--------------------------------------------------
        [ 0.672 , 0.682 ] & 4.00\%  &
        %Diversity
         [ 0.655, 0.686 ] & 6.43\% &
        %CLEAM
        [ 0.701 , 0.712 ]   & \textbf{1.71}\%
        \\ %\hdashline
         %Point Estimate--------------------------------------------------
        $p^*_0$=0.6 & 
          0.583 &   2.83\% & 
        %Diversity
          0.551 &   8.17\% & 
        %CLEAM
          0.607 &   \textbf{1.17}\%  &
        %Interval Estimate--------------------------------------------------
        [ 0.578 , 0.588 ] & 3.67\% &
        %Diversity
        [ 0.540, 0.562  ]  & 10.00\% &
        %CLEAM
        [ 0.602 , 0.613 ]   & \textbf{2.17}\% 
        \\ %\hdashline
        %Point Estimate--------------------------------------------------
        $p^*_0$=0.5 & 
          0.486  &   2.80\% & 
        %Diversity
          0.469 &   6.20\% & 
        %CLEAM
          0.505  &   \textbf{1.00}\%  &
        %Interval Estimate--------------------------------------------------
        [ 0.480 , 0.493 ] & 4.00\% &
        %Diversity
        [ 0.458, 0.480  ]  & 8.40\% &
        %CLEAM
        [ 0.498 , 0.511 ]  & \textbf{2.20}\% \\
        \hline 
          \multicolumn{2}{c}{Average Error:} &   3.52\% &     &   5.38\% &     &   \textbf{0.75}\%
        &  & 4.20\% &  & 7.09\% &  & \textbf{1.55}\%\\
        \bottomrule
    \end{tabular}
    \addtolength{\tabcolsep}{5pt}
    }
    \label{tab:fakeG_PE}
\end{table}
\clearpage
\FloatBarrier
\subsection{Measuring Varying Degrees of Bias with Additional Sensitive Attributes}
\label{subsec:pseudoGMoreLA}

In 
Sec. \ref{sec:psuedoG},
%Sec. 5.3 of the main paper, 
we demonstrate CLEAM's ability to improve accuracy in approximating $\bpstar$ for the sensitive attributes $\texttt{Gender}$ and $\texttt{BlackHair}$. In this section, we extend the experiment on 
%a new dataset 1) AFHQ with sensitive attribute \texttt{Cat/Dog} and 2) the same 
CelebA dataset but with harder (lower $\balpha$) sensitive attributes \ie $\texttt{Young}$, and $\texttt{Attractive}$. 
We did not include Diversity in this study, due to its poor performance on harder sensitive attribute, as discussed in \ref{sec:DiversitySetup}.

\textbf{ From our results in 
Tab. \ref{Tab:YoungAndAttractivePEIE}}, both $\texttt{Young}$ and $\texttt{Attractive}$ classifiers have relatively large errors ($\emubase$) in the baseline \ie on average 17.63\% and 12.69\%, respectively. Then utilizing CLEAM, even with the harder sensitive attributes, we are able to significantly reduce these errors to $0.68\%$ and $0.94\%$. See Sec. \ref{sec:DeeperAnlaysis} for more details regarding the effect that the different degrees of inaccuracies in $\balpha$ have on the Baseline error. 

\begin{table*}[h]
    \centering
    %New overall Caption
    \caption{Comparing \textbf{point estimate} and \textbf{interval estimate} of 
    Baseline \cite{choiFairGenerativeModeling2020a},
    and proposed CLEAM measurement framework on a pseudo-generator with sensitive attribute $\{\texttt{Young},\texttt{Attractive}\}$
    }
\resizebox{0.98\linewidth}{!}{
    \addtolength{\tabcolsep}{-2pt}
      \begin{tabular}
      %{c  c @{\hspace{0.5em}} c @{\hspace{1em}} c @{\hspace{0.5em}} c @{\hspace{1em}} c @{\hspace{0.5em}} c}
      {c  cc cc cc @{\hspace{1cm}} cc cc cc}
        \toprule
        %\multicolumn{7}{c}{\bf (A) Pseudo-Generator}\\
        %\midrule
        & \multicolumn{6}{c}{\bf Point Estimate} & \multicolumn{6}{c}{\bf Interval Estimate}\\
        \cmidrule(lr){2-7}\cmidrule(lr){8-13}
          \textbf{GT} & \multicolumn{2}{c}{\textbf{Baseline}
          %\cite{choiFairGenerativeModeling2020a}
          }& \multicolumn{2}{c}{\textbf{Diversity}
          %\cite{keswaniAuditingDiversityUsing2021}
          } & \multicolumn{2}{c}{\textbf{CLEAM (Ours)}}
          & \multicolumn{2}{c}{\textbf{Baseline}
          %\cite{choiFairGenerativeModeling2020a}
          }& \multicolumn{2}{c}{\textbf{Diversity}
          %\cite{keswaniAuditingDiversityUsing2021}
          } & \multicolumn{2}{c}{\textbf{CLEAM (Ours)}}
        \\
         \cmidrule(lr){1-1}\cmidrule(lr){2-3}\cmidrule(lr){4-5} \cmidrule(lr){6-7} 
         \cmidrule(lr){1-1}\cmidrule(lr){8-9}\cmidrule(lr){10-11} \cmidrule(lr){12-13}
         %Baseline
        &$\mu_{\texttt{Base}}$ & $e_\mu (\downarrow)$ 
         %Diversity
         &$\mu_{\texttt{Div}}$ & $e_\mu(\downarrow)$ 
         %CLEAM
        &$\mu_{\texttt{CLEAM}}$ & $e_\mu(\downarrow)$
         %Baseline
        &$\rho_{\texttt{Base}}$ & $e_\rho (\downarrow)$ 
         %Diversity
         &$\rho_{\texttt{Div}}$ & $e_\rho(\downarrow)$ 
         %CLEAM
        &$\rho_{\texttt{CLEAM}}$ & $e_\rho(\downarrow)$
        \\
                \midrule
        \multicolumn{13}{c}{$\balpha$=[0.749,0.852], \texttt{Young}}\\
        \midrule
         $p^*_0=0.9$ & 
         %Point estimate---------------------------------------------
        %Baseline
        0.690 & 23.33\%  & 
        %Diversity
        --- & --- & 
        %CLEAM
        0.905& \textbf{0.56}\%  &
        %Interval estimate---------------------------------------------
        %Baseline
        [0.684,0.695] & 24.00\% & 
        %Diversity
         --- & ---  & 
        %CLEAM
        [0.890,0.920] & \textbf{2.22}\%  
        \\  %\hdashline
        $p^*_0=0.8$ & 
        %Point estimate---------------------------------------------
        %Baseline
        0.630 & 21.25\%  &  
        %Diversity
        --- & --- & 
        %CLEAM
        0.804 & \textbf{0.50}\%  &
        %Interval estimate---------------------------------------------
        %Baseline
        [0.625,0.635] & 21.88\%   & 
        %Diversity
          --- &  --- & 
        %CLEAM
        [0.795,0.813] & \textbf{1.63}\%  
        \\ %\hdashline
        $p^*_0=0.7$ & 
        %Point estimate---------------------------------------------
        %Baseline
        0.570 & 18.57\%  & 
        %Diversity
        --- & ---  & 
        %CLEAM
        0.698 & \textbf{0.29}\% &
        %Interval estimate---------------------------------------------
        %Baseline
        [0.565,0.575] & 19.29\% & 
        %Diversity
        --- & ---  & 
        %CLEAM
        [0.690,0.706] & \textbf{1.43}\% 
        \\ %\hdashline
        $p^*_0=0.6$ & 
        %Point estimate---------------------------------------------
        %Baseline
        0.510 & 15.00\%  & 
        %Diversity
        --- & ---  & 
        %CLEAM
        0.595 & \textbf{0.83}\% &
        %Interval estimate---------------------------------------------
        %Baseline
        [0.505,0.515] &  15.83\% & 
        %Diversity
        --- & ---  & 
        %CLEAM
        [0.590,0.600]  & \textbf{1.67}\%
        \\ %\hdashline
        $p^*_0=0.5$ & 
        %Point estimate---------------------------------------------
        %Baseline
        0.450 & 10.0\%  & 
        %Diversity
        --- & --- &
        %CLEAM
        0.506 & \textbf{1.20}\% &
        %Interval estimate---------------------------------------------
        %Baseline
        [0.445,0.455] & 11.00\%  & 
        %Diversity
        --- & ---  & 
        %CLEAM
        [0.502,0.510] & \textbf{2.00}\%
        \\ \midrule
        %Point estimate---------------------------------------------
        \multicolumn{2}{c}{Avg Error}& 17.63\%
        && ---\% 
        && \textbf{0.68}\% 
        %Interval estimate---------------------------------------------
        &  & 18.40\% 
        &  & ---\% 
        &  & \textbf{1.79}\% \\
        \midrule
        \multicolumn{13}{c}{$\balpha$=[0.780,0.807], \texttt{Attractive}}\\
        \midrule
        %Point estimate---------------------------------------------
        $p^*_0=0.9$ & 
        %Baseline
        0.730 & 18.89\%  & 
        %Diversity
        --- & ---  & 
        %CLEAM
        0.908 & \textbf{0.89}\% &
        %Interval estimate---------------------------------------------
        %Baseline
        [0.724,0.736] & 19.56\%  & 
        %Diversity
        --- & ---  & 
        %CLEAM
        [0.900,0.916] &  \textbf{1.78}\%
        \\ %\hdashline
        $p^*_0=0.8$ & 
        %Point estimate---------------------------------------------
        %Baseline
        0.670 & 16.25\%  & 
        %Diversity
        --- & ---  & 
        %CLEAM
        0.804 & \textbf{0.50}\% &
        %Interval estimate------------------------------------------
        %Baseline
        [0.665,0.675] & 16.88\% & 
        %Diversity
        --- & ---  &  
        %CLEAM
        [0.795,0.813] & \textbf{1.63}\%
        \\ %\hdashline
        $p^*_0=0.7$ & 
        %Point estimate---------------------------------------------
        %Baseline
        0.600 & 14.29\%  & 
        %Diversity
        --- &  --- & 
        %CLEAM
        0.696 & \textbf{0.57}\% &
        %Interval estimate---------------------------------------------
        %Baseline
        [0.594,0.606] & 15.14\% & 
        %Diversity
        --- & ---  & 
        %CLEAM
        [0.690,0.712] & \textbf{1.71}\%
        \\ %\hdashline
        $p^*_0=0.6$ & 
        %Point estimate---------------------------------------------
        %Baseline
        0.540 & 10.00\% & 
        %Diversity
        --- & ---  & 
        %CLEAM
        0.592  & \textbf{1.33}\% &
        %Interval estimate---------------------------------------------
        %Baseline
        [0.534,0.546] & 11.00\% & 
        %Diversity
        --- & ---  & 
        %CLEAM
        [0.580,0.604] & \textbf{3.33}\%
        \\ %\hdashline
        $p^*_0=0.5$ & 
        %Point estimate---------------------------------------------
        0.480 & 4.00\%  & 
        %Diversity
        --- & ---  & 
        %CLEAM
        0.493 & \textbf{1.40}\%  &
        %Interval estimate---------------------------------------------
        [0.475,0.485] & 5.00\% & 
        %Diversity
        --- & ---  & 
        %CLEAM
        [0.487,0.499] & \textbf{2.60}\% 
        \\
        \hline 
        %Point estimate---------------------------------------------
        \multicolumn{2}{c}{Avg Error}& 12.69\% 
        && ---\% 
        && \textbf{0.94}\%
        %Interval estimate---------------------------------------------
        && 13.52\% &
         & ---\% &
         & \textbf{2.22}\%
        \\
        \bottomrule
    \end{tabular}
    \addtolength{\tabcolsep}{2pt}
    }
    \label{Tab:YoungAndAttractivePEIE}
\end{table*}

\FloatBarrier
\clearpage
\subsection{Measuring Varying Degrees of Bias with Additional Sensitive Attribute Classifier}
\label{subsec:pseudoGMoreClassifiers}
%\textcolor{red}{To update Discussion with AFHQ dataset}
In this section, we validate CLEAM's versatility with different sensitive attribute classifier architectures. In our setup, we utilise MobileNetV2 \cite{sandlerMobileNetV2InvertedResiduals2018} as in  \cite{frankelFairGenerationPrior2020}. 
Then similar to Sec. \ref{sec:psuedoG}, we utilize a pseudo-generator with known GT $\bpstar$ for \texttt{Gender} and \texttt{BlackHair} sensitive attribute from the CelebA \cite{liuDeepLearningFace2015} dataset, and \texttt{Cat/Dog} from the AFHQ \cite{choiStarGANV2Diverse2020} dataset, to evaluate CLEAM's effectiveness at determining bias.
%We then measure the known GT $\pstar$ from the pseudo-generator with LA $\texttt{Gender}$ and $\texttt{BlackHair}$, as per Sec 5 of the main paper. Note that, as per MobileNetV2 architecture requirements, we resize the input images into $244$x$244$. 

As seen in our results in Tab.\ref{MobileNetPEIE}, MobileNetV2 achieved reasonably high average accuracy $\in$[0.889,0.983]. Then, when evaluating $\bpstar$ of the pseudo-generator we observed similar behavior to ResNet-18 discussed in 
%the main paper.
sec. \ref{sec:psuedoG}.
In particular, we observed a significantly large $\emubase$ (for the baseline) of $1.42\%$, $9.74\%$ and $2.81\%$ for the \texttt{Gender}, \texttt{BlackHair} and \texttt{Cat/Dog}  sensitive attribute, respectively. Whereas, CLEAM reported an $\emucleam$ of $0.13\%$, $0.7\%$ and $0.62\%$, respectively. The same trend can be observed in 
the IE.
%$e_\rho$.
We thus demonstrate CLEAM's versatility and ability to be deployed as a post-processing method (without retraining), on models of varying architecture.

\begin{table*}[h]
    \centering
    %New overall Caption
    \caption{
    Comparing the {\em point estimates} and {\em interval estimate}  of Baseline \cite{choiFairGenerativeModeling2020a}, Diversity \cite{keswaniAuditingDiversityUsing2021} and proposed CLEAM measurement frameworks in estimating different $\bpstar$ of a {\em pseudo-generator}, based on the CelebA \cite{liuDeepLearningFace2015} and AFHQ \cite{choiStarGANV2Diverse2020} dataset. The $\bphat$ is computed with a MobileNetV2\cite{sandlerMobileNetV2InvertedResiduals2018} classifier and
    the error rate is reported using Eqn. 1 of the main manuscript.
    We repeat this on \texttt{Gender} (CelebA), \texttt{BlackHair} (CelebA) and \texttt{Cat/Dog}(AFHQ)  sensitive attribute.
    }
\resizebox{0.98\linewidth}{!}{
    \addtolength{\tabcolsep}{-2pt}
      \begin{tabular}
      %{c  c @{\hspace{0.5em}} c @{\hspace{1em}} c @{\hspace{0.5em}} c @{\hspace{1em}} c @{\hspace{0.5em}} c}
      {c @{\hspace{8pt}} c@{\hspace{-0.1pt}}c c@{\hspace{-1pt}}c c@{\hspace{-1pt}}c @{\hspace{0.5cm}} cc cc cc}
        \toprule
        %\multicolumn{7}{c}{\bf (A) Pseudo-Generator}\\
        %\midrule
        & \multicolumn{6}{c}{\bf Point Estimate} & \multicolumn{6}{c}{\bf Interval Estimate}\\
        \cmidrule(lr){2-7}\cmidrule(lr){8-13}
          \textbf{GT} & \multicolumn{2}{c}{\textbf{Baseline}
          %\cite{choiFairGenerativeModeling2020a}
          }& \multicolumn{2}{c}{\textbf{Diversity}
          %\cite{keswaniAuditingDiversityUsing2021}
          } & \multicolumn{2}{c}{\textbf{CLEAM (Ours)}}
          & \multicolumn{2}{c}{\textbf{Baseline}
          %\cite{choiFairGenerativeModeling2020a}
          }& \multicolumn{2}{c}{
          \textbf{Diversity}
          %\cite{keswaniAuditingDiversityUsing2021}
          } & \multicolumn{2}{c}{\textbf{CLEAM (Ours)}}
        \\
         \cmidrule(lr){1-1}\cmidrule(lr){2-3}\cmidrule(lr){4-5} \cmidrule(lr){6-7} 
         \cmidrule(lr){1-1}\cmidrule(lr){8-9}\cmidrule(lr){10-11} \cmidrule(lr){12-13}
         %Baseline
        &$\mu_{\texttt{Base}}$ & $e_\mu (\downarrow)$ 
         %Diversity
         &$\mu_{\texttt{Div}}$ & $e_\mu(\downarrow)$ 
         %CLEAM
        &$\mu_{\texttt{CLEAM}}$ & $e_\mu(\downarrow)$
         %Baseline
        &$\rho_{\texttt{Base}}$ & $e_\rho (\downarrow)$ 
         %Diversity
         &$\rho_{\texttt{Div}}$ & $e_\rho(\downarrow)$ 
         %CLEAM
        &$\rho_{\texttt{CLEAM}}$ & $e_\rho(\downarrow)$
        \\
        \midrule
        \multicolumn{13}{c}{$\balpha$=[0.980,0.986], \texttt{Gender} (CelebA)}\\
        \midrule
        $p^*_0=0.9$ & 
        %Point Estimate ---------------------------------------
        %Baseline
        0.882 & 2.00\%  & 
        %Diversity
        0.950 & 5.55\% & 
        %CLEAM
        0.899 & \textbf{0.11}\% &
        %Interval Estimate ---------------------------------------
        %Baseline
        [ 0.879 , 0.885 ] & 2.33\% & 
        %Diversity
        [0.913 , 0.986 ] & 9.56\% &
        %CLEAM
        [0.895,0.902] & \textbf{0.56}\%  
        \\ %\hdashline
        %Point Estimate ---------------------------------------
        $p^*_0=0.8$ & 
        %Baseline
         0.786 & 1.75\%  & 
        %Diversity
        0.785 & 1.88\% &
        %CLEAM
         0.800 & $\mathbf{0.00}\%$ &
         %Interval Estimate ---------------------------------------
          %Baseline
        [ 0.782 , 0.790 ] & 2.25\% & 
        %Diversity
        [0.762 , 0.809 ] & 4.75\% &
        %CLEAM
        [0.794,0.804] & \textbf{0.75}\%  
        \\ %\hdashline
        $p^*_0=0.7$ & 
        %Point Estimate ---------------------------------------
        %Baseline
        0.689 & 1.57\%  & 
        %Diversity
        0.709 & 1.30\% & 
        %CLEAM
        0.699 & \textbf{0.14}\% &
        %Interval Estimate ---------------------------------------
         %Baseline
        [ 0.685, 0.693 ] & 2.14\% & 
        %Diversity
        [0.696 , 0.722 ] & 3.14\% &
        %CLEAM
        [0.694,0.704] & \textbf{0.86}\%
        \\ %\hdashline
         %Point Estimate ---------------------------------------
        $p^*_0=0.6$ & 
        %Baseline
        0.593 & 1.17\%  & 
        %Diversity
        0.591 & 1.50\% &
        %CLEAM
        0.600 & \textbf{0.00}\% &
        %Interval Estimate ---------------------------------------
        %Baseline
        [ 0.585 , 0.597 ] & 2.50\%  & 
        %Diversity
        [0.581 , 0.612 ] & 3.17\% &
        %CLEAM
        [594,0.605] & \textbf{1.00}\%
        \\ %\hdashline
        $p^*_0=0.5$ & 
        %Point Estimate ---------------------------------------
        0.497 & 0.60\%  & 
        %Diversity
        0.481 & 3.80\% & 
        %CLEAM
        0.502 & \textbf{0.40}\% &
        %Interval Estimate ---------------------------------------
        %Baseline
        [ 0.491 , 0.502 ] & 1.80\%  & 
        %Diversity
        [0.473 , 0.490 ] & 5.40\% &
        %CLEAM
        [495,0.507] & \textbf{1.40}\%
        \\
        \hline 
        %Point Estimate ---------------------------------------
        \multicolumn{2}{c}{Avg Error} & 1.42\% && 2.81\% && \textbf{0.13}\%
        %Interval Estimate ---------------------------------------
        && 2.20\% 
        && 5.20\% 
        && \textbf{0.91}\%\\
        \midrule
        \multicolumn{13}{c}{$\balpha$=[0.861,0.916], \texttt{BlackHair} (CelebA)}\\
        \midrule
        $p^*_0=0.9$ &
        %Point Estimate ---------------------------------------
        %Baseline
        0.782 & 13.11\%  & 
        %Diversity
        0.803 & 10.78\% & 
        %CLEAM
        0.899 & \textbf{0.11}\% &
        %Interval Estimate ---------------------------------------
        %Baseline
        [ 0.777 , 0.787 ] & 13.67\%  & 
        %Diversity
        [0.791 , 0.815 ] & 9.44\% &
        %CLEAM
        [0.893,0.900] & \textbf{0.78}\%
        \\ %\hdashline
        %Point Estimate ---------------------------------------
        $p^*_0=0.8$ & 
        %Baseline
         0.705 & 11.88\%  & 
        %Diversity
        0.699 & 12.63\% &
        %CLEAM
         0.800 & \textbf{0.00}\% &
        %Interval Estimate ---------------------------------------
        %Baseline
        [ 0.699 , 0.710 ] & 12.63\% & 
        %Diversity
        [0.686 , 0.713 ] & 14.25\% &
        %CLEAM
        [0.793,0.807] & $\textbf{0.88}\%$
        \\ %\hdashline
        %Point Estimate ---------------------------------------
        $p^*_0=0.7$ & 
        %Baseline
        0.623 & 11.00\%  & 
        %Diversity
        0.661 & 5.56\% &
        %CLEAM
        0.700 & \textbf{0.00}\% &
        %Interval Estimate ---------------------------------------
        %Baseline
        [ 0.618 , 0.628 ] & 11.71\%  & 
        %Diversity
        [ 0.643 , 0.68 ] & 8.14\% &
        %CLEAM
        [0.694,0.706] & \textbf{0.86}\%
        \\ %\hdashline
        $p^*_0=0.6$ & 
        %Point Estimate ---------------------------------------
        %Baseline
        0.550 & 8.33\%  & 
        %Diversity
        0.609 & 1.50\% & 
        %CLEAM
        0.600 & \textbf{0.00}\% &
        %Interval Estimate ---------------------------------------
        %Baseline
        [ 0.544 , 0.556 ] & 9.33\% & 
        %Diversity
        [0.604 , 0.614 ] & 2.33\% &
        %CLEAM
        [0.593,0.608] & \textbf{1.17}\%
        \\ %\hdashline
        $p^*_0=0.5$ & 
         %Point Estimate ---------------------------------------
        0.478 & 4.40\%  & 
        %Diversity
        0.521 & 4.20\% & 
        %CLEAM
        0.506 & \textbf{1.20}\%  &
        %Interval Estimate ---------------------------------------
        [ 0.472 , 0.484 ] & 5.60\% & 
        %Diversity
        [0.506 , 0.536 ] & 7.20\% &
        %CLEAM
        [0.498,0.514] & \textbf{2.80}\%
        \\
        \hline 
        %Point Estimate ---------------------------------------
        \multicolumn{2}{c}{Avg Error}& 9.74\% && 6.93\% &&  \textbf{0.70}\%
        %Interval Estimate ---------------------------------------
        && $10.59\%$ &&$8.27\%$ &&$\mathbf{1.30}\%$
        \\
        \bottomrule
        \midrule
        \midrule
        \multicolumn{13}{c}{$\balpha$=[0.964,0.897], \texttt{Cat/Dog} (AFHQ)}\\
        \midrule
        $p^*_0=0.9$ &
        %Point Estimate ---------------------------------------
        %Baseline
        0.875 & 2.77\%  & 
        %Diversity
        0.880 & 3.26\% & 
        %CLEAM
        0.897 & \textbf{0.34}\% &
        %Interval Estimate ---------------------------------------
         %Baseline
        [ 0.872 , 0.878 ]  & 3.07\%  & 
        %Diversity
        [0.871 , 0.890] & 3.25\% &
        %CLEAM
        [ 0.894 , 0.900 ] & \textbf{0.68}\%
        \\ %\hdashline
        $p^*_0=0.8$ &
        %Point Estimate ---------------------------------------
        %Baseline
         0.784 & 2.00\%  & 
        %Diversity
        0.770 & 3.75\% &
        %CLEAM
         0.791 & \textbf{1.11}\% &
         %Interval Estimate ---------------------------------------
         %Baseline
        [ 0.780 , 0.788 ] & 2.53\% & 
        %Diversity
        [0.759 , 0.781 ] & 5.12\% &
        %CLEAM
        [ 0.786 , 0.796 ] & \textbf{0.42}\%
        \\ %\hdashline
        $p^*_0=0.7$ & 
        %Point Estimate ---------------------------------------
        %Baseline
        0.704 & 0.62\%  & 
        %Diversity
        0.692 & 1.08\% &
        %CLEAM
        0.698 & \textbf{0.20}\% &
        %Interval Estimate ---------------------------------------
         [ 0.700 , 0.708 ] & 1.19\%  & 
        %Diversity
        [ 0.684, 0.709 ] & 2.40\% &
        %CLEAM
        [ 0.694 , 0.703 ] & \textbf{0.86}\%
        \\ %\hdashline
        $p^*_0=0.6$ & 
        %Point Estimate ---------------------------------------
        %Baseline
        0.617 & 2.78\%  & 
        %Diversity
        0.611 & 1.83\% & 
        %CLEAM
        0.597 & \textbf{0.54}\% &
        %Interval Estimate ---------------------------------------
        %Baseline
        [ 0.611 , 0.622 ] & 2.78\% & 
        %Diversity
        [0.602 , 0.620 ] & 3.42\% &
        %CLEAM
        [ 0.591 , 0.603 ] & \textbf{1.58}\%
        \\ %\hdashline
        $p^*_0=0.5$ & 
        %Point Estimate ---------------------------------------
        0.529 & 5.87\%  & 
        %Diversity
        0.536 & 7.20\% & 
        %CLEAM
        0.495 & \textbf{0.93}\% &
        %Interval Estimate ---------------------------------------
        [ 0.523 , 0.536 ] & 7.17\% & 
        %Diversity
        [0.524 , 0.548 ] & 9.68\% &
        %CLEAM
        [ 0.488 , 0.503 ] & $\mathbf{2.44}\%$
        \\
        \hline 
        %Point Estimate ---------------------------------------
        \multicolumn{2}{c}{Avg Error}& 2.81\% && 3.42\% & & \textbf{0.62}\%
        %Interval Estimate ---------------------------------------
        && 3.35\% &Avg Error & 4.77\% && \textbf{1.20}\%
        \\
        \bottomrule
    \end{tabular}
    \addtolength{\tabcolsep}{2pt}
    }
    \label{MobileNetPEIE}
\end{table*}
\clearpage
\subsection{Measuring SOTA GANs 
%\textcolor{blue}{
and Diffusion Models
%} 
with Additional Classifier}

%\textcolor{red}{[To include CLIP being utilized as a classifier on SOTA GANs][{\bf Addressed}]} 

%\textcolor{blue}{
In this section, we further explore the utilization of CLIP as a sensitive attribute classifier; more details on CLIP in Sec. \ref{sec:applicationOfCLEAM}. Here, we follow the setup in Sec. 5.1 of our main manuscript to measure the bias in GenData-StyleGAN2 and GenData-StyleSwin \wrt \texttt{Gender}.
%\textcolor{blue}{
Additionally, we evaluate a publically available pre-trained Latent Diffusion Model (LDM) \cite{rombach2022high} on FFHQ \cite{karrasStyleBasedGeneratorArchitecture2019}, where we acquire the GT $\pstar$ \wrt \texttt{Gender} with the same procedure as GenData.
%}
%}

%\textcolor{blue}{
Our results in Tab. \ref{tab:G_PE_CLIP} and \ref{tab:LDM_PE_CLIP} shows that the Baseline is able to achieve reasonable accuracy in estimating the GT $\bpstar$. This is because CLIP's accuracy is very high ($\alpha_0$=0.998) on the bias class ($\pstar_0$) for both StyleGAN2, StyleSwin and 
%\textcolor{blue}{
LDM
%} 
resulting in less  mis-classification from occurring. Regardless, CLEAM is still able to further improve on the already very accurate baseline, further reducing the error from $\emubase \geq 0.91\%$, on StyleGAN2, StyleSwin and 
%\textcolor{blue}{
LDM
%} 
to $\emucleam \leq 0.47\%$.
A similar trend can be observed in the IE, where it is able to bound the GT $\pstar_0$.
%}

%\textcolor{red}{[Chris: Updating \dots]}
\begin{table*}[!h]
    \centering
    %New overall Caption
    \caption{
    %\textcolor{blue}{
    Comparing the {\em point estimates} and {\em interval estimates} of Baseline \cite{choiFairGenerativeModeling2020a} our CLEAM in estimating $\bpstar$ of 
    StyleGAN2 \cite{karrasStyleBasedGeneratorArchitecture2019} and StyleSwin \cite{zhang2021styleswin} with the GenData datasets.
    %the GenData-StyleGAN2/StyleSwin datasets. 
    We utilize SA classifier CLIP to classify sensitive attribute \texttt{Gender}.
    The $\pstar_0$ value of each GAN \wrt SA is determined by manually hand-labeling the generated data.
    We repeat this for 5 experimental runs and report the mean error rate, per Eqn. 1 of the main manuscript. 
   }
\resizebox{\textwidth}{!}{
    \addtolength{\tabcolsep}{-4pt}
      \begin{tabular}
      {ccc  cc cccc     cc cc cc}
        \toprule
        & & & \multicolumn{6}{c}{\bf Point Estimate} & \multicolumn{6}{c}{\bf Interval Estimate}
        \\
        \cmidrule(lr){4-9}\cmidrule(lr){10-15}
          %\textbf{Classifier} 
          &
          \textbf{  $\balpha=\{\alpha_0,\alpha_1\}$ } &
          Avg. $\balpha$ &
          %\textbf{  $\balpha=\{\alpha_0,\alpha_1\}$ } &
          \multicolumn{2}{c}{\textbf{Baseline}}
          %\cite{choiFairGenerativeModeling2020a}}
          & 
          \multicolumn{2}{c}{\textbf{Diversity}}
          %\cite{keswaniAuditingDiversityUsing2021}} 
          & \multicolumn{2}{c}{\textbf{CLEAM (Ours)}}
          %Interval Estimates
           & \multicolumn{2}{c}{\textbf{Baseline}}
           %\cite{choiFairGenerativeModeling2020a}}
           & 
           \multicolumn{2}{c}{\textbf{Diversity}}
           %\cite{keswaniAuditingDiversityUsing2021}} 
           & \multicolumn{2}{c}{\textbf{CLEAM (Ours)}}
        \\
        %\cmidrule(lr){1-1}
        \cmidrule(lr){2-2}\cmidrule(lr){3-3}\cmidrule(lr){4-5}\cmidrule(lr){6-7} \cmidrule(lr){8-9}
        \cmidrule(lr){10-11}\cmidrule(lr){12-13} \cmidrule(lr){14-15}
         %Baseline
        & & &
        $\mu_{\texttt{Base}}$ & $e_{\mu}(\downarrow)$ 
         %Diversity
         &$\mu_{\texttt{Div}}$ & $e_\mu(\downarrow)$
         %CLEAM
        &$\mu_{\texttt{CLEAM}}$ & $e_\mu(\downarrow)$
        &$\rho_{\texttt{Base}}$ & $e_\rho(\downarrow)$ 
        &$\rho_{\texttt{Div}}$ & $e_\rho(\downarrow)$ 
        &$\rho_{\texttt{CLEAM}}$ & $e_\rho(\downarrow)$\\
        \midrule
        \multicolumn{15}{c}{\cellcolor{lightgray!20} \bf (A) StyleGAN2}\\
        \midrule
        \multicolumn{15}{c}{\texttt{Gender} with GT class probability {\bfseries\boldmath $p^*_0$=0.642} }\\
        \midrule
        CLIP & \{0.998, 0.975\} & 0.987 &
        %Point Estimate----------------------------------------------------------
          0.653 &   1.71\% & 
        %Diversity
          --- & 
          --- & 
        %CLEAM
          0.645 &   \textbf{0.47\%} & 
        %Interval Estimate----------------------------------------------------------
        [0.649, 0.657] & 2.34\% & 
        %Diversity
        --- & --- & 
        %CLEAM
       [0.641, 0.649] & \textbf{1.09\%}
        \\  
        \midrule
        \multicolumn{15}{c}{\cellcolor{lightgray!20} \bf (B) StyleSwin}\\
        \midrule
        \multicolumn{15}{c}{\texttt{Gender} with GT class probability {\bfseries\boldmath $p^*_0$=0.656}}\\
        \midrule
        CLIP & \{0.998, 0.975\} & 0.987 & 
        %Point Estimate----------------------------------------------------------
          0.666 &   0.91\% & 
        %Diversity
          --- &   --- &  
        %CLEAM
          0.658 &   \textbf{0.30\%} 
        %Interval Estimate----------------------------------------------------------
        & [0.663,0.669] & 1.98\% & 
        %Diversity
        --- & --- & 
        %CLEAM
        [0.655,0.662] & \textbf{0.91\%}
        \\  %\hdashline
        \midrule
    \end{tabular}
    }
    \label{tab:G_PE_CLIP}
    \addtolength{\tabcolsep}{4pt}
    %\vspace{-4mm}
\end{table*}

\begin{table*}[!h]
    \centering
    %New overall Caption
    \caption{
    %\textcolor{blue}{
    Comparing the {\em point estimates} and {\em interval estimates} of Baseline \cite{choiFairGenerativeModeling2020a} our CLEAM in estimating $\bpstar$ of 
    a pretrained Latent Diffusion Model\cite{rombach2022high} on the FFHQ dataset.
    We repeat this for 5 experimental runs and report the mean error rate, per Eqn. 1 of the main manuscript. 
   % }
   }
\resizebox{\textwidth}{!}{
    \addtolength{\tabcolsep}{-4pt}
      \begin{tabular}
      {ccc  cc cccc     cc cc cc}
        \toprule
        & & & \multicolumn{6}{c}{\bf Point Estimate} & \multicolumn{6}{c}{\bf Interval Estimate}
        \\
        \cmidrule(lr){4-9}\cmidrule(lr){10-15}
          %\textbf{Classifier} 
          &
          \textbf{  $\balpha=\{\alpha_0,\alpha_1\}$ } &
          Avg. $\balpha$ &
          %\textbf{  $\balpha=\{\alpha_0,\alpha_1\}$ } &
          \multicolumn{2}{c}{\textbf{Baseline}}
          %\cite{choiFairGenerativeModeling2020a}}
          & 
          \multicolumn{2}{c}{\textbf{Diversity}}
          %\cite{keswaniAuditingDiversityUsing2021}} 
          & \multicolumn{2}{c}{\textbf{CLEAM (Ours)}}
          %Interval Estimates
           & \multicolumn{2}{c}{\textbf{Baseline}}
           %\cite{choiFairGenerativeModeling2020a}}
           & 
           \multicolumn{2}{c}{\textbf{Diversity}}
           %\cite{keswaniAuditingDiversityUsing2021}} 
           & \multicolumn{2}{c}{\textbf{CLEAM (Ours)}}
        \\
        %\cmidrule(lr){1-1}
        \cmidrule(lr){2-2}\cmidrule(lr){3-3}\cmidrule(lr){4-5}\cmidrule(lr){6-7} \cmidrule(lr){8-9}
        \cmidrule(lr){10-11}\cmidrule(lr){12-13} \cmidrule(lr){14-15}
         %Baseline
        & & &
        $\mu_{\texttt{Base}}$ & $e_{\mu}(\downarrow)$ 
         %Diversity
         &$\mu_{\texttt{Div}}$ & $e_\mu(\downarrow)$
         %CLEAM
        &$\mu_{\texttt{CLEAM}}$ & $e_\mu(\downarrow)$
        &$\rho_{\texttt{Base}}$ & $e_\rho(\downarrow)$ 
        &$\rho_{\texttt{Div}}$ & $e_\rho(\downarrow)$ 
        &$\rho_{\texttt{CLEAM}}$ & $e_\rho(\downarrow)$\\
        \midrule
        \multicolumn{15}{c}{\cellcolor{lightgray!20} \bf Latent Diffusion Model}\\
        \midrule
        \multicolumn{15}{c}{\texttt{Gender} with GT class probability {\bfseries\boldmath $p^*_0$=0.570} }\\
        \midrule
        CLIP & \{0.998, 0.975\} & 0.987 &
        %Point Estimate----------------------------------------------------------
          0.585 &   2.63\% & 
        %Diversity
          --- & 
          --- & 
        %CLEAM
          0.571 &   \textbf{0.18\%} & 
        %Interval Estimate----------------------------------------------------------
        [0.578, 0.593] & 4.04\% & 
        %Diversity
        --- & --- & 
        %CLEAM
       [0.564, 0.579] & \textbf{1.58\%}
    \end{tabular}
    }
    \label{tab:LDM_PE_CLIP}
    \addtolength{\tabcolsep}{4pt}
    %\vspace{-4mm}
\end{table*}

%\vfill

\FloatBarrier
\clearpage
\subsection{Comparing Classifiers Accuracy on Validation Dataset vs Generated Dataset}
\label{subsec:accOfGenaVsReala}

%\textcolor{red}{[Chris: Compare $\balpha$ of our CelebA-HQ validation vs SDM] [{\bf Addressed}]}

%\textcolor{red}{To update discussion with actual CLEAM results}
In our proposed CLEAM, we use 
$\balpha$ pre-measured on the validation dataset, denoted by $\balpha_{val}$. In this section, we show that 
$\balpha_{val}$ is a good approximate of the $\balpha$ when measured on the generated data, denoted by $\balpha_{gen}$.
% Note that $\balpha_{gen}$ is unknown in practice, therefore $\balpha_{val}$ is used instead. 
% As an accurate $\balpha$ is a crucial component to CLEAM, it is important to understand if our pre-evaluated $\balpha$ on the validation dataset, denoted by $\balpha_{val}$, is a good approximate of the $\balpha$ when measured on the generated data, denoted by $\balpha_{gen}$.
% Note that this is important as generally,
Note that
$\balpha_{gen}$ is not available in practice, therefore  $\balpha_{val}$ is used as approximation during CLEAM measurement. We further remark that our purpose of this experiment is only done to validate $\balpha_{val}$ as a good approximation for $\balpha_{gen}$ and is not necessary in the actual deployment of CLEAM.

%\textcolor{blue}{
{\bf Comparing $\balpha_{val}$ vs $\balpha_{gen}$ on GANs.}
%}
To validate that, we utilize our newly introduced generated dataset, with known labels
and measure the $\balpha_{gen}$ for both \texttt{Gender} and \texttt{Blackhair} on StyleGAN2 and StyleSwin and compared it against the $\balpha_{val}$.
The results in Tab. \ref{tab:evaluatingAlpha} show that $\balpha_{val}$ is a good approximation of the $\balpha_{gen}$ of the generated dataset.
%validation-----------------------------

In addition, in Tab. \ref{tab:G_PEValvsGenAlpha}, we further demonstrate the difference in effect when utilizing $\balpha_{gen}$ as opposed to $\balpha_{val}$ with CLEAM for sensitive attribute \texttt{Gender} on StyleGAN2 from the %\textcolor{blue}{
GenData
%} 
dataset. Overall, we observed only marginal improvements, for most cases, when utilizing the $\balpha_{gen}$. Additionally, as the improvements by CLEAM were still very significant when utilizing the $\balpha_{val}$, and as the labels for the generated dataset are not readily available to evaluate $\balpha_{gen}$, we found the $\balpha_{val}$ to be a good approximation for $\balpha_{gen}$ for fairness measurement.    

%\textcolor{blue}{
{\bf Comparing $\balpha_{val}$ vs $\balpha_{gen}$ on SDM.} 
Similarly when evaluating the SDM with CLEAM, we also utilize $\balpha_{val}$ in-place of $\balpha_{gen}$. However, as a validation dataset is not readily available for SDM, we explored the use of a poxy validation dataset whose domain is a close representation as our applications. More specifically, 
%Similarly to attain the approximated $\balpha_{val}$, as discussed in the main manuscript when a validation dataset isn't available for SDMs, 
we utilize CelebA-HQ as our proxy validation dataset (with known labels \wrt \texttt{Gender}) to attan $\balpha_{val}$. Then similarly, we compare $\balpha_{val}$ to $\balpha_{gen}$ from our labelled GenData-SDM  (per prompt).
%Then to validate this approximation, we pass a uniform distribution \wrt \texttt{Gender} from our labelled GenData-SDM (per prompt) to our CLIP classifier to attain $\balpha_{gen}$, which we compare against our approximated $\balpha_{val}$.
%we randomly sampled 500 images from the diffusion model \wrt the respective prompts and hand-labelled the sampled with the same procedure discussed in Sec. \ref{subsec:newDataset}. 
%Then with these labelled samples, we attain the GT $\pstar$, and additionally evaluate true $\balpha$ by passing a uniform sample of each class to the CLIP classifier.
As shown in Tab. \ref{tab:approxalpha} our approximated $\balpha_{val}$ (measured on CelebA-HQ), although not perfect, is a close approximation of $\balpha_{gen}$, thereby making it appropriate to be utilized with CLEAM.

%}
\begin{table}[ht!]
    \centering
    \caption{Comparing $\balpha_{val}$ ( $\balpha$ measured during the classifier's validation stage using real data), against $\balpha_{gen}$ ( $\balpha$ measured on the generated dataset). Notice that the $\balpha_{val}$ measured during the validation dataset is a close approximation of the generated dataset's $\balpha_{gen}$.}
    \resizebox{\textwidth}{!}{
    \begin{tabular}{c cccc  cccc}
    \toprule
                            & \multicolumn{4}{c}{StyleGAN2} & \multicolumn{4}{c}{StyleSwim}  \\
                            \cmidrule(lr){2-5} \cmidrule(lr){6-9} 
                            & ResNet18 & ResNet34 & MobileNetv2 & VGG16 & ResNet18 & ResNet34 & MobileNetv2 & VGG16 \\
    \midrule
                            & \multicolumn{8}{c}{\texttt{Gender}} \\
    \midrule
    Validated $\balpha$      & [0.947,0.983]  & [0.932,0.976] & [0.938, 0.975] & [0.801,0.919]
                            & [0.947,0.983]  & [0.932,0.976] & [0.958, 0.975] & [0.801,0.919]
    \\
    $\balpha_{gen}$           & [0.940,0.984]  & [0.928,0.982] & [0.948, 0.985] & [0.815,0.922]
                            & [0.957,0.966]  & [0.944,0.981] & [0.956, 0.977] & [0.804,0.924] \\
    \midrule
                            & \multicolumn{8}{c}{\texttt{Blackhair}} \\
    \midrule
    Validated $\balpha$      & [0.869,0.884] & [0.834,0.919] & [0.839,0.880] & [0.850,0.836]
                            & [0.869,0.884] & [0.834,0.919] & [0.839,0.880] & [0.850,0.836]
    \\
    $\balpha_{gen}$           & [0.870,0,885] & [0.830,0.914] & [0.845,0.886] & [0.837,0.824]
                            & [0.874,0.892] & [0.824,0.930] & [0.837,0.891] & [0.847,0.821]\\
                            \bottomrule
    \end{tabular}
    }
    \label{tab:evaluatingAlpha}
\end{table}
\begin{table}[!h]
    \centering
    %New overall Caption
    \caption{
    Comparing the {\em point estimates} and {\em interval estimates} of Baseline and our proposed CLEAM measurement framework in estimating $\bpstar$ of the GenData datasets sampled from (A) StyleGAN2 \cite{karrasStyleBasedGeneratorArchitecture2019}. 
    The $\pstar_0$ value for each GAN with a certain attribute is determined by manually hand-labeling the generated data.
    We then utilize a Resnet-18 to classify attributes \texttt{Gender} to obtain $\bphat$. Then with different accuracy $\balpha$, measured from the validation split (denoted by $\balpha_{val}$) and GenData datasets (denoted by $\balpha_{gen}$), we apply CLEAM.
    Each $\bphat$ utilizes $n=400$ samples and is evaluated for a batch-size of $s=30$.
    We repeat this for 5 experimental runs and report the mean error rate, per Eqn. 1 from the main manuscript. 
   }
\resizebox{\textwidth}{!}{
    \addtolength{\tabcolsep}{-4pt}
      \begin{tabular}
      {c cc cc cc cc @{\hspace{0.7cm}} cc cc cc}
        \toprule
        & \multicolumn{6}{c}{\bf Point Estimate} & \multicolumn{6}{c}{\bf Interval Estimate}
        \\
        \cmidrule(lr){2-7}\cmidrule(lr){8-13}
          \textbf{Classifier} &
          \multicolumn{2}{c}{\textbf{Baseline}\cite{choiFairGenerativeModeling2020a}} & \multicolumn{2}{c}{\textbf{CLEAM (Ours) with $\balpha_{val}$}} & \multicolumn{2}{c}{\textbf{CLEAM (Ours) with $\balpha_{gen}$}}
          %Interval Estimates
            & \multicolumn{2}{c}{\textbf{Baseline}\cite{choiFairGenerativeModeling2020a}}& \multicolumn{2}{c}{\textbf{CLEAM (Ours) with $\balpha_{val}$}} & \multicolumn{2}{c}{\textbf{CLEAM (Ours) with $\balpha_{gen}$}}
        \\
        \cmidrule(lr){1-1} \cmidrule(lr){2-3} \cmidrule(lr){4-5} \cmidrule(lr){6-7} \cmidrule(lr){8-9} \cmidrule(lr){10-11} \cmidrule(lr){12-13}
         %Baseline
        & 
        $\mu_{\texttt{Base}}$ & $e_{\mu}(\downarrow)$
         %CLEAM
        &$\mu_{\texttt{CLEAM}}$ & $e_\mu(\downarrow)$
        &$\mu_{\texttt{CLEAM}}$ & $e_\mu(\downarrow)$
        &$\rho_{\texttt{Base}}$ & $e_\rho(\downarrow)$ 
        &$\rho_{\texttt{CLEAM}}$ & $e_\rho(\downarrow)$
        &$\rho_{\texttt{CLEAM}}$ & $e_\rho(\downarrow)$\\
        \midrule
        \multicolumn{13}{c}{\bf (A) StyleGAN2}\\
        \midrule
        \multicolumn{13}{c}{\texttt{Gender} with GT class probability {\bfseries\boldmath $p^*_0$=0.642} }\\
        \midrule
        R18 &
        %Point Estimate----------------------------------------------------------
        0.610 & 4.98\% & 
        %CLEAM
        0.638 & 0.62\% & 
        0.639 & \textbf{0.44\%} &
        %Interval Estimate----------------------------------------------------------
        [0.602, 0.618] & 6.23\% & 
        %CLEAM
       [0.629, 0.646] & \textbf{2.02\%} &
       [0.629, 0.648] & \textbf{2.02\%}
        \\  
        %\hdashline
        %Point Estimate----------------------------------------------------------
        R34 &
        0.596 & 7.17\% &
        %CLEAM
        0.634 & 1.25\% &
        0.635 & \textbf{1.06\%} &
        %Interval Estimate----------------------------------------------------------
        [0.589, 0.599] & 8.26\% &
        %CLEAM
        [0.628, 0.638] & {2.18\%} &
        [0.630, 0.640] & \textbf{1.87\%}
        \\ 
        %\hdashline
        %Point Estimate----------------------------------------------------------
        MN2 &
        0.607 & 5.45\% & 
        %CLEAM
        0.637  & \textbf{0.78}\% &
        0.636 & 0.86\% &
        %Interval Estimate----------------------------------------------------------
        [0.602, 0.612] & 6.23\% &
        %CLEAM
        [0.632, 0.643] & \textbf{1.56\%} &
        [0.630, 0.642] & 1.82\%
        \\ 
        %\hdashline
        %Point Estimate----------------------------------------------------------
        V16 & 
        0.532 & 17.13\% & 
        %CLEAM
        0.636  & 0.93\% &
        0.640 & \textbf{0.36}\% &
        %Interval Estimate----------------------------------------------------------
        [0.526, 0.538] & 18.06\% &
        %CLEAM
        [0.628, 0.644] & {2.18\%} &
        [0.632, 0.647] & \textbf{1.53\%}
        \\
        \midrule
        %Point Estimate----------------------------------------------------------
        \multicolumn{2}{c}{Avg Error}& 8.68\% 
        && {0.90}\%
        &&\textbf{0.68}\%
        %Interval Estimate----------------------------------------------------------
        && 9.70\%
        && {1.99}\%
        && \textbf{1.81}\%
        \\
        \bottomrule
    \end{tabular}
    }
    \label{tab:G_PEValvsGenAlpha}
    \addtolength{\tabcolsep}{4pt}
\end{table}
\begin{table}[h]
    \centering
    \caption{
    %\textcolor{blue}{
    Comparing the approximate $\balpha_{val}$ measured on CelebA-HQ versus CLIP's $\balpha_{gen}$ evaluated on a fair distribution of GenData-SDM for each prompt \wrt \texttt{Gender}.
    %\textcolor{red}{This has been flipped to [F/M], same as main manuscript}
   % }
    }
    \resizebox{0.95\textwidth}{!}{
    \begin{tabular}{c c c c c c c}
      \toprule
                  & Dataset & \multicolumn{5}{c}{Stable Diffusion Model}\\
                   \cmidrule(lr){2-2}\cmidrule(lr){3-7} 
                  &  CelebA-HQ & "Somebody" & "an individual" & "a human being" & "a person" & "one person" \\
                  \midrule
      $\balpha$   & [0.998,0.975] & [1.0,0.970] & [1.0,0.980] & [1.0,0.970] & [0.990, 0.970] & [1.0, 0.980]  \\
      \bottomrule
    \end{tabular}
    }
    \label{tab:approxalpha}
\end{table}

%\clearpage
\FloatBarrier
\subsection{Comparing CLEAM with Classifier Correction Techniques (BBSE/BBSC)}

\label{sec:detailsonBBSEBBSC}

%\textcolor{red}{\bf [Milad: title updated!]}

%\textcolor{red}{
%Details to include
%\begin{itemize}
%    \item More detailed procedure for implementing BBSE and BBSC 
%    \item Table on BBSC results and a more detailed discussion of the results
%\end{itemize}
%}

%\textcolor{blue}{
In this section, we compare CLEAM against a few classifier correction techniques.
We remark that CLEAM, unlike the classifier correction techniques, does not aim to improve the sensitive attribute classifier's accuracy but instead accounts for its errors during fairness measurement. However, given that classifier correction techniques may improve bias measurement, we found it useful to make a comparison.
Specifically, we look into Black-Box shift estimator/correction (BBSE/BBSC) \cite{liptonDetectingCorrectingLabel2018}, methods previously proposed to address classifier inaccuracies
due to label shift. 
We demonstrate that even with BBSE/BBSC, errors in bias measurement still remain significant. 
%}
%We attribute these errors to the method's strict assumption that the classifier's inaccuracies result solely from label shifts, ignoring other factors. }

%\textcolor{blue}{
\textbf{Setup.}
To determine the effectiveness of BBSE/BBSC in tackling the errors of fairness measurement in generative models we evaluate it on the same setup as per 
%experiment \ref{sec:evalrealG} (GenData)
Sec. 5.1 of the main manuscript on GenData-StyleGAN and GenData-StyleSwin 
with ResNet-18. Specifically, for BBSE we follow Lipton \etal \cite{liptonDetectingCorrectingLabel2018} and first evaluate the confusion matrix for the trained ResNet-18 based on the validation dataset. 
Then, utilizing the confusion matrix, we calculate the weight vector which accounts for label shift of the generated data.
With this weight vector, we now implement a variant of CLEAM utilizing Algo.1  (with the weighted vector in-place of $\balpha$) in the main manuscript to evaluate the PE and IE.
%in Eqn. 4 and 5 in the main manuscript, in place of $\balpha$.
Similarly, for BBSC, we calculate the weight vector. However, unlike BBSE, we now utilize the weighted vector and fine-tune the classifier on the generated samples \cite{liptonDetectingCorrectingLabel2018}.
%See supp. \ref{sec:detailsonBBSEBBSC} for more details.
%Our experiment compares the following methods on the same setups as experiment 5.1 of our manuscript:
%}

%\textcolor{blue}{
Our results in Tab. \ref{tab:BBSE} shows that BBSE helps to marginally reduce $e_\mu$ and $e_\rho$ for the PE and IE, when compared against the baseline. However, these results still remain poor when compared to our original CLEAM implementation. 
One reason for this difference may be that, unlike CLEAM which is agnostic to the cause of the error, BBSE specifically corrects for label shifts while neglecting other sources of error \eg task hardness. 
%CLEAM on the other hand is error agnostic and generally addresses the existing error in the classifier.
Meanwhile, our results in Tab. \ref{tab:bbsc} show that utilizing BBSC makes no improvement in the improving the baseline fairness measurements. 
We hypothesize that this is due to the strong assumption of invariant conditional input distribution $p(x|y)$ used in BBSC, which may not hold in our problem. 
%See Supp. \ref{sec:detailsonBBSEBBSC} for full results.
Overall we conclude that while classifier correction techniques may improve fairness measurements in some cases, they may not always generalize as they are often tailored to a specific problem. 
%} 

\begin{table}[h]
    \centering
    \caption{
   % \textcolor{blue}{
    Comparing BBSE and CLEAM in estimating $\bpstar$ on GenData-StyleGAN2 and GenData-StyleSwin \wrt \{\texttt{Gender},\texttt{BlackHair}\}. %using measurement frameworks 1) Baseline \cite{choiFairGenerativeModeling2020a} 2) BBSE \cite{liptonDetectingCorrectingLabel2018}  3) CLEAM. 
    Here, we utilize a ResNet-18 trained on the CelebA-HQ dataset. }
  %  }
    \resizebox{\textwidth}{!}{
    \begin{tabular}{cc cc cc cc cc cc c }
   \toprule
        %\multicolumn{7}{c}{\bf (A) Pseudo-Generator}\\
        %\midrule
        %& 
        \multicolumn{6}{c}{\bf Point Estimate} & \multicolumn{6}{c}{\bf Interval Estimate}\\
        \cmidrule(lr){1-6}\cmidrule(lr){7-12}
          %\textbf{GT} & 
          \multicolumn{2}{c}{\textbf{Baseline}
          }& \multicolumn{2}{c}{\textbf{BBSE}
          } & \multicolumn{2}{c}{\textbf{CLEAM (Ours)}}
          & \multicolumn{2}{c}{\textbf{Baseline}
          }& \multicolumn{2}{c}{\textbf{BBSE}
          } & \multicolumn{2}{c}{\textbf{CLEAM (Ours)}}
        \\
         \cmidrule(lr){1-2} \cmidrule(lr){3-4} \cmidrule(lr){5-6}
         \cmidrule(lr){7-8} \cmidrule(lr){9-10} \cmidrule(lr){11-12}
         %Baseline
        $\mu_{\texttt{Base}}$ & $e_\mu (\downarrow)$ 
         %Diversity
         &$\mu_{\texttt{BBSE}}$ & $e_\mu(\downarrow)$ 
         %CLEAM
        &$\mu_{\texttt{CLEAM}}$ & $e_\mu(\downarrow)$
         %Baseline
        &$\rho_{\texttt{Base}}$ & $e_\rho (\downarrow)$ 
         %Diversity
         &$\rho_{\texttt{BBSE}}$ & $e_\rho(\downarrow)$ 
         %CLEAM
        &$\rho_{\texttt{CLEAM}}$ & $e_\rho(\downarrow)$
        \\
        \midrule
        \multicolumn{12}{c}{\cellcolor{lightgray!20} \bf (A) StyleGAN2}\\
        \midrule
        \multicolumn{12}{c}{\texttt{Gender} with GT class probability {\bfseries\boldmath $p^*_0$=0.642} }\\
        \midrule
        0.610 & 4.98\% & 0.621 & 3.38\% & 0.638 & {\bf 0.62\%} &
        [0.602,0.618] & 6.23\% & [0.613,0.628] & 4.52\% & [0.629,0.646] & {\bf 2.02\%} \\
        \midrule
        \multicolumn{12}{c}{\texttt{BlackHair} with GT class probability
        {\bfseries\boldmath $p^*_0$=0.643} }\\
        \midrule
        0.599 & 6.48\% & 0.630 & 2.02\% & 0.641 & {\bf 0.31\%} &
        [0.591,0.607] & 8.08\% &	[0.622,0.638] & 3.27\% & [0.631,0.652] & {\bf 1.40\%}
        \\
        \midrule
        \multicolumn{12}{c}{\cellcolor{lightgray!20} \bf (B) StyleSwin}\\
        \midrule
       \multicolumn{12}{c}{\texttt{Gender} with GT class probability {\bfseries\boldmath $p^*_0$=0.656} }\\
       \midrule
        0.620 &	5.49\% & 0.628 & 4.27\% & 0.648 & {\bf 1.22\%} &
        [0.612,0.629] &	6.70\% & [0.620,0.634] & 5.49\% & [0.639,0.658] & {\bf 2.59\%} \\
        \midrule
       \multicolumn{12}{c}{\texttt{BlackHair} with GT class probability {\bfseries\boldmath $p^*_0$=0.668} }\\
       \midrule
       0.612 & 8.38\% & 0.640 & 4.20\% & 0.659 & {\bf 1.35\%} &
       [0.605,0.620] &	9.43\% & [0.633,0.647] & 5.24\% & [0.649,0.670] & {\bf 2.84\%}
       \\
       \bottomrule
    \end{tabular}
     }
    \label{tab:BBSE}
\end{table}

\begin{table}[h]
    \centering
    \caption{
   % \textcolor{blue}{
    Comparing fairness distribution with ResNet-18 trained with and without Black-Box Shift Correction (BBSC) on the GenData dataset. Here we utilize the prior work’s fairness measurement framework (Baseline) and our proposed CLEAM to evaluate the fairness distribution.
   % }
    }
    \resizebox{\textwidth}{!}{
    \begin{tabular}{c c c cc cc cc cc }
   \toprule
        %\multicolumn{7}{c}{\bf (A) Pseudo-Generator}\\
        %\midrule
        %& 
        & & & \multicolumn{4}{c}{\bf Point Estimate} & \multicolumn{4}{c}{\bf Interval Estimate}\\
        \cmidrule(lr){4-8}\cmidrule(lr){9-11}
          %\textbf{GT} & 
         Setup & $\balpha$ & Avg $\balpha$ & \multicolumn{2}{c}{\textbf{Baseline}
          }
          %& \multicolumn{2}{c}{\textbf{BBSE}} 
          & \multicolumn{2}{c}{\textbf{CLEAM (Ours)}}
          & \multicolumn{2}{c}{\textbf{Baseline}
          }
          %& \multicolumn{2}{c}{\textbf{BBSE}} 
          & \multicolumn{2}{c}{\textbf{CLEAM (Ours)}}
        \\
         \cmidrule(lr){1-1} \cmidrule(lr){2-2} \cmidrule(lr){3-3}
         \cmidrule(lr){4-5}
         \cmidrule(lr){6-7} \cmidrule(lr){8-9} \cmidrule(lr){10-11}
         & & &
         %Baseline
        $\mu_{\texttt{Base}}$ & $e_\mu (\downarrow)$ 
         %Diversity
         %&$\mu_{\texttt{BBSE}}$ & $e_\mu(\downarrow)$ 
         %CLEAM
        &$\mu_{\texttt{CLEAM}}$ & $e_\mu(\downarrow)$
         %Baseline
        &$\rho_{\texttt{Base}}$ & $e_\rho (\downarrow)$ 
         %Diversity
         %&$\rho_{\texttt{BBSE}}$ & $e_\rho(\downarrow)$ 
         %CLEAM
        &$\rho_{\texttt{CLEAM}}$ & $e_\rho(\downarrow)$
        \\
        \midrule
        \multicolumn{11}{c}{\cellcolor{lightgray!20} \bf (A) StyleGAN2}\\
        \midrule
        \multicolumn{11}{c}{\texttt{Gender} with GT class probability {\bfseries\boldmath $p^*_0$=0.642} }\\
        \midrule
               Original Classifier & \{0.947,0.983\} & 0.97 & 0.610 & 4.98\% & 0.638 & 0.62\% &[0.602,0.618] &	6.23\% & [0.629,0.646] & 2.02\% \\
        Adapted Classifier w. BSSC & \{0.932,0.980\} & 0.96 & 0.609 & 5.28\% & 0.645 & 0.46\% & [0.601,0.616] & 6.53\% & [0.635,0.655] & 2.02\% \\
        \midrule
        \multicolumn{11}{c}{\texttt{BlackHair} with GT class probability
        {\bfseries\boldmath $p^*_0$=0.643} }\\
        \midrule
               Original Classifier & \{0.869,0.885\} & 0.88 & 0.599 & 6.48\% & 0.641 & 0.31\% & [0.591,0.607] &	8.08\% & [0.631,0.652] & 1.40\% \\
        Adapted Classifier w. BSSC & \{0.854,0.875\} & 0.86 & 0.588 & 8.55\% & 0.635 & 1.24\% & [0.581,0.596] & 9.64\% & [0.627,0.643] & 2.49\%
\\
        \midrule
        \multicolumn{11}{c}{\cellcolor{lightgray!20} \bf (B) StyleSwin}\\
        \midrule
       \multicolumn{11}{c}{\texttt{Gender} with GT class probability {\bfseries\boldmath $p^*_0$=0.656} }\\
       \midrule
               Original Classifier & \{0.947,0.983\} & 0.97 & 0.620 & 5.49\% & 0.648 & 1.22\% & [0.612,0.629] & 6.70\% & [0.639,0.658] & 2.59\% \\
        Adapted Classifier w. BSSC & \{0.932,0.980\} & 0.96 & 0.617 & 5.94\% & 0.655 & 0.15\% & [0.610,0.614] & 7.01\% & [0.649,0.661] & 1.06\%\\
        \midrule
       \multicolumn{11}{c}{\texttt{BlackHair} with GT class probability {\bfseries\boldmath $p^*_0$=0.668} }\\
       \midrule
               Original Classifier & \{0.869,0.885\} & 0.88 & 0.612 & 8.38\% & 0.659 & 1.35\% & [0.605,0.620] & 9.43\% & [0.649,0.670] & 2.84\% \\
        Adapted Classifier w. BSSC & \{0.854,0.875\} & 0.86 & 0.608 & 8.98\% & 0.663 & 0.75\% & [0.600,0.616] & 10.18\% & [0.655,0.671] & 1.95\% \\
       \bottomrule
    \end{tabular}
     }
    \label{tab:bbsc}
\end{table}
\clearpage
\FloatBarrier
\subsection{
%Interval Estimation Results of CLEAM for Bias Mitigation Algorithms
Applying CLEAM to Re-evaluate Bias Mitigation Algorithms
}

\label{subsec:IntervalEstimatesResults}
%\textcolor{red}{Chris: To do; the cyan parts are to be properly integrated with the current text. [{\bf Addressed}]}

%\textcolor{blue}{
{\em Importance-weighting} \cite{choiFairGenerativeModeling2020a} is a simple and effective method for bias mitigation. However, its performance in fairness improvement is measured by the Baseline, which could be erroneous. In this section, we re-evaluate the performance of importance-weighting with CLEAM, which has shown better accuracy in fairness estimation.
%}

%\textcolor{blue}{
Following Choi \etal \cite{choiFairGenerativeModeling2020a}, we utilize the original source code to train two BIGGANs \cite{brockLargeScaleGAN2019} on CelebA \cite{liuDeepLearningFace2015}: for the first GAN, without applying any bias mitigation (Unweighted), while in the second, we apply importance re-weighting (Weighted).
We do this for the originally proposed sensitive attribute \texttt{Gender}, and extend the experiment to \texttt{BlackHair}.
For a fair comparison, we follow \cite{choiFairGenerativeModeling2020a} and similarly use a ResNet-18 with a reasonably high average accuracy of  $88\%$ and $97\%$ for sensitive attributes \texttt{BlackHair} and \texttt{Gender}. 
%See Supp. \ref{subsec:trainingParam} for more details on training. 
Our results in Tab. \ref{tab:biasMitigation_PE} show that Baseline measures a $\mu_\texttt{Base}$ of $0.727$ and $0.680$ for Unweighted and Weighted, with SA $\texttt{Gender}$ (similar to reported results in \cite{choiFairGenerativeModeling2020a}). 
%Note that $\mu_\texttt{Base}(\pstar_0)=0.5$ is considered as a fair generator.
Meanwhile, CLEAM's results show that $\mu_\texttt{CLEAM}>\mu_\texttt{Base}$, implying that previous work could have underestimated the bias of the GANs. This could lead to an erroneous evaluation of a bias mitigation technique, or comparison across different bias mitigation techniques.
%See 
%Supp.\ref{subsec:trainingParam} for training details and 
%\ref{subsec:IntervalEstimatesResults} 
%for IE results.
%}
Then, when analyzing bias mitigation techniques using IE of CLEAM (as per Tab. \ref{tab:biasMitigation_IE}), since the IE of unweighted and weighted GANs do not overlap, we are provided with some statistical guarantees that the bias mitigation techniques, importance-weighting, is indeed effective.

\begin{table}[ht]
\centering
    \caption{
    %\textcolor{blue}{
    Re-evaluating the {\em point estimates} of previously proposed bias mitigation method, importance-weighting (imp-weighting) \cite{choiFairGenerativeModeling2020a} with CLEAM. 
    We first evaluate the bias of a BIGGAN \cite{brockLargeScaleGAN2019} with and without imp-weighting \ie unweighted and weighted, with the Baseline. 
    Then, we apply CLEAM to obtain a more accurate measurement.
    %and compare it against the Baseline. 
    We do this for both \texttt{Gender} and \texttt{BlackHair} sensitive attributes.
   % }
    }
    \resizebox{0.50\linewidth}{!}{
        \begin{tabular}{c  @{\hspace{0.5em}} c @{\hspace{1em}} c @{\hspace{1em}} c }
            \toprule
              \textbf{Setup} & \textbf{Baseline}& \textbf{Diversity} & \textbf{CLEAM (Ours)}
            \\
             \cmidrule(lr){1-1}\cmidrule(lr){2-2}\cmidrule(lr){3-3} \cmidrule(lr){4-4}
             %Baseline
            &$\mu_{\texttt{Base}}$
             %Diversity
            &$\mu_{\texttt{Div}}$
             %CLEAM
            &$\mu_{\texttt{CLEAM}}$ \\
            %\toprule
            %\multicolumn{4}{c}{\bf (B) Importance Re-weighting \cite{choiFairGenerativeModeling2020a} GAN}\\
            \toprule
            \multicolumn{4}{c}{$\balpha$=[0.976,0.979], \texttt{Gender}}\\
            \midrule
            Unweighted & 0.727 & 0.711 &  0.738 \\
            Weighted & 0.680 & 0.671   & 0.690 \\
            \midrule
            \multicolumn{4}{c}{$\balpha$=[0.881,0.887], \texttt{BlackHair}}\\
             \midrule
            Unweighted & 0.729   & 0.716 & 0.803 \\
            Weighted & 0.716  & 0.706 & 0.785 \\
            \bottomrule
        \end{tabular}
    }
    \label{tab:biasMitigation_PE}
\end{table}

%This section contains the interval estimates (IE) results for the experiments in Sec. 5.2. As discussed in the main paper, the point estimate can still contain some approximation error. As a result, we suggest that in addition to the PE, the IE is better to be reported. 
%Our results, in the main paper (Sec 5.1 and 5.3), show that in our experiments where the GT $\pstar$ is known CLEAM's IE, in most cases, demonstrates being able to to bound the $\pstar$.
%Then, when analyzing bias mitigation techniques using IE of CLEAM (as per Tab. \ref{tab:biasMitigation_IE}), since the IE of unweighted and weighted GANs do not overlap, we are provided with some statistical guarantees that the bias mitigation techniques, importance weighting, is indeed effective.
\FloatBarrier

\begin{table}[!h]
    \centering
    \caption{Re-evaluating the {\em interval estimates} of previously proposed bias mitigation method, importance-weighting (imp-weighting) \cite{choiFairGenerativeModeling2020a} with 
    %Baseline \cite{choiFairGenerativeModeling2020a}, Diversity \cite{keswaniAuditingDiversityUsing2021} and 
     CLEAM. 
    To do this, we first evaluate the bias of a BIGGAN \cite{brockLargeScaleGAN2019} with and without implementing imp-weighting \ie unweighted and weighted, with the Baseline. 
    Then, we apply CLEAM to obtain more accurate measurements, which we use to compare against the Baseline. We do this for both \texttt{Gender} and \texttt{BlackHair} sensitive attributes.}
\resizebox{8cm}{!}{
      \begin{tabular}{c  c @{\hspace{0.5em}} c @{\hspace{1em}} c @{\hspace{0.5em}}}
        \toprule
       % \multicolumn{7}{c}{\bf (A) Pseudo-Generator}\\
       % \midrule
        \textbf{Setup} & \textbf{Baseline
        %\cite{choiFairGenerativeModeling2020a}
        }& \textbf{Diversity}
        %\cite{keswaniAuditingDiversityUsing2021}
         & \textbf{CLEAM(Ours)}
        \\
         \cmidrule(lr){1-1}\cmidrule(lr){2-2}\cmidrule(lr){3-3} \cmidrule(lr){4-4}
        %Baseline
        % &P.Est($p^*_0$) & Error &I.Est.($p^*_0$) & Max Error
        &$\rho_{\texttt{Base}}$ 
         %Diversity
         %&P.Est.($p^*_0$) & Error &I.Est.($p^*_0$) & Max Error
         &$\rho_{\texttt{Div}}$ 
         %CLEAM
        %& P.Est.($p^*_0$) & Error & I.Est.($p^*_0$) & Max Error
        &$\rho_{\texttt{CLEAM}}$\\
        %\toprule
        %\multicolumn{7}{c}{\bf (B)   Importance Re-weighting \cite{choiFairGenerativeModeling2020a} GAN}\\
        \midrule
        \multicolumn{4}{c}{$\balpha$=[0.976,0.979], \texttt{Gender}}\\
        \midrule
        Unweighted & [$0.721  ,0.732 $]  & [$0.697,0.722$]  & [$0.733 ,0.744 $] 
        \\
        Weighted & [$0.674 ,0.685 $]  & [$0.658,0.684$] & [$0.686 ,0.693 $] \\
        \midrule
        \multicolumn{4}{c}{$\balpha$=[0.881,0.887], \texttt{BlackHair}}\\
         \midrule
        Unweighted & [$0.725 ,0.733 $]  & [$0.704,0.729$]  & [$0.798 ,0.809 $] \\
        Weighted & [$0.710 ,0.722 $] & [$0.696,0.716$]  & [$0.778 ,0.792 $] \\
        \bottomrule
    \end{tabular}
    }
    \label{tab:biasMitigation_IE}
\end{table}

\clearpage
\FloatBarrier
% \section{Details on applying CLEAM to evaluate SOTA generators}
\section{Details on Applying CLIP as a SA Classifier}
\label{sec:applicationOfCLEAM}

%\textcolor{red}{\bf [Milad: Title Updated! Please update page 1 accordingly.]}

%\textcolor{red}{
%[Chris: To include:]
%\begin{enumerate}
%    \item Make sure the tables and figures are all F/M
%    \item Expand on CLIP being utilized as a classifier
%\end{enumerate}
%}

%\textcolor{red}{To Include discussion on how we utilized CLIP as a attribute Classifier[{\bf Addressed}]}

%\textcolor{blue}{
{\bf CLIP as a Sensitive Attribute classifier.}
To utilize CLIP as a sensitive attribute classifier (with the VIT-B/32 architecture), we follow the best practices suggested by Radford \etal \cite{radfordLearningTransferableVisual2021}. Here, we first input two different prompts, describing the respective classes, to the CLIP text-encoder, as seen in Tab. \ref{tab:CLIPPrompts}. 
As suggested by Radford \etal we utilize the prompt starting with "A photo of a" \ie a scene description, followed by our sensitive attribute's classes \eg female/male.
Next, we also encode the generated images with the CLIP image encoder. 
Finally, for each encoded generated image and the two encoded text-prompt, we take the cosine similarities followed by the $\argmax$. The $\argmax$ output provides us with the respective hard label of the generated image.
%}
%between the encoded generated image and the encoded text to assign the image its respective class. 

%\textcolor{blue}{
{\bf Generated Image pre-processing.} We remark that as the stable diffusion model produces a mixture of colored and greyscale images, for a fair comparison, we transform all images from RGB to greyscale before %being input in
feeding into
CLIP for classification.
%}

%Here, we utilise the prompts "a photo of a female" and "a photo of a male" to differentiate between the different classes for \texttt{Gender}. 
%Then as the stable diffusion model produces a mixture of colored and greyscale images, for a fair comparison, we transform all images from rgb to greyscale before being input in CLIP for classification. 
%See \ref{fig:moreIllusratioofSD} for some illustration.
%Finally, to assign the respective label, we follow radford \etal and assign the label of the encoded prompt with the smallest cosine similarity to the encoded image. 
\begin{table}[h]
    \centering
    \caption{Prompts for using CLIP \cite{radfordLearningTransferableVisual2021} for sensitive attribute classification .}
    \begin{tabular}{c c c}
    \toprule
       {\bf Sensitive Attribute}  & {\bf Class 0 prompt} & {\bf Class 1 prompt}  \\
       \cmidrule(lr){1-1} \cmidrule(lr){2-2} \cmidrule(lr){3-3}
       \texttt{Gender}  & "A photo of a female" & "A photo of a male"\\
       \texttt{Smiling} & "A photo of a face not smiling" & "A photo of a face smiling"\\
    \bottomrule
    \end{tabular}
    \label{tab:CLIPPrompts}
\end{table}

\clearpage
\section{Ablation Study: Details of Hyper-Parameter Settings and Selection}
\label{subsec:trainingParam}

%\textcolor{red}{To Include:
%\begin{enumerate}
%    \item ResNet inference accuracy [{\bf Addressed}] 
%    \item C02 Computation [{\bf Addressed}] 
%\end{enumerate}
%}

%\textcolor{red}{Chris: Need to update for new experiments and datasets. To include more specific details on the bias mitigation setup e.g. perc and bias}
\paragraph{\bfseries\boldmath Sensitive Attribute Classifier {$C_{u}$}.} 
In our experiments, we utilized a Resnet-18/34 \cite{heDeepResidualLearning2016}, MobileNetv2 \cite{sandlerMobileNetV2InvertedResiduals2018} and VGG-16.
The respective datasets \ie CelebA, \cite{liuDeepLearningFace2015} CelebA-HQ  \cite{CelebAMask-HQ} and AFHQ \cite{choiStarGANV2Diverse2020} datasets are then segmented into \{Train, Test, Validation\} with respect to the ratio \{80\%,10\%,10\%\}, where each segmentation of the dataset contains uniform distribution \wrt the queried sensitive attribute.
The classifiers are then trained with the training datasets and the $\balpha$ are evaluated with the validation dataset. 
Each classifier is trained with an Adam optimizer\cite{kingmaAdamMethodStochastic2017} with a learning rate=$1e^{-3}$, Batch size=$64$ and input dim=$64$x$64$ from the CelebA dataset \cite{liuDeepLearningFace2015}, dim=$64$x$64$ from the AFHQ dataset and dim=$128$x$128$ from the CelebA-HQ dataset \cite{CelebAMask-HQ}. 
%\textcolor{blue}{
Tab. \ref{tab:ClassifierAlphaCelebAforSOTA} details the $\balpha$ of the ResNet-18 utilized in Sec.6 of our main manuscript.
%}

\begin{table}[h]
    \centering
    \caption{
    %\textcolor{blue}{
    Accuracy of ResNet-18 trained and evaluated on CelebA-HQ.
    %}
    }
    \label{tab:ClassifierAlphaCelebAforSOTA}
    \begin{tabular}{c c}
        \toprule
       Sensitive Attribute  & Accuracy, $\balpha$  \\
       \midrule
       NoBeard  & [0.968,0.898] \\
       HeavyMakeup  & [0.925,0.883] \\
       Bald & [0.930,0.972] \\
       Chubby  & [0.838,933] \\
       Mustache & [0.925,0.896]\\
       Smiling  & [0.933, 0.877] \\
       Young  & [0.871, 0.857]\\
       BlackHair  & [0.869,0.885] \\
       Gender  & [0.947,0.983]\\
       \bottomrule
    \end{tabular}
\end{table}

\paragraph{\bfseries\boldmath Generator $G_\phi$ used in sec.\ref{subsec:IntervalEstimatesResults}.}
As mentioned in 
sec. \ref{subsec:IntervalEstimatesResults},
%the main paper, 
we utilized the setup in Choi \etal \cite{choiFairGenerativeModeling2020a}\footnote{https://github.com/ermongroup/fairgen} for the training of our imp-weighted and unweighted GANs. With this, we replicate their hyperparameter selection of $64$ x $64$ celebA \cite{liuDeepLearningFace2015} images with a learning rate=$2e^{-4}$, $\beta_1=0$, $\beta_2=0.99$ and four discriminator steps per generator step. We utilize a single RTX3090 for the training of our models.
%For MaGNET \cite{humayunMaGNETUniformSampling2021a}, we utilised the saved model weights for a StyleGANv2 \cite{karrasStyleBasedGeneratorArchitecture2019} as well as, the pre-computed latent vectors $z$ and $MaGNET(z)$ provided by \cite{humayunMaGNETUniformSampling2021a} to generate the respective $s*n$ samples.  

%In our experiments, where we generated $n=400$ samples with our pseudo-generator setup for $s$ batches
%, followed by applying 
%CLEAM to approximate $\pstar$, we found that the batch size $s=30$ to be suitable to approximate the known GT $\pstar$.
%Increasing $n$ did not lead to noticeable improvement in the error as discussed in the main paper. Similarly, as seen in Fig. \ref{fig:varyingS}, increasing $s$ did not result in significant improvements by both baseline and CLEAM. However, decreasing $s$ did observe some marginal degradation in performance \ie increase in $e_\mu$.

%\textcolor{blue}{
\textbf{Evaluating CLEAM with Different $\mathbf{n}$.} 
%We investigate the effectiveness of CLEAM with different batch sizes $n$. 
Utilizing the same setup 
in Sec. 5.1 of our main manuscript 
%in Sec. \ref{sec:evalrealG}
-- with the GenData-StyleGAN and GenData-StyleSwin datasets, we repeated the experiment with ResNet-18 and $n\in [100,600]$. Our results in Fig.\ref{fig:ResNet18Permuten} show that there is a marginal increase in error for both the Baseline and CLEAM as 
$n$ approaches 100,
%$n\rightarrow$100, 
while the converse occurs when 
$n$ approaches 600.
%$n\rightarrow$600.
However, given the diminishing improvements for $n>400$, we found $n=400$ to be ideal- a balance between computational cost and measurement accuracy.
%}

\begin{figure}[h]
    \centering
    \includegraphics[width=0.5\textwidth]{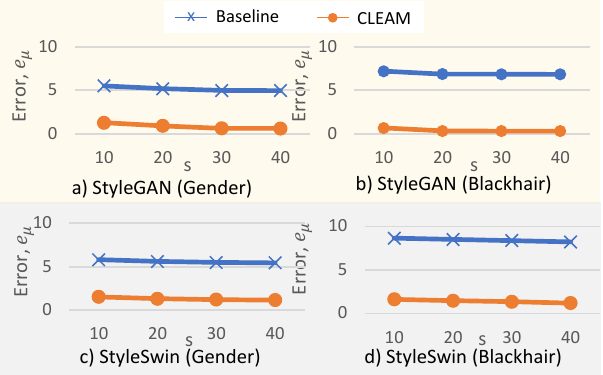}
    \caption{Comparing the point error $e_{\mu}$ for Baseline and CLEAM when evaluating the bias of the generated data with ResNet-18, for varying sample the number of batches, $s$.}
    \label{fig:varyingS}
\end{figure}

\paragraph{\bfseries\boldmath Batch Size $s$.}
In our experiments, we utilized $s$ batches of data each of which contains $n$ images to approximate $\bpstar$ with the Baseline and CLEAM methods. In the previous experiment, we found $n=400$ samples to be the ideal balance between computational time and minimizing fairness measurement error.
Here, we repeat the same hyper-parameter search, utilizing the real generator setup in Sec 5.1 of the main paper with ResNet-18, but instead varied the number of batches, $s$. Our results in Fig. \ref{fig:varyingS} found $s=30$ to be the optimal value when approximating $\bpstar$. Increasing $s$ did not result in significant improvements by both baseline and CLEAM. However, decreasing $s$ did observe some significant degradation in performance \ie increase in $e_\mu$.

\paragraph{\bfseries\boldmath Computational Time.}
In our main paper, we note that CLEAM is a lightweight correction to the existing baseline method, that requires no additional parameter to be computed during evaluation. To support this, we evaluated the computational time for the Baseline, Diversity, and our proposed CLEAM. Our results in Tab. \ref{tab:computationalTime} show that there is only a small difference in computational time ($\approx 0.1$s) between the Baseline and our proposed CLEAM. This difference is solely to facilitate the computation of Algo. 1. 
%\textcolor{blue}{
See Tab. \ref{tab:carbonemission} for discussion on carbon emission.
%}

\begin{table}[h]
    \centering
    \includegraphics[width=0.4\textwidth]{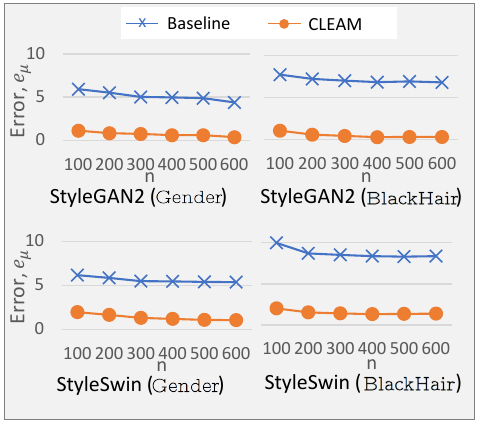}
    \captionof{figure}{
    %\textcolor{blue}{
    Comparing the point error $e_{\mu}$ for Baseline and CLEAM when evaluating the bias of GenData with ResNet-18, with varying sample size, $n$.
    %}
    }
    \label{fig:ResNet18Permuten}
\end{table}

%\textcolor{red}{[Chris: Include C02 computation \dots]}

%when evaluating the bias of real GANS in Sec. \ref{sec:evalrealG}, we found that each test point, on average of 5 test runs, took approximately 100 seconds to compute with CLEAM, 97 second for the Baseline and 600 seconds with diversity. This demonstrate that CLEAM does not contribute a significant amount overhead in computational time. 

\begin{table}[h]
    \centering
    \caption{Average computation time for estimating $\bpstar$ with $s$=30 and $n$=400 for Baseline \cite{choiFairGenerativeModeling2020a}, Diversity \cite{keswaniAuditingDiversityUsing2021} and our proposed CLEAM with a single RTX3090 for 5 consecutive runs. }
    \resizebox{0.8\linewidth}{!}{
    \begin{tabular}{c c c c}
    \toprule
    & Baseline \cite{choiFairGenerativeModeling2020a} & Diversity \cite{keswaniAuditingDiversityUsing2021} &
    CLEAM (Ours)\\
    \cmidrule(lr){2-2}\cmidrule(lr){3-3}\cmidrule(lr){4-4}
    %\midrule
    %\multicolumn{4}{c}{\bf CelebA, 64x64}\\
    %\midrule
     {\bf CelebA, 64x64} , $s$ & 99.9 & 600.4 & 100.0 \\
     {\bf AFHQ, 64x64} , $s$ & 99.8 & 601.2 & 99.9 \\
     %\midrule
     %\multicolumn{4}{c}{\bf CelebA-HQ, 128x128}\\
    %\midrule
     {\bf CelebA-HQ, 128x128}, $s$ & 135.9 & 820.4 & 136.0 \\
     \bottomrule
    \end{tabular}
    }
    \label{tab:computationalTime}
\end{table}
%Following this, we utilised same parameters; $s=30$ and $n=400$ samples generated from our trained GAN and approximate $\pstar$ with CLEAM.
%, as per Table. 1 in the main paper. 
%During which we computed the average time taken from the generation of $s=30$ and $n=400$ samples over 5 runs for each of the GANs, followed by the application of CLEAM, to be one minute and forty seconds. 

\begin{table}[h!]
    \centering

    \caption{\textbf{Estimated Computation time}. The carbon emission values are computed using \url{https://mlco2.github.io/impact}.}
    \resizebox{\columnwidth}{!}{%
    \begin{tabular}{c @{\hspace{1em}} c @{\hspace{1em}} c @{\hspace{1em}} c @{\hspace{1em}}} 
    \toprule
    Experiment & Hardware & GPU Hours & Carbon emitted (kg)\\
    \cmidrule(lr){1-1}\cmidrule(lr){2-2}\cmidrule(lr){3-3}\cmidrule(lr){4-4} 
    Training of SA Classifiers & RTX3090  & 2.0 & 0.39\\
    Comparing CLEAM on GANs, Main Paper Tab. 1 & RTX3090  & 4.8 & 0.94\\
    Comparing CLEAM on DGN, Main Paper Tab. 2 & RTX3090  & 0.3 & 0.1\\
    Inferring with CLEAM on DGN, Main Paper Fig. 3a & RTX3090  & 0.3 & 0.1\\
    Inferring CLEAM on GANs, Main Paper Fig. 3b & RTX3090  & 0.52 & 0.15\\
    Comparing CLEAM on PsuedoG, Supp Tab \ref{tab:fakeG_PE} & RTX3090  & 4.5 & 0.88\\
    Comparing CLEAM on PsuedoG Additional SA, Supp Tab \ref{Tab:YoungAndAttractivePEIE} & RTX3090  & 3 & 0.59\\
    Comparing CLEAM on PsuedoG Additional classifier, Supp Tab \ref{MobileNetPEIE} & RTX3090  & 4.5 & 0.88\\
    Comparing CLEAM on DGN with CLIP, Supp Tab. \ref{tab:G_PE_CLIP} & RTX3090  & 0.15 & 0.05\\
   Comparing CLEAM with BBSE/BBSC, Supp Tab. \ref{tab:BBSE} & RTX3090  & 0.25 & 0.07\\
    Applying CLEAM on Bias mitigation, Subb Tab \ref{tab:biasMitigation_PE} & RTX3090 & 0.88 & 0.17\\
   \midrule
   \multicolumn{2}{c}{\bf Total:} & 21.2 & 4.32 \\
    \bottomrule
    
    \end{tabular}%
    }
    \label{tab:carbonemission}
\end{table}

\FloatBarrier
\clearpage
\section{Related Work
}
\label{sec:relatedWork}
%\textcolor{red}{[Chris: Include disucssion on Bianchi \etal [{\bf Addressed}]]}

\textbf{Fairness in Generative Models.} Fairness in machine learning is mostly studied for discriminative learning, where usually the objective is to handle a classification task independent of a 
sensitive attribute 
in the input data, \eg making a hiring decision independent of the applicant \texttt{Gender}. 
However, the definition of fairness is quite different for generative learning, where it is considered as  equal representation/generation probability \wrt a sensitive attribute. 
%For example, a generative model that has an equal probability of producing \texttt{Male} and \texttt{Female} samples is fair \wrt \texttt{Gender}. 
Because of this difference, the conventional fairness metrics used for classification, like Equalised Odds, Equalised Opportunity \cite{hardtEqualityOpportunitySupervised2016} and Demographic Parity \cite{feldmanCertifyingRemovingDisparate2015}, cannot be applied to generative models.
Instead, the similarity between the probability distribution of the generated sample \wrt a sensitive attribute ($\bpstar$) and a target distribution $\bpbar$ (a uniform distribution) \cite{choiFairGenerativeModeling2020a} is utilized as fairness metric.
{ See sec. \ref{subsec:MoreOnFD} for details.}
%Since the attribute labels are not available for the generated images, usually $\pstar$ is estimated by an attribute classifier. In our work, we argue that due to inaccuracies in the attribute classifier, the estimated value for $\pstar$ can deviate from the unknown ground-truth value and degrade the fairness measurement accuracy.

\textbf{Existing Works on Fair Generative Models.}
Existing works focus on \textbf{\em bias mitigation} in generative models.
The importance reweighting algorithm is proposed by Choi \etal \cite{choiFairGenerativeModeling2020a} 
where a re-weighting algorithm favours a reference fair dataset \wrt the sensitive attribute in-place of a larger biased dataset.
%for the training of a Fair-GAN model.
%In this algorithm, a reference fair dataset \wrt the sensitive attribute is used during training, while still exposing the model to the larger biased dataset, but down-weighting it. This allows the generator to output high-quality samples while encouraging fairness \wrt the sensitive attribute.
Frankel \etal \cite{frankelFairGenerationPrior2020} introduces the concept of prior modification, where an additional smaller network is added to modify the prior of a GAN to achieve a fairer output. Tan \etal \cite{tanImprovingFairnessDeep2020} learns the latent input space \wrt the sensitive attribute, which they can later sample accordingly to achieve a fair output.
MaGNET \cite{humayunMaGNETUniformSampling2021a} demonstrates that enforcing uniformity in the latent feature space of a GAN, through a sampling process, improves fairness.
%\textcolor{blue}{
Um \etal \cite{umFairGenerativeModel2021} improves fairenss through the utilization of total variation distance which quantifies the unfairness between a small reference dataset and the generated samples.
Teo \etal \cite{teo2022fair} introduces fairTL++, which utilizes a small fair dataset to implement fairness adaptation via transfer learning.
%}
In all of these works, the focus is on improving fairness of the generative model (where the performance of the model is measured with a framework, in which the inaccuracies in the sensitive attribute classifier has been ignored). However, our proposed CLEAM method focuses on improving \textbf{\em fairness measurement}, by compensating for the inaccuracies in the sensitive attribute classifier through a statistical model. Therefore, it can be used to evaluate the bias mitigation algorithms more accurately.

\textbf{Equal Representation.}
Some literature also use a similar notion of equal representation (used in generative models) to address fairness. 
For example, fair clustering variation \cite{chierichettiFairClusteringFairlets2017} is proposed by enforcing the clusters to represent each attribute equally, and fair data summarization \cite{celisFairDiverseDPPBased2018} is used to mitigate the bias in creating a representative subset for a given dataset, while handling the trade-offs between fairness and diversity during sampling.
However, unlike our 
setup, %framework (Fig. \ref{fig:overall_framework}(a)) 
these works assume to have access to the attribute labels. 
Meanwhile, in data mining, a similar problem was recently studied.
Given a large dataset of unlabelled mined data, the objective is to evaluate the disparity of the dataset \wrt an attribute. To do this, an evaluation framework called diversity \cite{keswaniAuditingDiversityUsing2021} was introduced.
To measure this, a pre-trained classifier is used as a feature extractor. The unlabelled dataset is then compared against a  controlled reference dataset (with known labels) via a similarity algorithm. 
%(see Supp. \ref{subsec:calibrationTechniques} for more related work).

%\textcolor{blue}{
\textbf{Biases in Text-Image generation.} Some literature have attempted to look into the biases in text-to-image generators \cite{bianchiEasilyAccessibleTexttoImage2022}. 
Specifically, Bianchi \etal study existing biases in occupations-based prompts for popular text-to-image generators \eg stable diffusion models. 
They found the biases to exasperate existing occupation stereotypes, \eg nurses being over-represented as non-Caucasian females.
To measure these biases, \cite{bianchiEasilyAccessibleTexttoImage2022} has a simple approach utilizing a pre-trained feature extractor to assign the sensitive attribute labels to a small batch of generated images (100 samples).
%based on their similarity to a labelled reference dataset.
%To measure these biases the work compares the latent-embeddings, utilizing a pre-trained feature extractor, of a small batch of generated image (100 samples) against a reference dataset \wrt a sensitive attributes and assign the respective labels based on a similarly metric.
%}
%\textcolor{blue}{
%1) Differences in the measurements method to our proposed method
We remark that this approach is similar to Diversity, a method which we found to also demonstrate significant errors due to the lack of consideration for the classifier's error. 
%We remark that, in our work, we look into the accuracy of similar classification methods (Diversity). Here, we found that these bias measurement methods also demonstrate significant errors due to the lack of consideration for the classifiers error. 
%CLEAM is then propose to improve upon these erroneous measurements.
%2) Differences in the they studied application and ours
%Then in 
Furthermore, we emphasize the difference between our study and Bianchi \etal. Specifically, in
our application of CLEAM (Sec. 6 of the main manuscript),  we examine the impact of using prompts with indefinite pronouns/nouns that are synonymous to each another.
Our objective, unlike Bianchi \etal's work, is to investigate the influence of 
subtle changes in the prompts on bias, which is studied on a large dataset ($\approx 2k$ samples). Our results are the first to demonstrate that even subtle changes to the prompt (which are semantically unchanged), could result in drastically different biases.
%\textcolor{red}{[This is to be reviewed carefully, to ensure that the information doesn't "trap" us.]}
%various synonymous indefinite (neutral) prompts on bias. 
%
%Surprisingly our findings demonstrate that these different synonymous exhibit varying degrees of bias. This outcome contradicts our initial expectations, as we anticipated that the semantically similar prompts would produce similar biases, and since they were neutral, the biases would be negligible.
%}

%In the application section (Sec. 6 of the main manuscript), we examine the impact of using prompts with indefinite pronouns/nouns that are synonymous with each other. Our objective, in contrast to Bianchi et al.'s work, is to investigate the influence of various synonymous indefinite (neutral) prompts on bias. Surprisingly, our findings demonstrate that these different synonymous prompts exhibit varying degrees of bias. This outcome contradicts our initial expectations, as we anticipated that the semantically similar prompts would produce similar biases, and since they were neutral, the biases would be negligible.

%\subsection{CLEAM versus Other Calibration Techniques}
%\label{subsec:calibrationTechniques}
\textbf{Classifier Calibration.} The proposed CLEAM can be seen from a classifier calibration point of view as it refines the output of the classifier.
However, CLEAM should not be mistaken with conventional calibration algorithms, \eg temperature scaling \cite{guoCalibrationModernNeural2017}, Platt Scaling \cite{platt1999probabilistic} and Isotonic regression \cite{nyberg2021reliably}. Unlike these algorithms that concern themselves with the confidence of prediction, CLEAM focuses on sensitive attribute distribution, thereby making these algorithms ineffective.

\begin{figure}[h]
    \centering
    \includegraphics[scale=0.5]{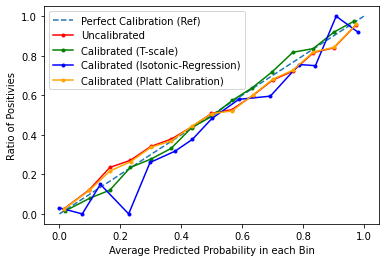}
    \caption{Calibration Curve on ResNet-18 for $\texttt{Attractive}$ sensitive attribute. We observe that the T-scaling is the most effective technique in improving soft label calibration and Isotonic regression the worst. However, this same trend does not follow in the hard label errors of Fig \ref{fig:Calibration_Error}.}
    \label{fig:Calibration_curve}
\end{figure}

More specifically, 
conventional classifier calibration methods usually work on soft labels (probabilities). Note that in our framework, the $argMax$ is applied to the output probabilities to determine the hard label. %In our application, regular classifier calibration does not affect the hard label that much. 
Therefore, in our application that deals with hard labels, regular classification techniques are less effective.
To investigate this, we conduct a few calibration experiment by applying some popular classifier calibration techniques; temperature scaling(T-scaling) \cite{guoCalibrationModernNeural2017}, Isotonic Regression\cite{nyberg2021reliably} and Platt Scaling\cite{platt1999probabilistic}  on a pre-trained ResNet-18\cite{heDeepResidualLearning2016} senstive attribute classifier. In Fig. \ref{fig:Calibration_curve}, we see that T-scaling is the most effective in correcting the calibration curve to the ideal Ref line.  
Note that, this Ref line indicates that the classifier is perfectly calibrated \wrt the soft labels.
%However, as shown in Fig. \ref{fig:Calibration_Error} T-scaling shows negligible improvement in normalized error, demonstraing that traditional calibration technique may not have a direct correlation to hard label calibration, which CLEAM aims to address.

\begin{figure}[!h]
    \centering
    \includegraphics[scale=0.4]{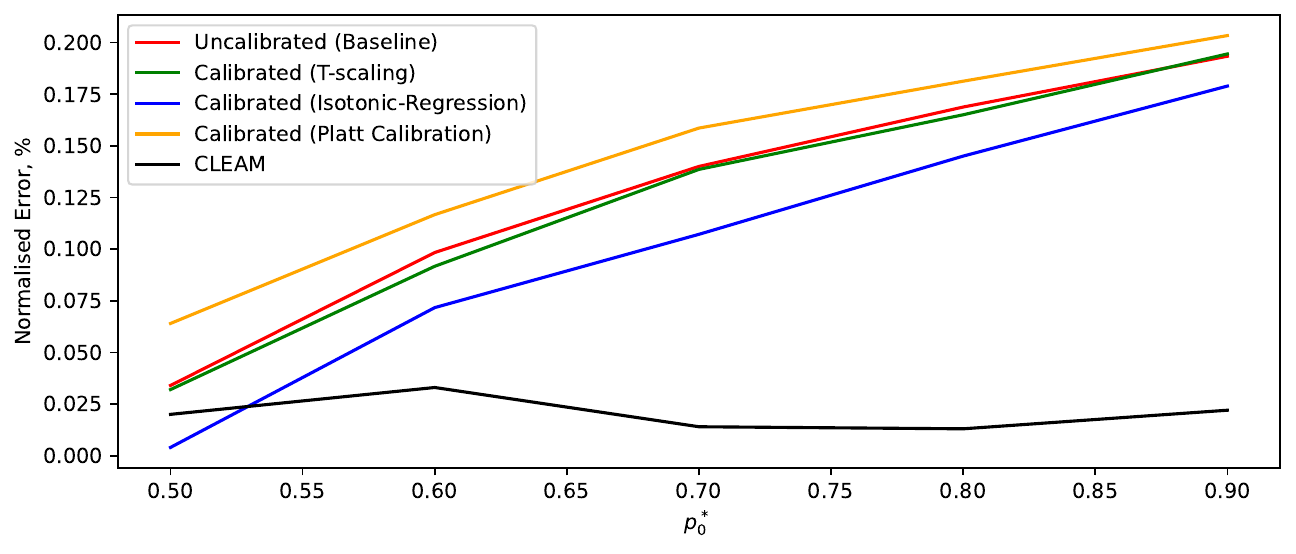}
    \caption{\textbf{Comparing Calibration Techniques}: Using the pseudo-generator, we compare CLEAM against well known calibration techniques, overall we observe that previous techniques are significantly less effective, achieving an average error of; T-Scaling: $12.4\%$, Isotonic Regression: $10.1\%$, Platt Calibration: $14.5\%$ and uncalibrated (baseline): $12.4\%$ against CLEAM: $2.0\%$ }
    \label{fig:Calibration_Error}
\end{figure}

Next, using the pseudo-generator from Sec. \ref{sec:psuedoG}, we utilised the calibrated sensitive attribute classifiers earlier and compare them against CLEAM (which was applied on an uncalibrated model).
%against the other well known calibration techniques.
%, namely T-scaling, Isotonic regression and Platt Scaling \cite{guoCalibrationModernNeural2017}. 
\textbf{In our results,} seen in Fig. \ref{fig:Calibration_Error}, we observe that these traditional calibration methods are less effective in correcting the sensitive attribute distribution error. 
In fact, methods like Platt scaling worsen the error, and T-scaling ---which is shown in \cite{guoCalibrationModernNeural2017} and our experiment to be one of the most effective traditional calibration methods--- does not change class predictions (hard labels), but merely perturb the soft labels. 
This demonstrates that traditional calibration technique are not direct correlation to hard label calibration, which CLEAM aims to address.

\label{sub:traditionalCalibration}
\clearpage
\section{Details of the GenData: A New Dataset of Labeled Generated Images}
\label{subsec:newDataset}
%\textcolor{red}{[Chris: Update details with SDM details and sample images]}

In this section, we provide more \hypertarget{datasetPorotcol}{information} on our new dataset, containing generated samples labeled based on sensitive attributes from StyleGAN2\footnote{https://github.com/NVlabs/stylegan2-ada-pytorch} \cite{karrasStyleBasedGeneratorArchitecture2019} and StyleSwin \footnote{https://github.com/microsoft/StyleSwin} \cite{zhang2021styleswin} trained on CelebA-HQ \cite{CelebAMask-HQ}, 
%\textcolor{blue}{
and a Stable Diffuson Model(SDM)\cite{rombach2021highresolution}.
%}
 More specifically, our dataset contains $\approx$9k randomly generated samples based on the original saved weights and codes of the respective GANs, 
 %\textcolor{blue}{
 and  $\approx$2k samples for four different prompts inputted in the SDM.
 %}. 
 %\textcolor{blue}{
 These samples are then hand labeled \wrt the sensitive attributes. More specifically, \texttt{Gender} and \texttt{BlackHair} for both the GANs and \texttt{Gender} for the SDM. 
 %}
 %These samples are then hand labeled \wrt the sensitive attribute \texttt{Gender} and \texttt{Blackhair}, which is perceived to be consistent with human perception. 
 Then with these labeled datasets, we can approximate the ground-truth sensitive attribute distribution, $\bpstar$, of the respective GANs.

\begin{figure}[h!]
\centering
     \begin{subfigure}[b]{0.49\textwidth}
         \centering
         \includegraphics[width=\textwidth]{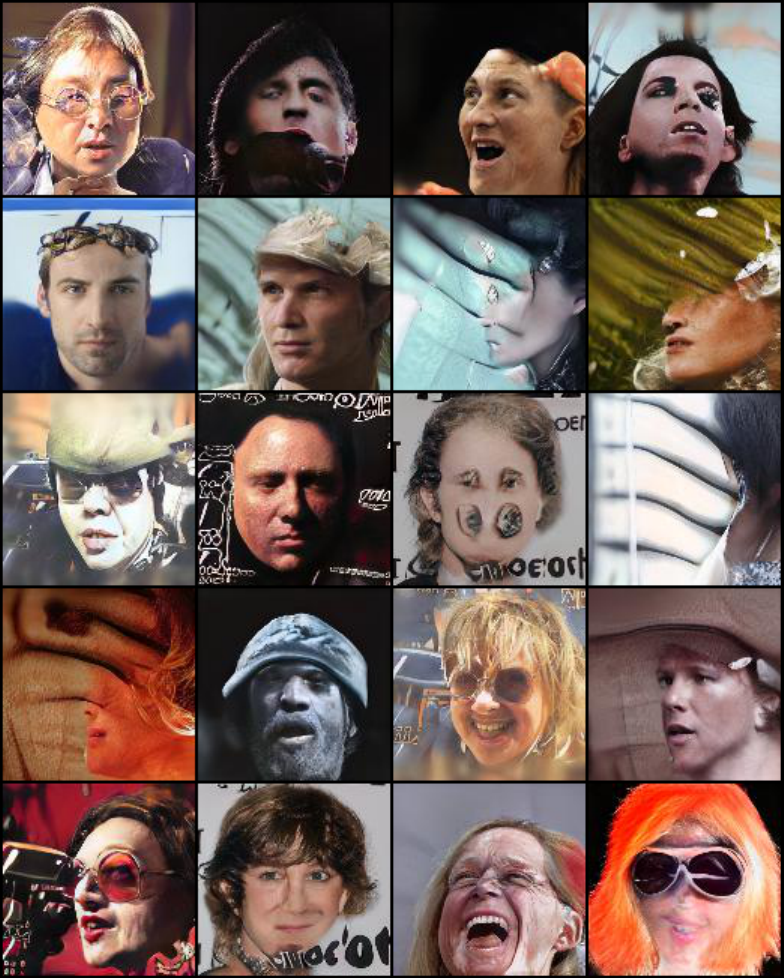}
         \caption{StyleGAN2}
         \label{fig:RejectedStyleGAN2}
     \end{subfigure}
     \hfill
     \begin{subfigure}[b]{0.49\textwidth}
         \centering
         \includegraphics[width=\textwidth]{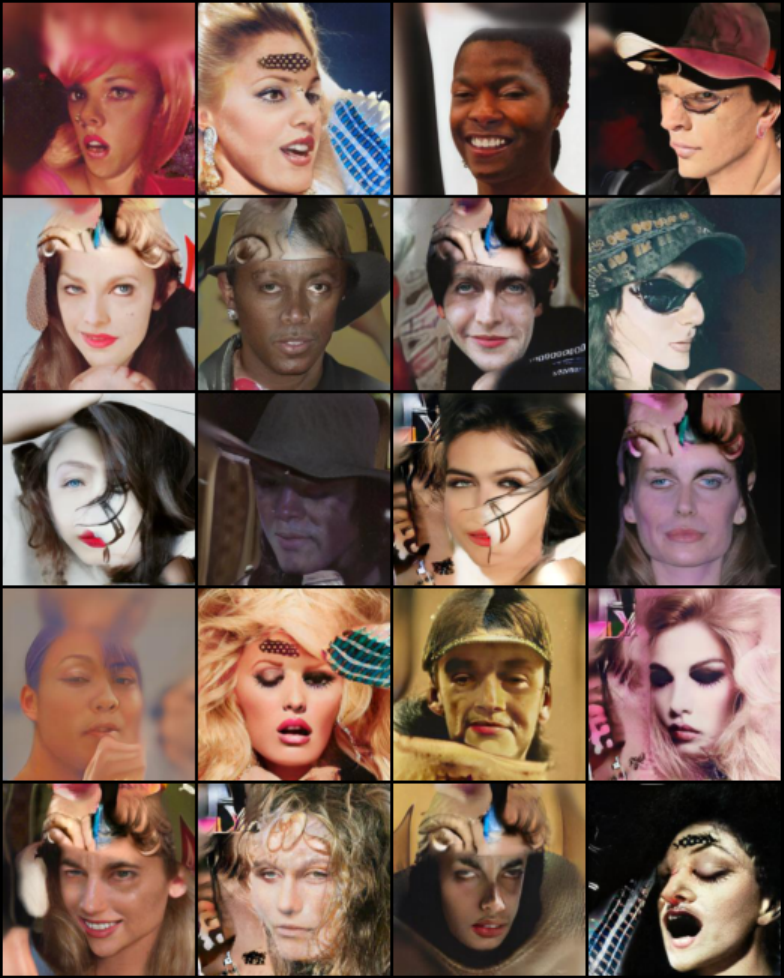}
         \caption{StyleSwin}
         \label{fig:RejectedStyleSwin}
     \end{subfigure}
     \caption{Examples of rejected samples during hand-labeling due to poor quality.}
    \label{fig:RejectedSamples}
\end{figure}

\paragraph{\textbf{Dataset Labeling Protocol.}}
To ensure high-quality samples and labels in our dataset, we passed the dataset through Amazon Mechanical Turk, where labelers were given detailed guidelines and examples for identifying the individual sensitive attributes. In addition to the sensitive attribute option \eg \texttt{Gender(Male)} or \texttt{Gender(Female)}, labelers were also given an ``unidentifiable'' option which they were instructed to select for low-quality samples, as per Fig, \ref{fig:RejectedSamples} and \ref{fig:SDMRejected}. We repeated this process for 4 runs \st each sample had the opinions of four independent labelers. Finally, each sample was assigned the label that the majority had selected.

\begin{figure}[h!]
\centering
     \begin{subfigure}[b]{0.49\textwidth}
         \centering
         \includegraphics[width=\textwidth]{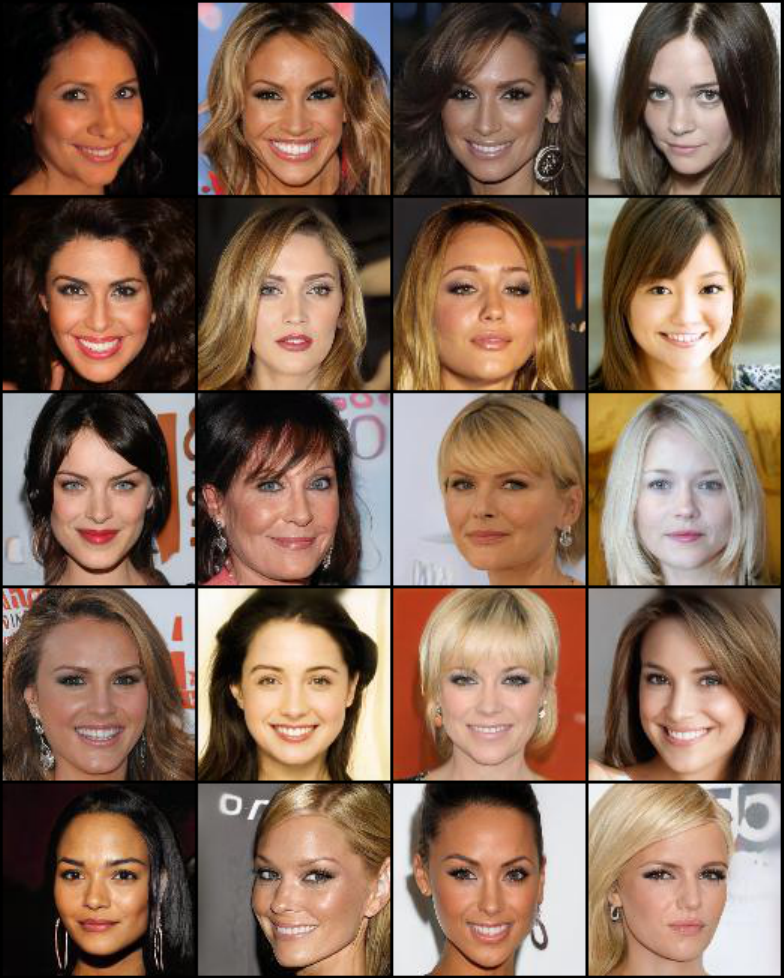}
         \caption{\texttt{Gender} (Female) Samples}
         \label{fig:GenderFemale}
     \end{subfigure}
     \hfill
     \begin{subfigure}[b]{0.49\textwidth}
         \centering
         \includegraphics[width=\textwidth]{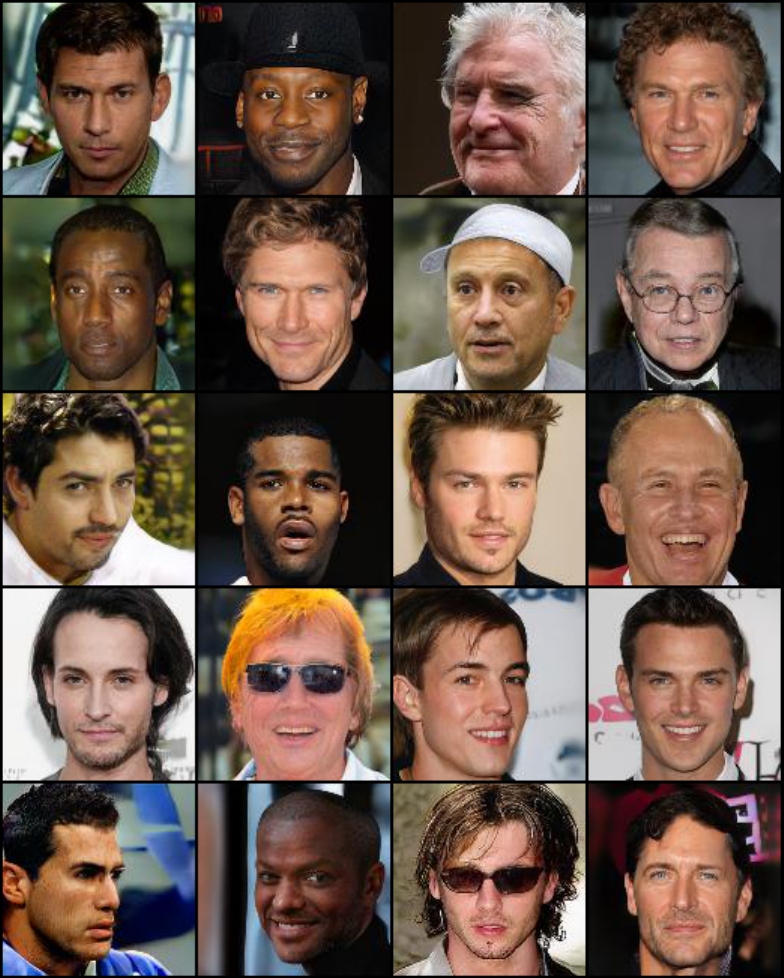}
         \caption{\texttt{Gender} (Male) Samples}
         \label{fig:GenderMale}
     \end{subfigure}
     \caption{Examples of samples \wrt \texttt{Gender} sensitive attribute in our 
     proposed GenData dataset.
     }
    \label{fig:GenderSamples}
\end{figure}

\begin{figure}[h!]
\centering
     \begin{subfigure}[b]{0.49\textwidth}
         \centering
         \includegraphics[width=\textwidth]{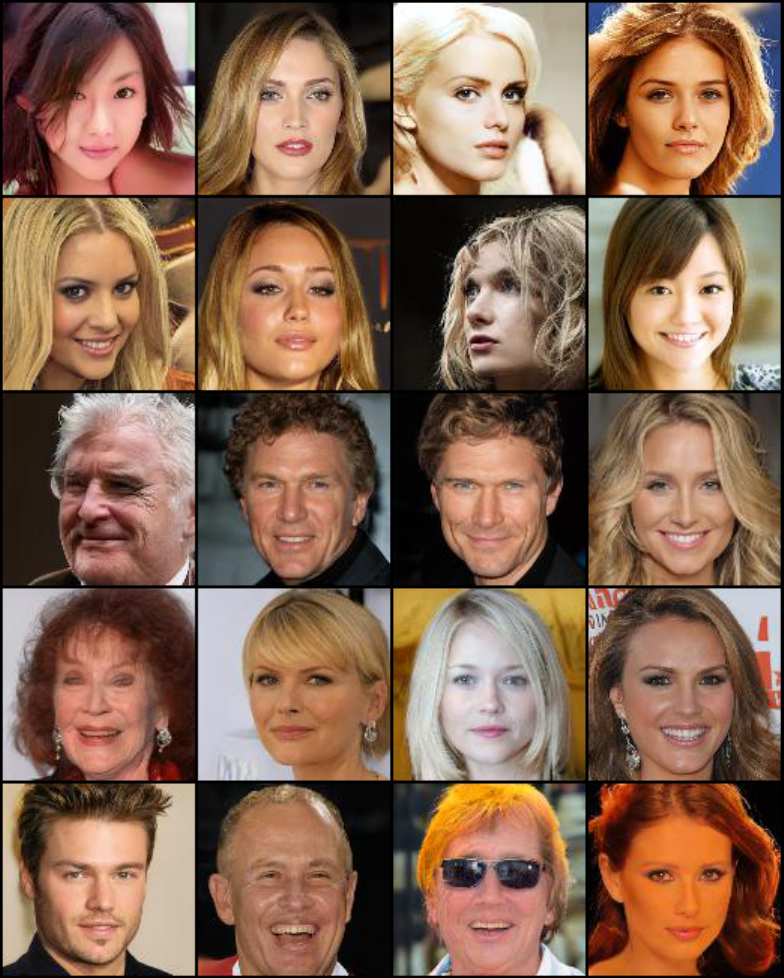}
         \caption{no-\texttt{BlackHair} Samples}
         \label{fig:noBlackhair}
     \end{subfigure}
     \hfill
     \begin{subfigure}[b]{0.49\textwidth}
         \centering
         \includegraphics[width=\textwidth]{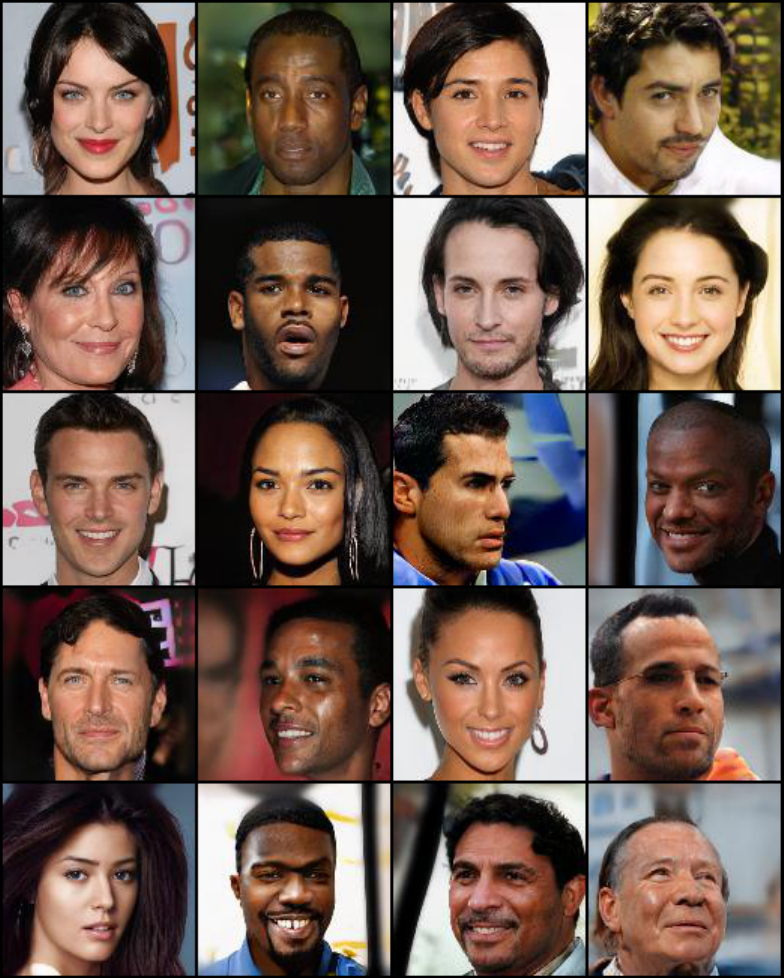}
         \caption{\texttt{BlackHair} Samples}
         \label{fig:Blackhair}
     \end{subfigure}
     \caption{Examples of samples \wrt \texttt{BlackHair} sensitive attribute
     in our 
     proposed GenData dataset.
     }
    \label{fig:BlackHairSamples}
\end{figure}

\begin{figure}[h]
     \centering
     \includegraphics[width=\textwidth]{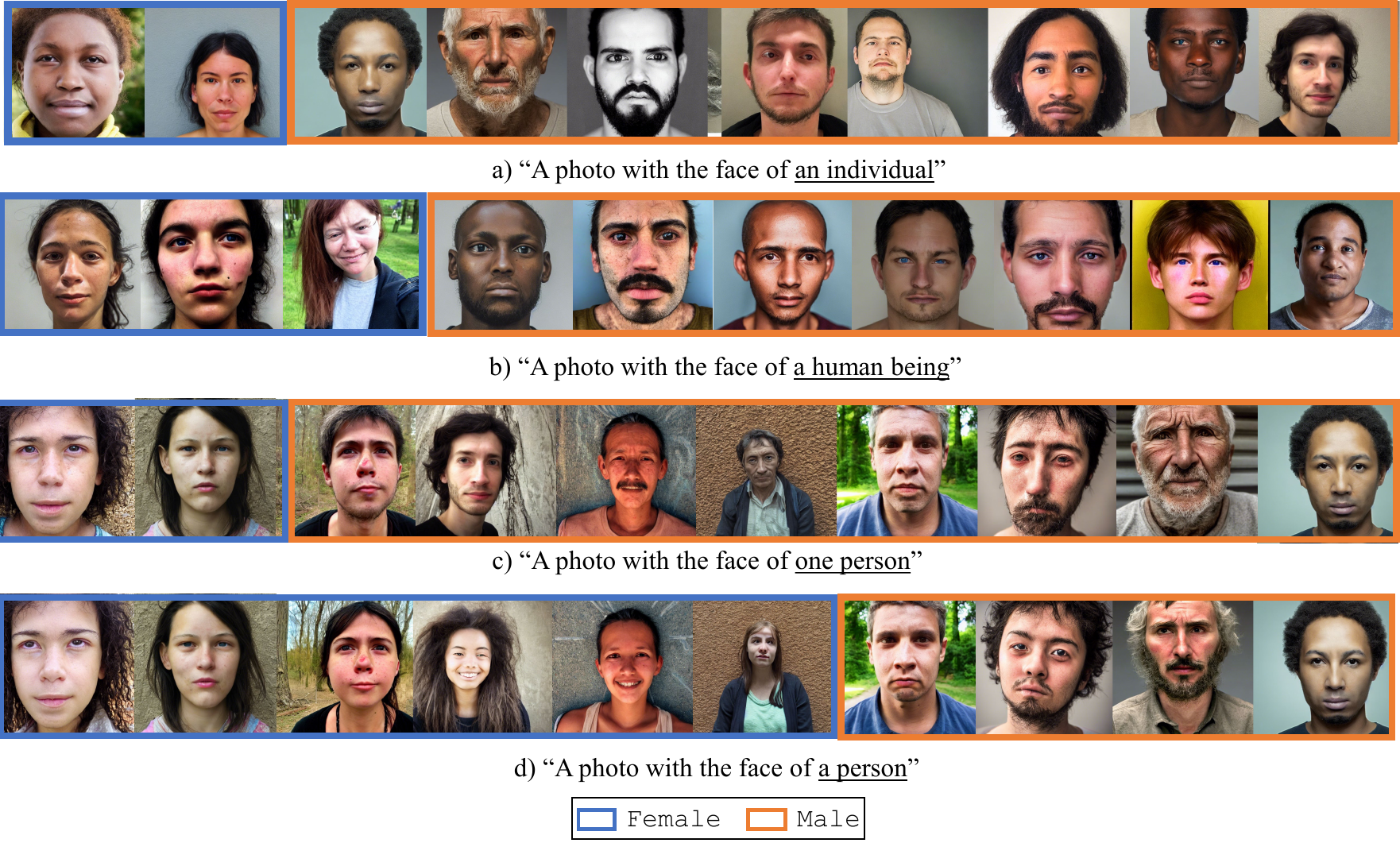}
     \caption{Examples of randomly generated samples based on the prompts "A photo with the face of \underline{an individual}" and "A photo with the face of \underline{a human being}" \wrt the sensitive attribute \texttt{Gender}. }
     \label{fig:SDMmoreillustration}
 \end{figure}

\begin{figure}[h]
     \centering
         \includegraphics[width=\textwidth]{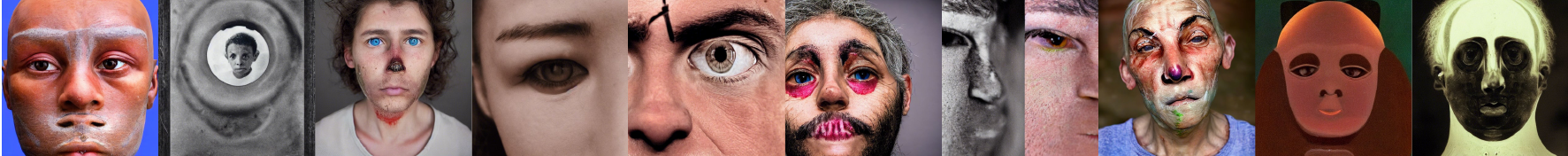}
     \caption{Examples of rejected samples from the SDM.}
     \label{fig:SDMRejected}
 \end{figure}

%\textcolor{blue}{
Overall, the GANs and SDM received 97\% and 99\%  unanimous agreement rates.
%}
This for example includes male, female, or unidentifiable, for the sensitive attribute \texttt{Gender}. We discard the samples that had been labeled unidentifiable and were left with a high-quality dataset as per Fig. \ref{fig:GenderSamples}, \ref{fig:BlackHairSamples} and \ref{fig:SDMmoreillustration}.
%\textcolor{blue}{
We remark that the discarded samples consist only a small portion of the generated samples \ie 3\% of the GANs, and 1\% of the SDM. 
Upon further evaluation, we found that the sensitive attribute classifiers appear to uniformly assign these (rejected) ambiguous samples a random class with low confidence. As a result, we can assume that the impact of disregarding these samples was insignificant to CLEAM's evaluation.
%}
%\textcolor{blue}{Furthermore, in addition to the fact that these samples make a small portion of generated samples, a further evaluation found that the classifiers appears to uniformly assign these (rejected) ambiguous samples a random class with low confidence. As a result, we can assume that the impact of disregarding these samples was insignificant to CLEAM's evaluation. }

\FloatBarrier
\clearpage
\section{Limitations and Considerations}
%\textcolor{red}{To include limitations of CLIP as a classifier}

%\textcolor{blue}{
{\bf Ethical consideration.} In general, we note that our work does not introduce any social harm but instead improves on the existing fairness measurement framework to better gauge progress. However, we stress that it is important to consider the limitations of the existing fairness measurement framework, which we discuss in the following.
%}

%\textcolor{blue}{
{\bf Sensitive Attribute Labels.}
Certain sensitive attributes may exist on a spectrum \eg \texttt{Young}. However, given that this work aims to improve fairness measurement, and the current widely used definition is based on binary outcomes, we utilize the same setup in our work. 
Additionally, it is also important to be aware that certain sensitive attributes may be ambiguous \eg \texttt{Big Nose} (which exist in popular datasets like CelebA-HQ), but definitions could differ based on different cultural expectations.
In our work, we try to select less ambiguous sensitive attributes e.g., \texttt{BlackHair}.
%}

%\textcolor{blue}{
{\bf Human and Auto Labelling.}
Labeling sensitive attributes in generative models is essential to better understand the possible biases that may exist in some proposed generative model algorithms. To do this, researchers often utilize either human labelers or machines for automated labeling. However, when utilizing such labeling procedures it is important to consider ethical implications, especially in many cases where sensitive information such as gender is involved.
One particular concern is that there could be potential discrimination in the assignment of labels such as gender. For example, if only certain facial features are considered when assigning gender labels, some individuals may be inaccurately labeled due to their unique characteristics that deviate from traditional notions of male and female identity.
%}

%\textcolor{blue}{
Human labelers may bring their own biases, subjectivity, and cultural background to the labeling process, which can lead to inaccuracies or reinforce stereotypes. Additionally, it is important to ensure that the labelers represent a diverse range of backgrounds and perspectives, particularly if the samples being labeled are from a diverse population. This can help mitigate potential discrimination against some social identities and improve the accuracy of the labeling process.
%}

%\textcolor{blue}{
In the case when utilizing machines for labeling, it is important to be aware that labeling algorithms may be biased, depending on the data set it was trained on. If the data set is not diverse or balanced, the algorithm may produce inaccurate or biased results that reinforce stereotypes or discrimination against certain social identities. 
%}

%\textcolor{blue}{
{\bf Utilizing Zero Shot Classifiers.} 
%consider application domain vs training domain
When utilizing pre-trained classifiers it is important to carefully select proxy validation dataset with a similar domain to the generated images. A significant mismatch in these two domains could result in an inaccurate approximation of $\balpha$, resulting in poor performance by CLEAM. 
%Ambigious sensitive attributes may mean different things between the validation dataset and the pre-trained classifier
Then similar to our previous discussion, we would also refrain from ambiguous sensitive attributes, as this may result in a mismatch between the proxy validation dataset and the pre-trained sensitive attribute classifier. 
%}

%we would recommend to refrain from measuring ambiguous sensitive attributes, as discussed earlier, as the proxy validation dataset and the generated images  }

%\newpage
%\input{Appendix sections/supp_header}

%%%%%%%%%%%%%%%%%%%%%%%%%%%%%%%%%%%%%%%%%%%%%%%%%%%%%%%%%%%%

\end{document}